\documentclass[11pt]{article}
\usepackage{acl}  

\usepackage{times}
\usepackage{latexsym}
\usepackage[T1]{fontenc}
\usepackage[utf8]{inputenc}
\usepackage{microtype}
\usepackage{inconsolata}
\usepackage{amsmath}
\usepackage{amssymb}
\usepackage{mathtools}
\usepackage{booktabs}
\usepackage{multirow}
\usepackage{multicol}
\usepackage{array}
\usepackage{graphicx}
\usepackage{xcolor}
\usepackage{colortbl}
\definecolor{heat}{RGB}{66,120,190}   
\definecolor{refgray}{gray}{0.5}     
\usepackage{url}
\usepackage{hyperref}
\usepackage{enumitem}
\usepackage{listings}
\usepackage{caption}
\usepackage{subcaption}
\usepackage{tabularx}
\usepackage[breakable]{tcolorbox}
\usepackage{tikz}
\usepackage{float}
\usepackage{rotating}
\usepackage{xspace}

\newcommand{\dialfork}{\textsc{DialFork}\xspace}
\newcommand{\sdialog}{\textsc{SDialog}\xspace}
\newcommand{\STAR}{\textsc{STAR}\xspace}

\newcommand{\eg}{\textit{e.g.,}\xspace}
\newcommand{\ie}{\textit{i.e.,}\xspace}

\newcommand{\llm}[1]{\texttt{#1}} 
\newcommand{\logo}[1]{\makebox[2.4ex][c]{\ensuremath{\vcenter{\hbox{\includegraphics[height=1.9ex,width=2.4ex,keepaspectratio]{img/logos/#1}}}}}\hspace{0.35em}} 

\newcommand{\htype}[1]{\textsc{#1}}
\newcommand{\WIC}{\htype{WIC}\xspace}    
\newcommand{\SSPC}{\htype{SSPC}\xspace}  
\newcommand{\GR}{\htype{GR}\xspace}      
\newcommand{\CSC}{\htype{CSC}\xspace}    

\newcommand{\flowkl}{\textsc{FlowKL}\xspace}
\newcommand{\flowdtw}{\textsc{FlowDTW}\xspace}
\newcommand{\flowscore}{\textsc{FlowScore}\xspace}
\newcommand{\NC}{\textsc{NC}\xspace}
\renewcommand{\SS}{\textsc{SS}\xspace}      
\newcommand{\IT}{\textsc{IT}\xspace}      

\tcbuselibrary{skins,breakable}
\definecolor{lightgray}{RGB}{240,240,240}
\tcbset{
  dialogbox/.style={
    breakable,
    colback=lightgray,
    colframe=gray!60,
    fonttitle=\small\bfseries,
    title=#1,
    left=4pt, right=4pt, top=4pt, bottom=4pt,
    boxrule=0.5pt,
    arc=2pt
  }
}

\definecolor{promptbg}{RGB}{246,248,251}
\definecolor{promptrule}{RGB}{66,120,190}
\definecolor{phcolor}{RGB}{176,58,46}     
\definecolor{jsonstr}{RGB}{30,120,70}
\lstdefinestyle{prompt}{
  basicstyle=\footnotesize\ttfamily\color{black!85},
  backgroundcolor=\color{promptbg},
  frame=l, framerule=1.6pt, rulecolor=\color{promptrule},
  framesep=5pt, xleftmargin=7pt, xrightmargin=0pt,
  aboveskip=8pt, belowskip=6pt,
  columns=fullflexible, keepspaces=true,
  breaklines=true, breakatwhitespace=true,
  breakindent=0pt,
  showstringspaces=false,
  moredelim={[s][\bfseries\color{phcolor}]{\{\{}{\}\}}},
  moredelim={[s][\bfseries]{**}{**}},
}
\lstdefinestyle{promptsmall}{style=prompt, basicstyle=\scriptsize\ttfamily\color{black!85}}
\lstdefinestyle{json}{
  style=promptsmall,
  breakindent=4em,
  moredelim={[s][\color{jsonstr}]{"}{"}},
}
\lstdefinestyle{transcript}{
  style=promptsmall, breakindent=2ex,
  moredelim={[s][\bfseries\color{promptrule}]{[}{]}},
}
\definecolor{happycolor}{RGB}{60,60,60}
\definecolor{wiccolor}{RGB}{180,50,50}
\definecolor{sspccolor}{RGB}{50,100,180}
\definecolor{grcolor}{RGB}{30,140,80}
\definecolor{csccolor}{RGB}{160,80,0}
\definecolor{intendedcolor}{HTML}{5E5E5C}
\definecolor{corruptedcolor}{HTML}{1C5BA8}

\title{\raisebox{-0.2em}{\includegraphics[height=1em]{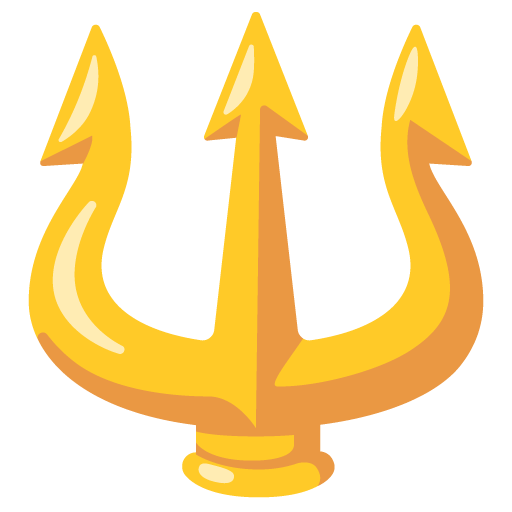}}\,\dialfork: Going Beyond Factuality to Evaluate Procedural Hallucination in Conversational Agents}

\author{
  Sergio Burdisso$^{1}$ \quad
  Esaú Villatoro-Tello$^1$ \quad
  Dairazalia Sánchez-Cortés$^1$ \\
  {\bf Yanis Labrak}$^1$ \quad
  {\bf Jordi Luque}$^3$ \quad
  {\bf Milos Cernak}$^2$ \quad
  {\bf Petr Motlicek}$^{1,4}$ \\
  \textsuperscript{1}Idiap Research Institute, Martigny, Switzerland \\
  \textsuperscript{2}Logitech, Lausanne, Switzerland \\
  \textsuperscript{3}Telefónica Innovación Digital, Barcelona, Spain \\
  \textsuperscript{4}Brno University of Technology, Brno, Czech Republic \\
  \texttt{sergio.burdisso@idiap.ch}
}

\begin{document}

\maketitle

\begin{abstract}
Large language models (LLMs) deployed as conversational agents are increasingly expected to follow procedural logic. While factual hallucination has received extensive attention, a different failure mode has been overlooked: \emph{procedural hallucination}, in which an agent produces individually fluent and factually plausible utterances while deviating from the intended conversational process. We introduce \dialfork, the first benchmark to measure this failure mode on raw dialogue text. \dialfork is built by \emph{forking} conformant dialogues at controlled turn positions and injecting one of four hallucination types, yielding 36{,}278 dialogues across 24 tasks in 12 domains; 30 human annotators validate the construction. We define three evaluation tasks that characterize procedural deviation at three granularities: how far a set of dialogues deviates from conformant ones (monitoring deployed agents), how far a single dialogue does (procedural retrieval), and at which turn a dialogue first goes off-track (real-time guardrails). For the first two, we propose dialog-flow-aware metrics, \flowkl and \flowscore, that compare dialogues by \emph{how they unfold} rather than their \emph{content}. At the dialogue level, \flowscore significantly outperforms all baselines across seven encoders, showing that semantic similarity is no substitute for procedural similarity. We release \dialfork under open licenses at \href{https://github.com/idiap/dialfork}{\nolinkurl{github.com/idiap/dialfork}}, hoping the community will measure conversational \emph{process} alongside content.
\end{abstract}

\section{Introduction}
\label{sec:intro}
As LLMs are deployed in increasingly agentic roles (customer support, scheduling, guided troubleshooting, clinical intake), they are implicitly expected to follow structured procedural logic: a sequence of prescribed steps that moves a conversation from a user's initial request to a completed goal~\citep{ijcai2024p890,10.1007/978-3-031-93418-6_9}. This expectation extends well beyond classical task-oriented dialogues (TOD); it
applies to any system where the \emph{process} of a conversation matters as much as its content.

The dominant conception of hallucination in LLMs concerns \emph{factual} errors~\citep{maynez-etal-2020-faithfulness, ji2023survey,10.1145/3703155}, for instance: confident assertions of incorrect dates, invented product details, or wrong numbers. 
A qualitatively different and practically critical failure mode has, however, received comparatively little attention: an agent that produces entirely truthful, fluent utterances while violating the procedural logic it was instructed to follow. The closest prior notions, hallucination in inter-agent communication~\citep{lin2025llmbasedagents} and faithfulness to grounding knowledge in dialogue~\citep{dziri-etal-2022-faithdial}, concern the accuracy of \emph{content}, not adherence to a conversational \emph{process}.

We call this \textbf{procedural hallucination}: a breakdown in \emph{process} rather than \emph{content}. A procedurally hallucinating agent may skip required steps, pursue a different goal than instructed, repeat already-completed actions, or introduce conflicting objectives that derail task completion, all while generating individually plausible responses. Crucially, this failure is \emph{orthogonal} to factual correctness: no single utterance need be false for the overall conversation to be wrong. Standard hallucination benchmarks~\citep{li-etal-2023-halueval,lin-etal-2022-truthfulqa} are blind to such failures because they evaluate the truthfulness of content, not the conversational process.

\begin{figure}[t]
  \centering
  \includegraphics[width=\linewidth]{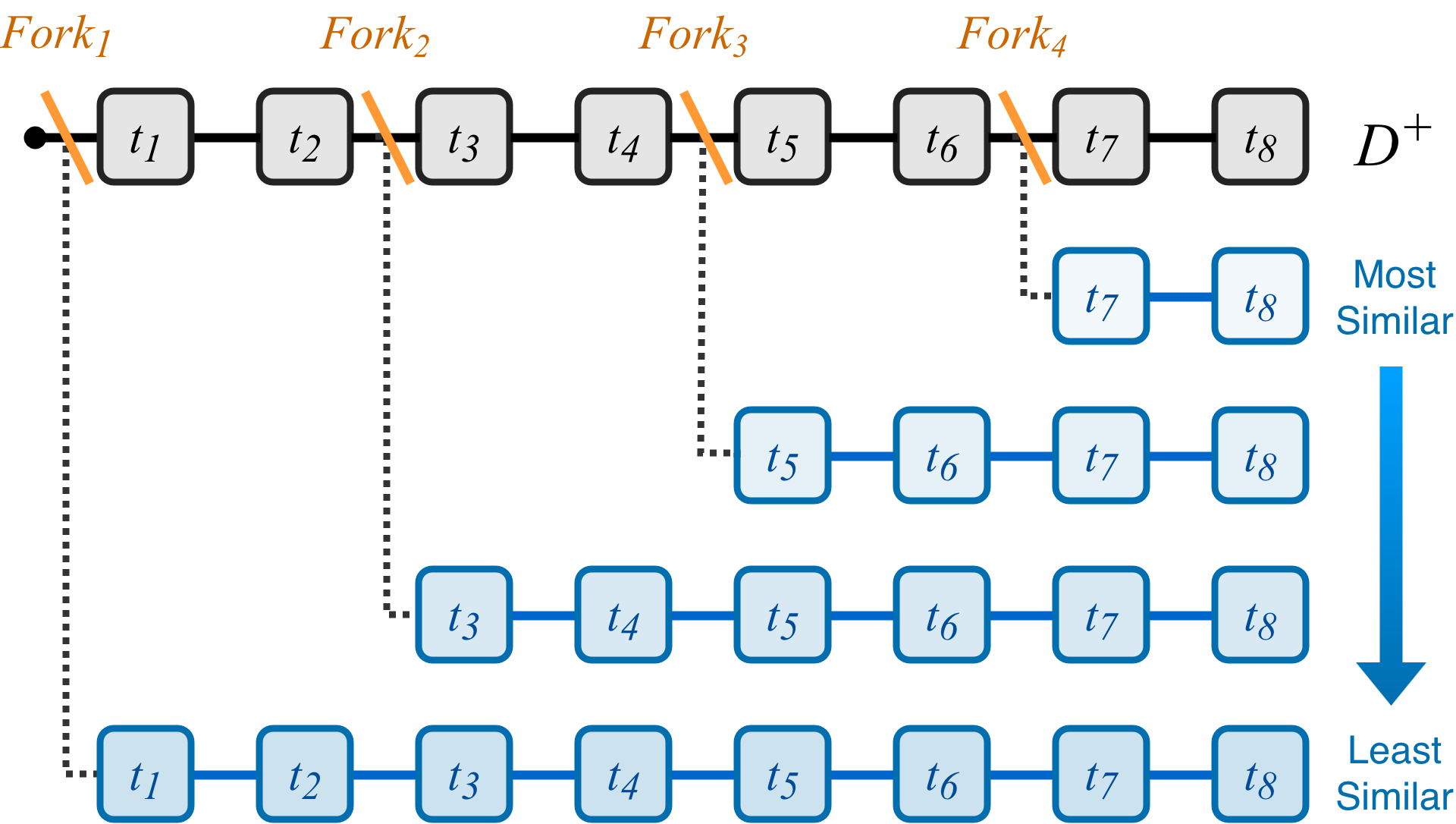}
  \caption{Construction of \dialfork by controlled forking: the simulation of a conformant dialogue $D^+$ is re-run with the same seed, reproducing its prefix, and a hallucination instruction is injected just before the fork turn. Earlier forks deviate more, inducing a ground-truth ordering by \emph{procedural} (dialog-flow) similarity.} 
  \label{fig:overview}
\end{figure}

To address this gap, we present \dialfork, a benchmark constructed by \emph{forking} dialogues at controlled turn positions (Figure~\ref{fig:overview}) to simulate different types of procedural hallucinations.
Our main contributions are the following: (a) We define \emph{procedural hallucination} as a phenomenon distinct from factual hallucination, and decompose it into four complementary types grounded in a general workflow model (\S\ref{sec:taxonomy}). (b) We release \dialfork,\footnote{\url{https://github.com/idiap/dialfork}: dataset, generation pipeline, evaluation code, and annotation tool.} a 36,278-dialogue benchmark constructed via controlled forking that spans 24 tasks across 12 domains and four temporal deviation depths, along with its seed-controlled multi-agent generation pipeline, applicable to any workflow-annotated dialogue dataset, and validate its construction in a human study with 30 annotators (\S\ref{sec:dataset}). (c) We define three evaluation tasks that characterize procedural deviation at three granularities, directly from raw dialogue text: how far a set of dialogues deviates from conformant ones, how far a single dialogue does, and at which turn a dialogue first goes off-track (\S\ref{sec:tasks}). (d) Enabled by the new dataset, we introduce \flowkl and \flowscore, two dialog-flow-aware distance metrics that compare dialogues by how they unfold rather than by their content, and report initial baselines on all three tasks as a starting point for future work (\S\ref{sec:experiments}).

\section{Related Work}
\label{sec:related}

\paragraph{Hallucination in LLMs.} The dominant view concerns factual accuracy of generated content~\citep{maynez-etal-2020-faithfulness, kryscinski-etal-2020-evaluating, lin-etal-2022-truthfulqa, mallen-etal-2023-trust}. Surveys~\citep{ji2023survey, zhang2025siren,10.1145/3703155} taxonomize hallucination along faithfulness and factuality dimensions, and dialogue-level benchmarks such as DiaHalu~\citep{chen-etal-2024-diahalu} extend them to multi-turn conversations; all these categories, however, concern \emph{what} is said, not the correctness of \emph{how} a conversation unfolds. 


\paragraph{Procedural adherence and dialogue similarity.} A parallel line studies whether systems follow prescribed steps, from policy-constrained action sequences~\citep{chen-etal-2021-action} and procedure-guided action prediction~\citep{white-etal-2023-leveraging} to condition--action instruction following~\citep{ghazarian2025} and workflow generation~\citep{ICLR2025_adbe9369}, typically over annotated actions, intents, or slots. Dialogue similarity is largely text similarity: embeddings collapse the turns into one vector~\citep{reimers-gurevych-2019-sbert,zhou-etal-2022-dse} compared by cosine, discarding turn order; TaskDiff~\citep{bhaumik-etal-2023-taskdiff} relies on annotated intents and slots. These benchmark how well models follow procedural instructions; \dialfork instead benchmarks methods that measure how far a conversation \emph{deviates} from a procedure, \emph{directly on raw dialogue text}, whether an LLM or a human produced it.

\section{Procedural Hallucination}
\label{sec:taxonomy}






\subsection{Formal Model}

\begin{table*}[t]
\centering
\small
\setlength{\tabcolsep}{3pt}
\caption{Procedural hallucination types and the components of $\mathcal{W} = (\mathcal{S}, \mathcal{A}, \delta, \mathcal{O})$ they corrupt (\checkmark: preserved; {\color{red}$\times$}: replaced; $\sim$: partially overridden). Simplified example (a real one is in Appendix~\ref{app:examples}): after the shared opening, each row shows only the first two turns from the fork turn $t_k$ on; every version then continues (\ldots) as a full dialogue. Gray: intended workflow; blue: corrupted.}
\label{tab:taxonomy}
\newcommand{\pk}{{\color{gray}\checkmark}}%
\newcommand{\px}{{\color{red}$\boldsymbol{\times}$}}%
\begin{tabularx}{\textwidth}{@{}lcccl>{\raggedright\arraybackslash}X@{}}
\toprule
\textbf{Type} & $(\mathcal{S},\mathcal{A})$ & $\delta$ & $\mathcal{O}$ & \textbf{What is broken} & \textbf{Example} (simplified): first two turns after the fork \\
\midrule
\multicolumn{5}{@{}l}{\textit{\color{intendedcolor}Shared opening before the fork}} & \textcolor{intendedcolor}{\textbf{User:} I'd like to book a viewing. \textbf{System:} Your name? \textbf{User:} Mark.} \\
\addlinespace[2pt]
Conformant & \pk & \pk & \pk & --- (original) & \textcolor{intendedcolor}{\textbf{System:} Which apartment do you want to view? \textbf{User:} Center Apts.} \ldots \\
\WIC  & \px & \px & \pk & Workflow identity & \textcolor{corruptedcolor}{\textbf{System:} Let's search apartments! budget? \textbf{User:} No, just a viewing!} \ldots \\
\SSPC & \pk & \px & \pk & Step selection & \textcolor{corruptedcolor}{\textbf{System:} Your viewing is booked! \textbf{User:} What?! Which one?!} \ldots \\
\GR   & \pk & \pk & \px & Task goal & \textcolor{corruptedcolor}{\textbf{System:} Want to try co-living instead? \textbf{User:} No, no, I already know} \ldots \\
\CSC  & \pk & \pk & $\sim$ & Goal conflict & \textcolor{corruptedcolor}{\textbf{System:} Just to confirm, your name is Mark? \textbf{User:} Yes, Mark.} \ldots \\
\bottomrule
\end{tabularx}
\end{table*}

Most conversational systems do not declare their procedural logic explicitly: states, available actions, and transitions are implicit in the agent's instructions, the underlying business process, or the assumed conversational norms. To talk about \emph{deviations} from this logic precisely, we introduce a lightweight abstraction that does not assume any particular implementation. We model the intended logic of a conversational agent as a tuple $\mathcal{W} = (\mathcal{S}, \mathcal{A}, \delta, \mathcal{O})$, where $\mathcal{S}$ is a set of conversational states (\eg steps or phases of the interaction), $\mathcal{A}$ is a set of agent actions (\eg the communicative functions it can perform), $\delta: \mathcal{S} \times \mathcal{U} \rightarrow \mathcal{S}$ is a transition function from the current state and user input $\mathcal{U}$ to the next state, and $\mathcal{O}$ is the agent's objective.

This formulation is intentionally general. It applies to any goal-directed conversational agent. For example, a task-oriented dialogue system has the dialogue policy as $\delta$ and task completion (\eg booking a hotel room, retrieving a balance) as $\mathcal{O}$; a medical-intake assistant has a clinical protocol as $\delta$ and patient triage as $\mathcal{O}$; a customer-support agent has a troubleshooting procedure as $\delta$ and issue resolution as $\mathcal{O}$. A \emph{conformant} dialogue $D^+$ is one whose turns follow $\mathcal{W}$.

\subsection{Hallucination Types}

We define four complementary types, each targeting a distinct component of $\mathcal{W}$ (summarized in Table~\ref{tab:taxonomy}, with an example of each):

\begin{description}[nosep, leftmargin=0pt]
\item[\textbullet~Workflow Identity Corruption (\WIC).] $\mathcal{W}$ is replaced by a different workflow $\mathcal{W}'$; the agent follows a valid workflow but the wrong one (modeling task disambiguation failures and prompt injection).

\item[\textbullet~Step Selection Policy Corruption (\SSPC).] $\mathcal{O}$ and $(\mathcal{S}, \mathcal{A})$ are preserved but $\delta$ is replaced by a corrupted policy $\delta'$ that, \eg repeats, skips, loops, or randomizes steps.

\item[\textbullet~Goal Replacement (\GR).] $\mathcal{W}$ is preserved but the objective $\mathcal{O}$ is replaced by a qualitatively different goal $\mathcal{O}'$ (\eg collecting feedback instead of completing a booking).

\item[\textbullet~Conflicting System Constraints (\CSC).] $\mathcal{O}$ is retained but augmented with a competing auxiliary objective $\mathcal{O}_c$ that intermittently overrides $\mathcal{O}$ (\eg always responding with a question, or confirming every detail).
\end{description}

In practice, for both diversity and fine-grained controllability, we define a small pool of concrete variants per hallucination type (five corrupted policies for \SSPC such as \textit{repeat}, \textit{skip}, and \textit{invert order}, six alternative goals for \GR, and six conflicting constraints for \CSC); the full lists are in Appendix~\ref{app:variants}.

\section{The \dialfork Dataset}
\label{sec:dataset}

\begin{figure*}[t]
  \centering
  \includegraphics[width=\textwidth]{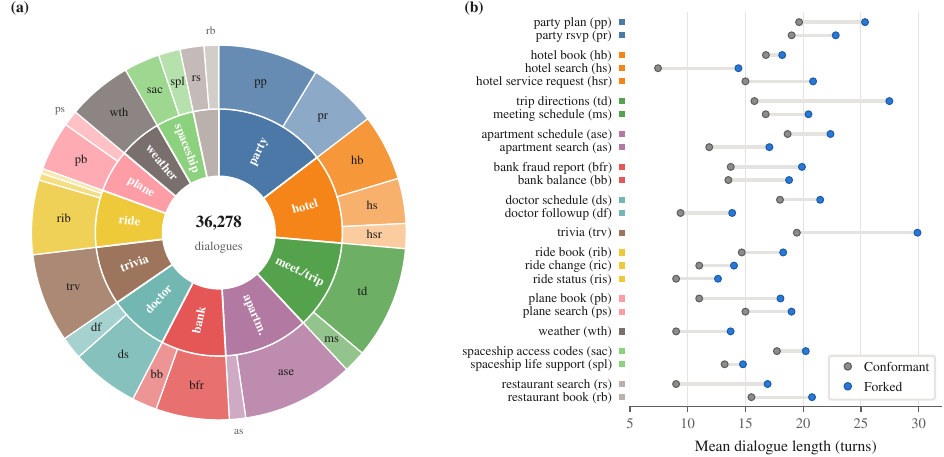}
  \caption{Per-task composition of \dialfork. (a)~Number of dialogues per \STAR domain (inner ring) and task (outer ring; abbreviations as in (b)). Since every conformant dialogue yields 16 forks, the proportions are the same for conformant and forked dialogues. (b)~Mean dialogue length per task, conformant vs.\ forked, grouped by domain in the same order and colors as (a).}
  \label{fig:task_stats}
\end{figure*}

\subsection{Construction Pipeline}

\paragraph{STAR as a generation basis.} We base \dialfork on \STAR~\citep{mosig-etal-2020-star}, which spans 24 tasks across 12 domains (hotel, ride, restaurant, medical, party, spaceship, trivia, weather, etc.). \STAR is uniquely suited to our pipeline because its dialogues were originally produced by \emph{human role-play}: for each dialogue, one human participant was given the system role, with the task's workflow graph and per-node response templates (Appendix~\ref{app:workflow}), and another was given the user role, with a persona, a task, and a goal; the two then conversed in these roles. Crucially, \STAR releases not only the resulting dialogues but also these role-play instructions, exactly as they were given to the two human participants of each dialogue (see Appendix~\ref{app:roleplay_example} for a complete example). We give the same instructions to two LLM agents that take the place of the humans: the system agent receives the instructions given to the human who played the system, and the user agent those given to the human who played the user. To our knowledge, no other public dialogue corpus releases both an explicit workflow for every task, which serves as procedural ground truth, and the instructions given to the humans who produced each conversation. Since every \dialfork dialogue is newly generated, only \STAR's instructions and workflows, not its original dialogues, enter the benchmark.

\paragraph{Conformant dialogue generation.} For each \STAR scenario, we use \sdialog~\citep{burdisso-etal-2025-sdialog} to run the two agents: the \textbf{system agent} is instructed to follow the workflow faithfully, and the \textbf{user agent} to interact naturally without volunteering all details upfront. Both agents are \llm{Gemma4-31B}~\citep{gemma4_31b_2026} served locally\footnote{Selected among several open-weight LLMs from pilot dialogues (Appendix~\ref{app:generation}). An open-weight model stays fixed (closed models may change behind the same name), lets anyone regenerate or extend \dialfork, and avoids the cost of a commercial API at this scale.}; prompts and full implementation details are in Appendix~\ref{app:generation}. The two agents converse until the task is completed or a per-task turn budget is reached. After filtering for dialogues that include enough system turns to admit all four fork positions, we retain an initial set of \textbf{2,134 valid conformant dialogues}.

\paragraph{Forked hallucination generation via seeded replay.} For each conformant dialogue $D^+$ with $T$ system turns, we identify four \emph{fork points} at evenly-spaced positions $t_k = 1 + \operatorname{round}((k-1) \cdot T / 4)$, $k \in \{1,2,3,4\}$. To produce a forked variant, we re-run the same scenario with the same random seed, so that turns $1$ through $t_k - 1$ are again generated under the intended workflow and reproduce those of $D^+$;\footnote{Up to minor rewordings that the LLM serving stack may introduce despite the fixed seed. These do not affect the construction: every turn before $t_k$ is generated under the intended workflow, and every turn from $t_k$ on under the corrupted one.} before turn $t_k$ is generated, a hallucination instruction is added as a system message to the system agent's context, specifying the type and the sampled variant (drawn from the per-type pool defined in \S\ref{sec:taxonomy}; full prompts in Appendix~\ref{app:prompts}). From $t_k$ onward, the system agent follows the injected instruction and produces turns that diverge from $D^+$ in the manner specified by the chosen hallucination type (Table~\ref{tab:taxonomy}; worked examples in Appendix~\ref{app:examples}). Earlier forks produce more deeply deviant dialogues, while later forks produce subtler deviations%
, naturally inducing a similarity ranking over the forked dialogues (Figure~\ref{fig:overview}).

\begin{figure}[t]
  \centering
  \includegraphics[width=\linewidth]{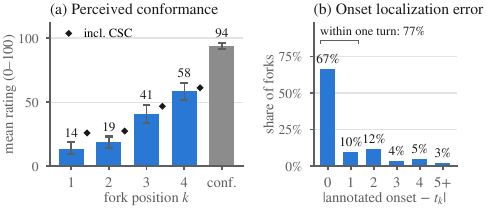}
  \caption{Human validation of \dialfork (Appendix~\ref{app:human}). (a)~Mean blind conformance rating per fork position $k$ for \WIC, \SSPC, and \GR forks (bars, with 95\% bootstrap CIs) and for all four types (markers), and for conformant dialogues. (b)~Distance, in assistant turns, between the annotated onset of the injected behavior and the fork turn $t_k$.}
  \label{fig:human_main}
\end{figure}

\paragraph{Quality assurance.} \dialfork's usefulness rests on a single property: \emph{forking must produce controlled deviation from the conformant baseline}. Automatic metrics confirm it across the whole dataset (Task~1, \S\ref{sec:experiments}), but cannot tell \emph{how} or \emph{when} each fork deviates. We therefore ran a human study: 30 annotators judged 428 conversations from five tasks (526 judgments), answering in stages so that they rated each conversation before learning how it was built. The study confirms the four labels the benchmark relies on. (i)~\emph{Fork order:} for the types that break the workflow (\WIC, \SSPC, \GR), blind conformance ratings rise with fork position, from 14/100 at $k{=}1$ to 58 at $k{=}4$ and 94 for conformant dialogues ($\tau_b{=}0.64$ across the four forks of a dialogue; Figure~\ref{fig:human_main}a). (ii)~\emph{Hallucination type:} told which instruction had been injected, annotators confirmed it in 91\% of forks, vs.\ 8\% for foil instructions on conformant dialogues. (iii)~\emph{Fork turn:} they placed the onset of the injected behavior within one turn of $t_k$ in 77\% of cases (exactly at $t_k$ in 67\%; Figure~\ref{fig:human_main}b). (iv)~\emph{Prefix:} the turns before the fork follow the workflow in 95\% of cases. Appendix~\ref{app:human} details the study; we release its annotation tool and anonymized judgments. An LLM judge calibrated on them, excluding the generator, extends these checks to the whole corpus (Appendix~\ref{app:human_judge}).

\subsection{Dataset Statistics}

\begin{figure}[t]
  \centering
  \includegraphics[width=\linewidth]{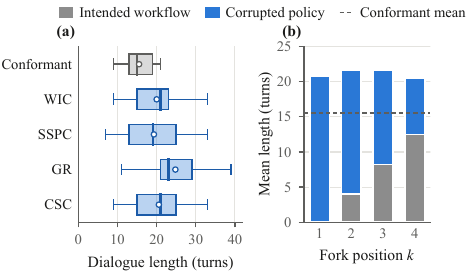}
  \caption{\dialfork statistics. (a)~Dialogue length per split (box: interquartile range; whiskers: 5th--95th percentiles; line: median; dot: mean). (b)~Mean number of turns per fork position $k$, split into turns generated before the fork, under the intended workflow, and after it, under the corrupted policy (averaged over the four hallucination types).}
  \label{fig:dataset_stats}
\end{figure}

Each of the 2,134 conformant dialogues yields 16 forked variants (4 hallucination types $\times$ 4 fork positions), for a total of \textbf{36,278 dialogues} (about 750K turns and 10.4M words, generated in about 33 GPU-days), 8,536 per hallucination type (per-task composition in Figure~\ref{fig:task_stats}; exact counts in Tables~\ref{tab:forkbench_composition} and~\ref{tab:forkbench_task_details}, Appendix~\ref{app:per_task_stats}). Figure~\ref{fig:dataset_stats}a shows that forked dialogues are longer than conformant ones (21.18 vs.\ 15.52 turns on average), 
with \GR the longest (verbose goals such as information collection add content) and \SSPC the shortest (skipped or repeated steps add little content). As $k$ increases (Figure~\ref{fig:dataset_stats}b), the intended-workflow prefix grows from 0 to 12.5 turns while the corrupted continuation shrinks from 20.9 to 8.0, keeping total length roughly constant.

\section{Evaluation Tasks and Methods}
\label{sec:tasks}

We define three evaluation tasks on \dialfork at increasing levels of granularity: set-level, dialogue-level, and turn-level. The first two probe procedural (dialog-flow) deviation as a measurable distance, an axis distinct from semantic similarity: two dialogues can be near-identical semantically (\eg same topics, same vocabulary) yet follow entirely different conversational procedures, and vice versa. The third localizes the deviation, asking where in a dialogue a hallucination first manifests. Throughout, dialogues are raw transcripts, with no annotations of any kind and no access to the model that produced them: the tasks measure procedural deviation in the text itself, whatever its source (an LLM agent or a human operator), rather than how prone a particular model is to deviate.
\subsection{Task 1: Set-to-Set Distance}
\label{subsec:task1}
\begin{table}[t]
\centering
\small
\setlength{\tabcolsep}{2pt}
\caption{Task 1 (Set-to-Set Distance): Kendall's $\tau$ between predicted and ground-truth fork-position rankings, averaged across the 24 tasks. Top block: controls; \textbf{bold}: best per column; shading: darker is better. All three distance metrics are significantly above chance ($p<0.001$); differences among them are not significant ($p \ge 0.43$; Appendix~\ref{app:stats}).}
\label{tab:s2s}
\begin{tabular}{lccccc}
\toprule
\textbf{Method} & \textbf{\WIC} & \textbf{\SSPC} & \textbf{\GR} & \textbf{\CSC} & \textbf{Avg.} \\
\midrule
\textcolor{refgray}{Random}            & \textcolor{refgray}{0.000} & \textcolor{refgray}{0.000} & \textcolor{refgray}{0.000} & \textcolor{refgray}{0.000} & \textcolor{refgray}{0.000} \\
\textcolor{refgray}{Turn-count diff.} & \textcolor{refgray}{$-$0.076} & \textcolor{refgray}{$-$0.340} & \textcolor{refgray}{0.278} & \textcolor{refgray}{$-$0.208} & \textcolor{refgray}{$-$0.087} \\
\textcolor{refgray}{Word-count diff.} & \textcolor{refgray}{0.222} & \textcolor{refgray}{0.208} & \textcolor{refgray}{0.583} & \textcolor{refgray}{$-$0.208} & \textcolor{refgray}{0.201} \\
\midrule
FBD          & \cellcolor{heat!38}\textbf{0.972} & \cellcolor{heat!25}0.792 & \cellcolor{heat!39}\textbf{0.986} & \cellcolor{heat!33}\textbf{0.903} & \cellcolor{heat!33}\textbf{0.913} \\
PRD          & \cellcolor{heat!35}0.931 & \cellcolor{heat!23}0.764 & \cellcolor{heat!39}\textbf{0.986} & \cellcolor{heat!28}0.833 & \cellcolor{heat!31}0.878 \\
\midrule
\flowkl       & \cellcolor{heat!29}0.847 & \cellcolor{heat!33}\textbf{0.903} & \cellcolor{heat!36}0.944 & \cellcolor{heat!28}0.833 & \cellcolor{heat!31}0.882 \\
\bottomrule
\end{tabular}
\end{table}

Consider a company that deploys an LLM agent to handle its routine customer requests, or that upgrades its assistant to a newer model. Before deployment, it needs to know whether the new agent's conversations unfold like the reference ones (\eg past conversations handled by its human agents or by the previous assistant), and, after deployment, whether they drift over time. Neither question concerns single dialogues; Task~1 captures this setting. \emph{Input:} two sets of dialogues, a reference and a target; \emph{output:} a scalar distance between them. To evaluate it with \dialfork, for each hallucination type $h$ and each of the 24 tasks of \dialfork, we group the forked dialogues by their fork position to obtain four sets $\mathcal{D}^h_k$ ($k\in\{1,2,3,4\}$), and take the set of conformant dialogues for that task, $\mathcal{D}^+$, as the reference. Writing $\mu_k \equiv \mu(\mathcal{D}^+, \mathcal{D}^h_k)$ for the distance under set-distance metric $\mu$, we expect the natural \emph{procedural-similarity} ordering $\mu_1 > \mu_2 > \mu_3 > \mu_4$, since earlier forks share less of the conformant prefix and are therefore more procedurally deviant. We report Kendall's $\tau$~\citep{kendall1938measure} between the metric-induced ranking and this ground-truth ordering, averaged across the 24 tasks. Because this ordering follows directly from how forks are constructed, Task~1 also serves as a \emph{sanity check} of \dialfork itself: if independent metrics recover it, the forked dialogues indeed depart from the conformant ones at their designated positions.

\paragraph{Baselines.} To the best of our knowledge, FBD and PRD, introduced together by \citet{xiang-etal-2021-assessing}, are the only two existing distribution-wise metrics specifically designed to compare sets of dialogues. Both operate on embeddings of consecutive (system, user) turn pairs extracted by a pretrained language model. \textbf{FBD} (Fr\'echet BERT Distance) fits a multivariate Gaussian to each set's embedding distribution and reports the Fr\'echet distance between the two Gaussians. \textbf{PRD} (Precision--Recall Distance) treats the two sets as distributions and reports the maximum harmonic mean of precision and recall over a sweep of relative-density thresholds~\citep{sajjadi-etal-2018-prd}.

\paragraph{Controls.} Since forks are longer than conformant dialogues (Figure~\ref{fig:dataset_stats}), \textbf{Turn-/Word-count diff.} use as distance $|\overline{\mathrm{len}}(\mathcal{D}^h_k) - \overline{\mathrm{len}}(\mathcal{D}^+)|$, the difference in mean length (in turns or words), checking whether length alone recovers the ordering. \textbf{Random} is the exact chance level.

\paragraph{\flowkl.} We propose \flowkl, an interpretable alternative grounded in an explicit dialog flow graph $G$, computed in three steps:
\begin{enumerate}[nosep, leftmargin=*]
\item \textbf{Graph construction.} Following \citet{burdisso-etal-2024-dialog2flow}, we cluster all utterances in $\mathcal{D}^+$ in the embedding space of Dialog2Flow (D2F), an encoder that groups utterances by communicative function, to obtain workflow-step nodes; directed edges encode observed transitions with frequency weights, yielding a bigram model $P_G(n_t \mid n_{t-1})$ over node sequences (Figure~\ref{fig:dialog2flow_graph} in Appendix shows the actual graph induced for the \textit{doctor\_schedule} task).
\item \textbf{Per-dialogue flow likelihood.} 
Each turn $u_t$ of a dialogue $D=(u_1,\ldots,u_T)$ is mapped to the node $n_t$ whose centroid is nearest to its D2F embedding $\phi(u_t)$. We then score how well $D$ follows the reference workflow as the geometric mean of its transition probabilities under $G$:
\[
\mathcal{L}(D) = \Big(\prod_{t \in \mathcal{K}_D} \hat{P}_G(n_t \mid n_{t-1})\Big)^{1/|\mathcal{K}_D|},
\]
where $\mathcal{K}_D$ is the set of transitions of $D$ observed in $G$ and $\hat{P}_G$ the Laplace-smoothed transition probability.
\item \textbf{Distribution distance.} Let $p_{\mathcal{D}}$ be the kernel density estimate of $\{\mathcal{L}(D) : D \in \mathcal{D}\}$ for a set $\mathcal{D}$. \flowkl is finally the KL divergence between the reference and target sets:
\[
\text{\flowkl}(\mathcal{D}_{\mathrm{ref}}, \mathcal{D}_{\mathrm{tgt}}) = \mathrm{KL}\big(p_{\mathcal{D}_{\mathrm{ref}}} \,\|\, p_{\mathcal{D}_{\mathrm{tgt}}}\big),
\]
with $\mathcal{D}_{\mathrm{ref}} = \mathcal{D}^+$ and $\mathcal{D}_{\mathrm{tgt}} = \mathcal{D}^h_k$.
\end{enumerate}
Unlike FBD and PRD, which compare opaque embedding distributions, \flowkl relies on an explicit, human-readable model of what conformant dialogues look like, the graph $G$, and measures how differently the two sets fit it, \ie how likely their dialogues are to follow its transitions.

\subsection{Task 2: Dialog-to-Dialog Distance}

\begin{table}[t]
\centering
\scriptsize
\setlength{\tabcolsep}{3.7pt}
\caption{Task 2 (Dialog-to-Dialog Distance): Kendall's $\tau$ between predicted and ground-truth fork rankings, averaged across all conformant dialogues. Top block: controls. \textbf{Bold}: best per encoder; \underline{\textbf{bold+underline}}: best overall; shading: darker is better. Mean/Full = mean-turn/full-dialog embedding. \flowscore's improvements over every other method are statistically significant for every encoder (one-sided Wilcoxon, Holm-corrected, $p<0.001$; Table~\ref{tab:d2d_significance}).}
\label{tab:d2d}
\begin{tabular}{llccccc}
\toprule
\textbf{Encoder} & \textbf{Method} & \textbf{\WIC} & \textbf{\SSPC} & \textbf{\GR} & \textbf{\CSC} & \textbf{Avg.} \\
\midrule
\multicolumn{2}{l}{\textcolor{refgray}{Random}} & \textcolor{refgray}{0.000} & \textcolor{refgray}{0.000} & \textcolor{refgray}{0.000} & \textcolor{refgray}{0.000} & \textcolor{refgray}{0.000} \\
\multicolumn{2}{l}{\textcolor{refgray}{Turn-count diff.}} & \textcolor{refgray}{0.025} & \textcolor{refgray}{0.251} & \textcolor{refgray}{0.226} & \textcolor{refgray}{0.002} & \textcolor{refgray}{0.126} \\
\multicolumn{2}{l}{\textcolor{refgray}{Word-count diff.}} & \textcolor{refgray}{0.220} & \textcolor{refgray}{0.242} & \textcolor{refgray}{0.390} & \textcolor{refgray}{0.040} & \textcolor{refgray}{0.223} \\
\multicolumn{2}{l}{\textcolor{refgray}{Shared prefix}} & \textcolor{refgray}{0.688} & \textcolor{refgray}{0.668} & \textcolor{refgray}{0.712} & \textcolor{refgray}{0.662} & \textcolor{refgray}{0.682} \\
\midrule
\multirow{4}{*}{\logo{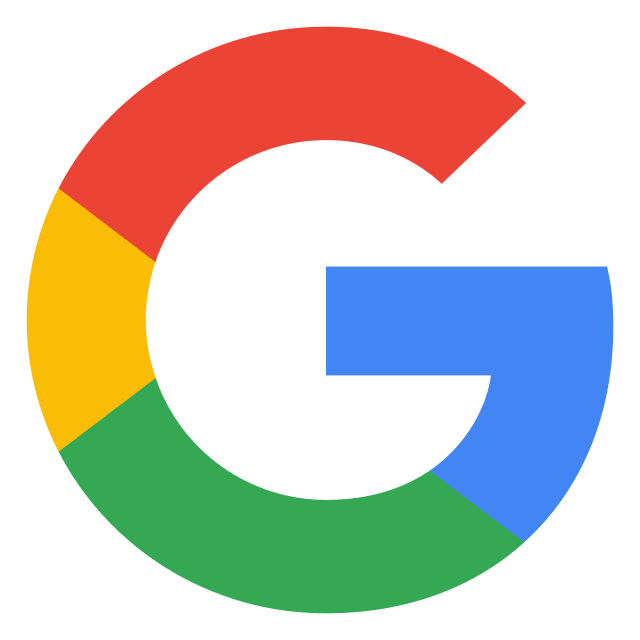}LaBSE} & Mean & \cellcolor{heat!27}0.816 & \cellcolor{heat!22}0.745 & \cellcolor{heat!32}0.895 & \cellcolor{heat!13}0.579 & \cellcolor{heat!23}0.759 \\
 & Full & \cellcolor{heat!23}0.755 & \cellcolor{heat!20}0.705 & \cellcolor{heat!26}0.813 & \cellcolor{heat!7}0.428 & \cellcolor{heat!18}0.675 \\
 & \flowdtw & \cellcolor{heat!28}0.842 & \cellcolor{heat!22}0.738 & \cellcolor{heat!33}0.907 & \cellcolor{heat!9}0.479 & \cellcolor{heat!22}0.742 \\
 & \flowscore & \cellcolor{heat!34}\underline{\textbf{0.925}} & \cellcolor{heat!26}\textbf{0.813} & \cellcolor{heat!35}\textbf{0.941} & \cellcolor{heat!18}\textbf{0.680} & \cellcolor{heat!28}\textbf{0.840} \\
\midrule
\multirow{4}{*}{\logo{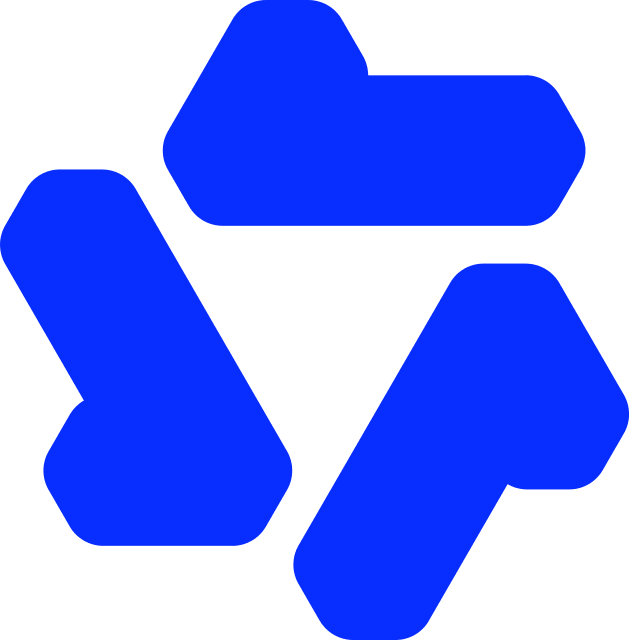}Qwen3-Emb.} & Mean & \cellcolor{heat!25}0.788 & \cellcolor{heat!16}0.631 & \cellcolor{heat!24}0.772 & \cellcolor{heat!6}0.400 & \cellcolor{heat!17}0.648 \\
 & Full & \cellcolor{heat!13}0.580 & \cellcolor{heat!13}0.576 & \cellcolor{heat!17}0.650 & \cellcolor{heat!3}0.270 & \cellcolor{heat!11}0.519 \\
 & \flowdtw & \cellcolor{heat!28}0.844 & \cellcolor{heat!21}0.720 & \cellcolor{heat!31}0.881 & \cellcolor{heat!8}0.440 & \cellcolor{heat!21}0.722 \\
 & \flowscore & \cellcolor{heat!32}\textbf{0.899} & \cellcolor{heat!25}\textbf{0.789} & \cellcolor{heat!34}\textbf{0.917} & \cellcolor{heat!16}\textbf{0.642} & \cellcolor{heat!26}\textbf{0.812} \\
\midrule
\multirow{4}{*}{\logo{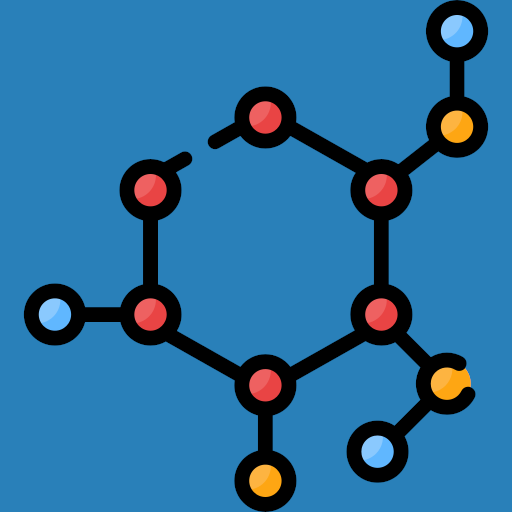}SBERT} & Mean & \cellcolor{heat!26}0.810 & \cellcolor{heat!20}0.714 & \cellcolor{heat!29}0.858 & \cellcolor{heat!11}0.530 & \cellcolor{heat!21}0.728 \\
 & Full & \cellcolor{heat!18}0.671 & \cellcolor{heat!16}0.628 & \cellcolor{heat!22}0.739 & \cellcolor{heat!8}0.453 & \cellcolor{heat!16}0.623 \\
 & \flowdtw & \cellcolor{heat!28}0.843 & \cellcolor{heat!21}0.727 & \cellcolor{heat!32}0.892 & \cellcolor{heat!10}0.492 & \cellcolor{heat!22}0.738 \\
 & \flowscore & \cellcolor{heat!34}\textbf{0.922} & \cellcolor{heat!27}\underline{\textbf{0.815}} & \cellcolor{heat!35}\textbf{0.934} & \cellcolor{heat!20}\underline{\textbf{0.700}} & \cellcolor{heat!28}\textbf{0.843} \\
\midrule
\multirow{4}{*}{\logo{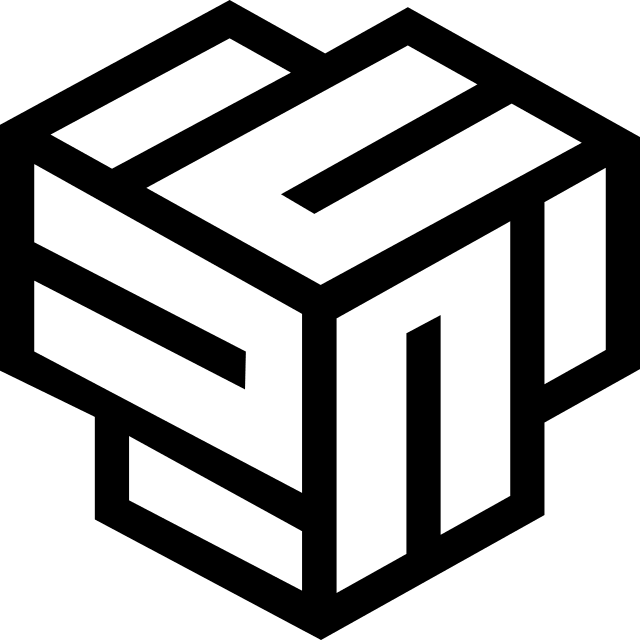}BGE-M3} & Mean & \cellcolor{heat!27}0.822 & \cellcolor{heat!22}0.737 & \cellcolor{heat!31}0.882 & \cellcolor{heat!12}0.550 & \cellcolor{heat!22}0.748 \\
 & Full & \cellcolor{heat!24}0.780 & \cellcolor{heat!20}0.704 & \cellcolor{heat!29}0.856 & \cellcolor{heat!10}0.506 & \cellcolor{heat!20}0.711 \\
 & \flowdtw & \cellcolor{heat!27}0.823 & \cellcolor{heat!21}0.727 & \cellcolor{heat!32}0.891 & \cellcolor{heat!9}0.468 & \cellcolor{heat!21}0.727 \\
 & \flowscore & \cellcolor{heat!34}\textbf{0.924} & \cellcolor{heat!27}\textbf{0.814} & \cellcolor{heat!35}\underline{\textbf{0.942}} & \cellcolor{heat!20}\underline{\textbf{0.700}} & \cellcolor{heat!29}\underline{\textbf{0.845}} \\
\midrule
\multirow{4}{*}{\logo{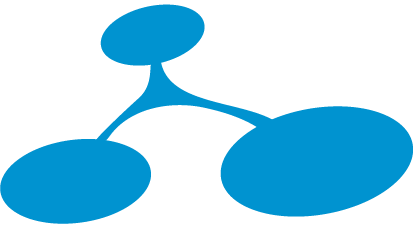}D2F} & Mean & \cellcolor{heat!29}0.848 & \cellcolor{heat!22}0.743 & \cellcolor{heat!32}0.895 & \cellcolor{heat!13}0.561 & \cellcolor{heat!23}0.762 \\
 & Full & \cellcolor{heat!27}0.817 & \cellcolor{heat!21}0.724 & \cellcolor{heat!28}0.843 & \cellcolor{heat!15}0.603 & \cellcolor{heat!22}0.747 \\
 & \flowdtw & \cellcolor{heat!30}0.864 & \cellcolor{heat!20}0.702 & \cellcolor{heat!33}0.913 & \cellcolor{heat!10}0.489 & \cellcolor{heat!22}0.742 \\
 & \flowscore & \cellcolor{heat!34}\underline{\textbf{0.925}} & \cellcolor{heat!25}\textbf{0.798} & \cellcolor{heat!35}\textbf{0.936} & \cellcolor{heat!18}\textbf{0.675} & \cellcolor{heat!28}\textbf{0.834} \\
\midrule
\multirow{4}{*}{\logo{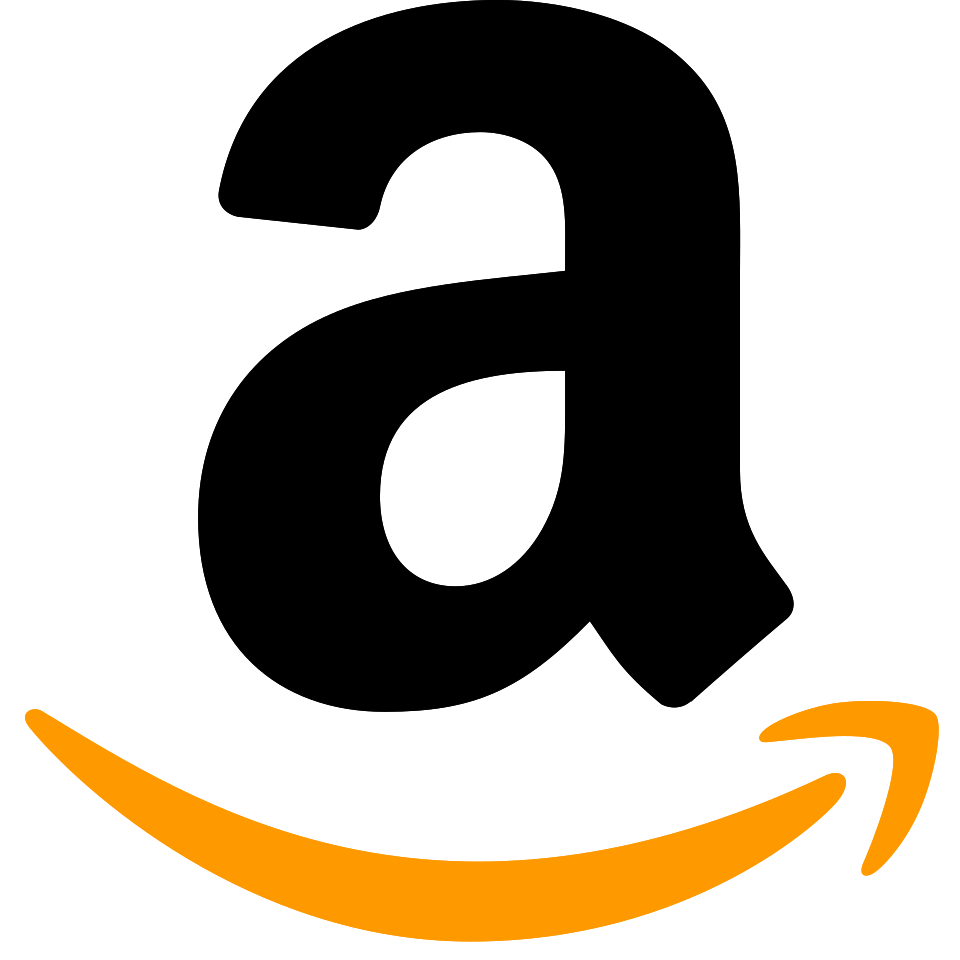}DSE} & Mean & \cellcolor{heat!28}0.832 & \cellcolor{heat!21}0.719 & \cellcolor{heat!31}0.883 & \cellcolor{heat!11}0.531 & \cellcolor{heat!22}0.741 \\
 & Full & \cellcolor{heat!24}0.768 & \cellcolor{heat!21}0.724 & \cellcolor{heat!26}0.801 & \cellcolor{heat!11}0.524 & \cellcolor{heat!20}0.704 \\
 & \flowdtw & \cellcolor{heat!29}0.848 & \cellcolor{heat!19}0.698 & \cellcolor{heat!32}0.897 & \cellcolor{heat!10}0.505 & \cellcolor{heat!22}0.737 \\
 & \flowscore & \cellcolor{heat!34}\textbf{0.922} & \cellcolor{heat!25}\textbf{0.794} & \cellcolor{heat!35}\textbf{0.929} & \cellcolor{heat!19}\textbf{0.685} & \cellcolor{heat!28}\textbf{0.833} \\
\midrule
\multirow{4}{*}{\logo{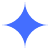}Emb.Gemma} & Mean & \cellcolor{heat!27}0.817 & \cellcolor{heat!21}0.726 & \cellcolor{heat!31}0.876 & \cellcolor{heat!12}0.539 & \cellcolor{heat!22}0.739 \\
 & Full & \cellcolor{heat!21}0.725 & \cellcolor{heat!18}0.670 & \cellcolor{heat!29}0.845 & \cellcolor{heat!9}0.462 & \cellcolor{heat!18}0.676 \\
 & \flowdtw & \cellcolor{heat!28}0.833 & \cellcolor{heat!20}0.701 & \cellcolor{heat!31}0.886 & \cellcolor{heat!9}0.463 & \cellcolor{heat!21}0.721 \\
 & \flowscore & \cellcolor{heat!33}\textbf{0.911} & \cellcolor{heat!26}\textbf{0.801} & \cellcolor{heat!35}\textbf{0.935} & \cellcolor{heat!19}\textbf{0.692} & \cellcolor{heat!28}\textbf{0.835} \\
\bottomrule
\end{tabular}
\end{table}
Task 2 probes procedural (dialog-flow) similarity at the level of individual dialogues, as needed, for instance, to retrieve past conversations that unfolded like a given one (\eg to find how similar cases were handled). \emph{Input:} two dialogues; \emph{output:} a scalar distance between them. To evaluate it with \dialfork, for each conformant dialogue $D^+$, the four forked dialogues $D^h_k$ provide an unambiguous 4-item ranking of distance from $D^+$ (Figure~\ref{fig:overview}), and a good dialogue-distance metric should recover this ranking. We report Kendall's $\tau$ between predicted and ground-truth rankings, averaged across all conformant dialogues. As, to our knowledge, no metric has been designed for dialogue-to-dialogue distance on raw dialogue text, we compare standard pooling baselines with two new flow-aware methods. All methods below define a distance between two dialogues $D$ and $D'$ (in our evaluation, $D = D^+$ and $D' = D^h_k$), with per-turn embedding sequences $E=(e_1,\ldots,e_n)$ and $E'=(e'_1,\ldots,e'_m)$.

\paragraph{Pooling baselines.} \textbf{Mean-turn}: $\text{Mean}(D, D') = 1 - \cos(\bar{e}, \bar{e}')$, where $\bar{e}$ and $\bar{e}'$ average the per-turn embeddings. \textbf{Full-dialog}: $\text{Full}(D, D') = 1 - \cos(\phi(D), \phi(D'))$, where $\phi$ encodes the concatenated transcript once.

\noindent
\textbf{Controls.} \textbf{Turn-/Word-count diff.} use $|\mathrm{len}(D) - \mathrm{len}(D')|$ (in turns or words) as distance; \textbf{Random} has expected $\tau = 0$. \textbf{Shared prefix} uses the number of opening turns $D$ and $D'$ share verbatim. This control exploits the construction mechanism itself and has no counterpart in real dialogues, which do not come with a conformant twin; we report it only to verify that our metrics do not reduce to it.

\paragraph{\flowdtw.} We align $E$ and $E'$ using Dynamic Time Warping (DTW)~\citep{sakoe-chiba-1978-dtw}. Local costs are $c(i,j) = 1 - e_i^\top e'_j$ and the DTW recurrence $\Gamma(i,j) = c(i,j) + \min\{\Gamma(i{-}1,j), \Gamma(i,j{-}1), \Gamma(i{-}1,j{-}1)\}$ with $\Gamma(0,0)=0$ gives the optimal aligned cost, normalized by the length $L$ of the optimal warping path: $\text{\flowdtw}(D, D') = \Gamma(n,m)/L$.

\paragraph{\flowscore.} A BERTScore-inspired~\citep{zhang-etal-2020-bertscore} matching applied at the turn level. We build $S(i,j) = e_i^\top e'_j$ and compute
\[
P = \tfrac{1}{n}\sum\limits_i \max\limits_j S(i,j),\; R = \tfrac{1}{m}\sum\limits_j \max\limits_i S(i,j),
\]
and $F_{\text{flow}} = 2PR/(P+R)$, and define the distance
\[
\text{\flowscore}(D, D') = 1 - F_{\text{flow}}
\]
\flowscore captures soft turn-level alignment without requiring the strict monotonic warping of DTW. Like BERTScore, \flowscore is meant to operate on contextualized embeddings, which encode each element in its context and thus make the matching sensitive to order. To our knowledge, however, no contextualized \emph{turn} encoder is publicly available: with the sentence encoders used here, each turn is embedded independently, so our instantiation of \flowscore does not yet capture turn order, a natural direction for future work.

\subsection{Task 3: Hallucination Turn Prediction}
\label{subsec:task3}
\begin{table*}[t]
\centering
\scriptsize
\setlength{\tabcolsep}{4pt}
\caption{Task 3 (Turn Prediction): results on the Task~3 test set; MAEs are in turns. \textbf{Bold}: best within a regime; \underline{\textbf{bold+underline}}: best overall; shading: darker is better.}
\label{tab:tp_main}
\begin{tabular}{lccccccccc}
\toprule
& \multicolumn{4}{c}{\textbf{Hallucination MAE} $\downarrow$} & \multicolumn{2}{c}{\textbf{Hallucination}} & \multicolumn{2}{c}{\textbf{Conformance}} & \\
\cmidrule(lr){2-5}\cmidrule(lr){6-7}\cmidrule(lr){8-9}
\textbf{Method} & \textbf{\CSC} & \textbf{\GR} & \textbf{\SSPC} & \textbf{\WIC} & \textbf{Avg. MAE\ $\downarrow$} & \textbf{Cov.\ $\uparrow$} & \textbf{MAE $\downarrow$} & \textbf{Cov.\ $\downarrow$} & \textbf{Score $\uparrow$} \\
\midrule
\multicolumn{10}{l}{\textbf{\textit{Baselines}}} \\
\midrule
\textcolor{refgray}{Random} & \textcolor{refgray}{6.31} & \textcolor{refgray}{7.92} & \textcolor{refgray}{5.74} & \textcolor{refgray}{6.44} & \textcolor{refgray}{6.60} & \textcolor{refgray}{100.00} & \textcolor{refgray}{7.50} & \textcolor{refgray}{100.00} & \textcolor{refgray}{0.94} \\
\textcolor{refgray}{First turn} & \textcolor{refgray}{5.64} & \textcolor{refgray}{5.64} & \textcolor{refgray}{5.64} & \textcolor{refgray}{5.64} & \textcolor{refgray}{5.64} & \textcolor{refgray}{100.00} & \textcolor{refgray}{14.12} & \textcolor{refgray}{100.00} & \textcolor{refgray}{0.00} \\
\textcolor{refgray}{Last turn} & \textcolor{refgray}{12.01} & \textcolor{refgray}{15.57} & \textcolor{refgray}{10.26} & \textcolor{refgray}{12.22} & \textcolor{refgray}{12.52} & \textcolor{refgray}{100.00} & \textcolor{refgray}{1.00} & \textcolor{refgray}{100.00} & \textcolor{refgray}{0.00} \\
\midrule
\multicolumn{10}{l}{\textbf{\textit{Streaming}}} \\
\midrule
Surprisal Spike (\SS) & \cellcolor{heat!36}4.21 & \cellcolor{heat!34}1.56 & \cellcolor{heat!39}1.30 & \cellcolor{heat!35}1.68 & \cellcolor{heat!37}2.19 & \cellcolor{heat!34}93.24 & \cellcolor{heat!15}\textbf{6.33} & \cellcolor{heat!35}\textbf{12.02} & \cellcolor{heat!31}\textbf{73.69} \\
\quad w/ \NC & \cellcolor{heat!37}4.13 & \cellcolor{heat!33}1.61 & \cellcolor{heat!38}1.33 & \cellcolor{heat!34}1.71 & \cellcolor{heat!37}2.20 & \cellcolor{heat!35}93.56 & \cellcolor{heat!12}6.58 & \cellcolor{heat!34}15.47 & \cellcolor{heat!27}72.45 \\
Improbable Transition (\IT) & \cellcolor{heat!37}4.13 & \cellcolor{heat!38}0.93 & \cellcolor{heat!39}\textbf{1.25} & \cellcolor{heat!35}1.63 & \cellcolor{heat!39}1.98 & \cellcolor{heat!35}93.86 & \cellcolor{heat!13}6.52 & \cellcolor{heat!35}12.49 & \cellcolor{heat!31}73.60 \\
\quad w/ \NC & \cellcolor{heat!38}\textbf{3.94} & \cellcolor{heat!38}\textbf{0.92} & \cellcolor{heat!39}1.29 & \cellcolor{heat!36}\textbf{1.60} & \cellcolor{heat!39}\textbf{1.94} & \cellcolor{heat!37}\textbf{94.80} & \cellcolor{heat!12}6.55 & \cellcolor{heat!33}16.31 & \cellcolor{heat!30}73.23 \\
\midrule
\multicolumn{10}{l}{\textbf{\textit{Non-streaming}}} \\
\midrule
\logo{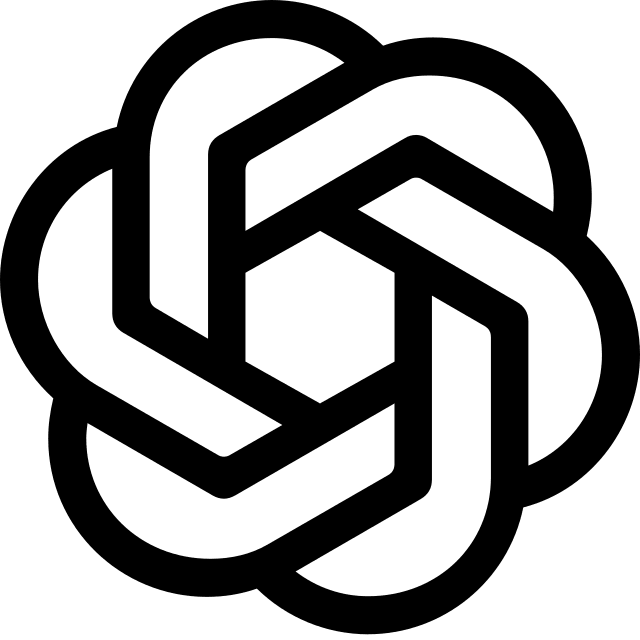}\llm{GPT-OSS-20B} & 10.10 & \cellcolor{heat!3}6.70 & 4.77 & 4.34 & 6.48 & 72.38 & \cellcolor{heat!40}\underline{\textbf{4.40}} & \cellcolor{heat!31}20.04 & \cellcolor{heat!1}62.65 \\
\logo{qwen}\llm{Qwen3.6-35B} & \cellcolor{heat!10}8.45 & \cellcolor{heat!4}6.46 & \cellcolor{heat!17}3.28 & \cellcolor{heat!1}4.28 & \cellcolor{heat!7}5.62 & \cellcolor{heat!17}\textbf{82.94} & \cellcolor{heat!34}4.86 & \cellcolor{heat!26}29.92 & \cellcolor{heat!6}64.56 \\
\logo{gemma}\llm{Gemma4-31B} & 10.13 & 7.12 & \cellcolor{heat!28}2.26 & \cellcolor{heat!6}3.88 & \cellcolor{heat!5}5.85 & \cellcolor{heat!2}73.33 & \cellcolor{heat!37}4.63 & \cellcolor{heat!37}\underline{\textbf{9.79}} & \cellcolor{heat!9}65.62 \\
\logo{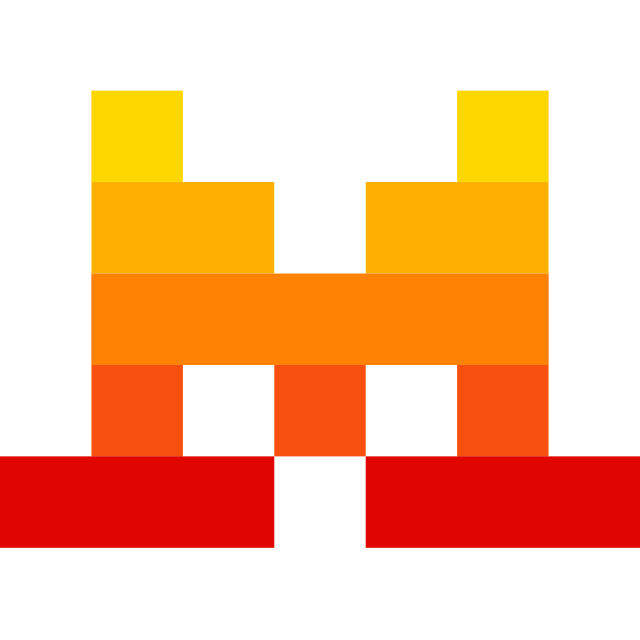}\llm{Mistral-Small-3.2-24B} & \cellcolor{heat!10}8.51 & \cellcolor{heat!9}5.56 & \cellcolor{heat!21}2.88 & \cellcolor{heat!5}3.98 & \cellcolor{heat!11}5.23 & \cellcolor{heat!10}78.20 & \cellcolor{heat!37}4.60 & \cellcolor{heat!35}12.12 & \cellcolor{heat!16}\textbf{68.14} \\
\logo{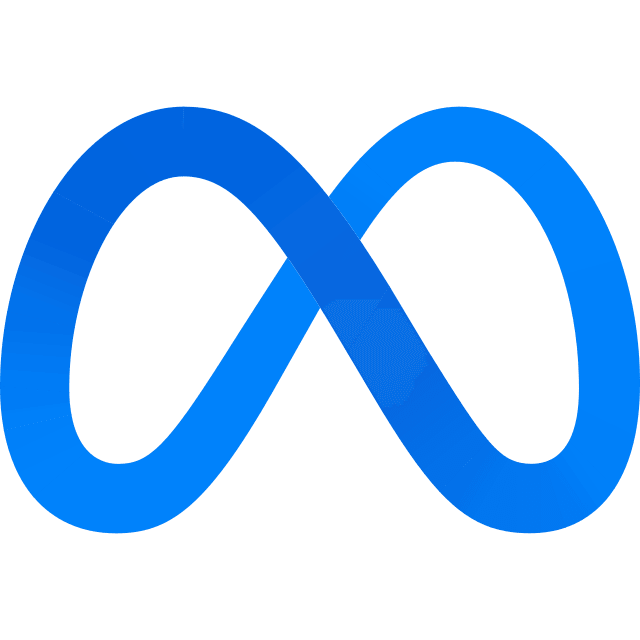}\llm{Llama-3.3-70B} & \cellcolor{heat!12}\textbf{8.10} & \cellcolor{heat!13}\textbf{5.00} & \cellcolor{heat!31}\textbf{2.04} & \cellcolor{heat!19}\textbf{2.85} & \cellcolor{heat!17}\textbf{4.50} & \cellcolor{heat!13}80.62 & \cellcolor{heat!23}5.69 & \cellcolor{heat!32}18.36 & \cellcolor{heat!13}67.09 \\
\midrule
\multicolumn{10}{l}{\textbf{\textit{Non-streaming + workflow}}} \\
\midrule
\logo{openai}\llm{GPT-OSS-20B} & \cellcolor{heat!30}5.18 & \cellcolor{heat!35}1.37 & \cellcolor{heat!38}1.40 & \cellcolor{heat!25}2.45 & \cellcolor{heat!33}2.60 & \cellcolor{heat!40}\underline{\textbf{96.88}} & \cellcolor{heat!39}4.45 & 81.64 & 62.31 \\
\logo{qwen}\llm{Qwen3.6-35B} & \cellcolor{heat!32}4.96 & \cellcolor{heat!36}1.14 & \cellcolor{heat!38}1.36 & \cellcolor{heat!33}1.85 & \cellcolor{heat!36}2.33 & \cellcolor{heat!40}96.71 & \cellcolor{heat!38}4.57 & \cellcolor{heat!14}53.96 & \cellcolor{heat!26}71.87 \\
\logo{gemma}\llm{Gemma4-31B} & \cellcolor{heat!30}5.30 & \cellcolor{heat!40}\underline{\textbf{0.51}} & \cellcolor{heat!39}1.31 & \cellcolor{heat!40}\underline{\textbf{1.28}} & \cellcolor{heat!38}2.10 & \cellcolor{heat!34}93.03 & \cellcolor{heat!40}\textbf{4.41} & \cellcolor{heat!26}30.48 & \cellcolor{heat!40}77.06 \\
\logo{mistral}\llm{Mistral-Small-3.2-24B} & \cellcolor{heat!31}5.03 & \cellcolor{heat!35}1.36 & \cellcolor{heat!34}1.74 & \cellcolor{heat!29}2.14 & \cellcolor{heat!34}2.57 & \cellcolor{heat!34}92.91 & \cellcolor{heat!33}4.94 & \cellcolor{heat!33}\textbf{16.68} & \cellcolor{heat!38}76.38 \\
\logo{meta}\llm{Llama-3.3-70B} & \cellcolor{heat!39}\underline{\textbf{3.83}} & \cellcolor{heat!39}0.69 & \cellcolor{heat!40}\underline{\textbf{1.19}} & \cellcolor{heat!36}1.61 & \cellcolor{heat!40}\underline{\textbf{1.83}} & \cellcolor{heat!35}93.89 & \cellcolor{heat!30}5.18 & \cellcolor{heat!31}21.06 & \cellcolor{heat!40}\underline{\textbf{77.09}} \\
\bottomrule
\end{tabular}
\end{table*}

Task 3 probes whether a method can localize \emph{where} a dialogue first goes off-track, as needed, for instance, by a guardrail that monitors a live conversation and intervenes as soon as the agent deviates from its intended process. The output is the first turn flagged as off-track, given as its position in the dialogue (counting the turns of both speakers), or none if the dialogue is conformant; for forked dialogues, the goal is to locate the hallucination \emph{onset}, the fork turn $t_k$, as closely as possible. We consider three regimes of decreasing realism: 
\begin{enumerate}[nosep, leftmargin=*]
\item \textbf{Streaming}: the dialogue's turns arrive one at a time, and the method must decide after each turn whether the conversation has gone off-track, with no access to future turns, and with no workflow. The most realistic and hardest setting.
\item \textbf{Non-streaming}: the complete dialogue, with no workflow.
\item \textbf{Non-streaming + workflow}: the complete dialogue \emph{and} its ground-truth workflow, typically unavailable in practice; reported as a reference with privileged access to the workflow.
\end{enumerate}
In Tasks~1 and~2, conformant dialogues are the reference every method compares against; in Task~3, they are also scored, for false alarms. A method that learns from conformant dialogues (\eg the workflow) must therefore not be scored on the same ones: \dialfork defines Task~3 train and test sets: methods learn from the train dialogues and are scored on the test dialogues (Appendix~\ref{app:task3_split}).

\paragraph{Streaming.} \emph{First}, for each dialogue we map turn embeddings to discrete nodes via the encoder's clustering (as in step~1 of \flowkl, \S\ref{subsec:task1}) and fit an $N$-gram language model over the node sequences of the train dialogues. Let $p_t = P_{\mathrm{LM}}(n_t \mid n_{t-N+1},\ldots,n_{t-1})$. Because the node assignment may be uncertain when the turn embedding $\phi(u_t)$ lies between clusters, we also consider weighting each transition by a \emph{node confidence} (\NC), $c_t = P_{\NC}(n_t \mid u_t) = e^{\sigma_1} / \sum_{j=1}^{K} e^{\sigma_j}$, where $\sigma_j$ is the cosine similarity between $\phi(u_t)$ and the $j$-th nearest centroid. Since the encoders we use are trained with contrastive InfoNCE-style objectives~\citep{oord2018representation}, this softmax approximates the posterior that $u_t$ belongs to $n_t$ (\ie that $n_t$'s centroid is the \emph{positive} for $u_t$) given the $K{-}1$ neighbors as \emph{negatives}. The \NC-weighted transition probability is $p_t^{\NC} = p_t \cdot c_t$. \emph{Then}, on top of this LM, we define two detectors. \textbf{Improbable Transition (\IT)}: predict the first $t$ with $p_t < \theta_{\mathrm{off}}$ (or $p_t^{\NC} < \theta_{\mathrm{off}}$ in the \NC variant); $\theta_{\mathrm{off}}$ is the 5th percentile of $\min_t p_t$ over the same conformant dialogues ($\approx 5\%$ false alarms on them). \textbf{Surprisal Spike (\SS)}: let $s_t = -\log p_t$ and $\Delta_t = s_t - s_{t-1}$; predict the first $t$ with $\Delta_t > \theta_{\mathrm{jump}}$, calibrated as the 95th percentile of $\max_t \Delta_t$ over the same dialogues (\NC: $s_t = -\log p_t^{\NC}$).

\paragraph{Non-streaming.} In the non-streaming regimes we use zero-shot LLM prompting: the LLM is asked to identify the system turn where the dialogue first deviates from the intended workflow (or to indicate that no deviation occurs). Without workflow access, the prompt includes only the dialogue; with workflow access, the ground-truth workflow is also provided. The exact prompts are in Appendix~\ref{app:llm_prompts}.

\noindent
\textbf{Baselines.} Three trivial predictors always predict the \textit{first}, the \textit{last}, or a \textit{random} turn.

\noindent
\textbf{Metrics.} For hallucinated dialogues, we report the mean absolute error (MAE) of the predicted turn against the fork turn and the \emph{hallucinated coverage}: the fraction of forked dialogues flagged. For conformant dialogues, we report the MAE relative to the dialogue end and the \emph{conformant coverage} (the false-alarm rate, to be minimized). A composite Score combines these, weighting MAE (0.35 each) above coverage (0.15 each), since localization is the task's defining output and coverage mainly guards against always- or never-flag strategies; per-regime winners are robust to the weights (Appendix~\ref{app:score}).

\section{Experiments}
\label{sec:experiments}

\subsection{Task 1: Set-to-Set Distance}

Table~\ref{tab:s2s} reports set-to-set ranking results. Despite relying on different principles (embedding distributions vs.\ an explicit flow graph), all three distance metrics recover the expected ordering with high agreement ($\tau \geq 0.88$ on average), while length-only controls do not. This provides the sanity check announced in \S\ref{subsec:task1}: forks depart from the conformant dialogues at their designated positions, not merely by growing longer.

The differences among the three metrics are not statistically significant (see Appendix~\ref{app:stats} for the testing protocol): \flowkl performs on par with FBD and PRD while relying on an interpretable human-readable transition graph. The metrics do differ in which hallucination types they capture best. \flowkl is the best metric on \SSPC, the type that targets step ordering, suggesting that an explicit transition model captures step order better than implicit embedding distributions. 

\subsection{Task 2: Dialog-to-Dialog Distance}

Table~\ref{tab:d2d} reports the dialog-to-dialog ranking results across seven encoder families, covering general-purpose sentence embeddings (SBERT~\citep{reimers-gurevych-2019-sbert}, LaBSE~\citep{feng-etal-2022-labse}, BGE-M3~\citep{chen-etal-2024-bgem3}, Qwen3-Embedding~\citep{zhang2025qwen3embedding}, EmbeddingGemma~\citep{vera-etal-2025-embeddinggemma}) and dialogue-specific embeddings (DSE~\citep{zhou-etal-2022-dse} and D2F~\citep{burdisso-etal-2024-dialog2flow}). For each family we report the best-performing variant (checkpoints in Appendix~\ref{app:encoders}; per-task results and significance tests in Appendix~\ref{app:d2d_significance}).

With the baseline methods, the choice of encoder matters: D2F is the best (or tied-best) encoder for Mean, Full, and \flowdtw. \flowscore, in contrast, is the strongest method for all seven encoders, with statistically significant gains over every other method (Table~\ref{tab:d2d_significance}), and lifts all of them above $\tau{=}0.81$, making results largely independent of the encoder. This supports \emph{semantic} and \emph{procedural} similarity being distinct axes: two dialogues can be close in content yet differ in how they unfold. For every encoder, comparing whole-dialogue embeddings (Full) ranks forks markedly worse ($\tau{=}0.519$--$0.747$), since a fork and its source largely share their content and thus remain semantically close, whereas matching the same turn embeddings with \flowscore recovers the ordering ($0.812$--$0.845$). The gap is largest for Qwen3-Embedding, a strong semantic encoder ($0.519$ vs.\ $0.812$), and smallest for D2F ($0.747$ vs.\ $0.834$), the encoder trained to capture dialog flow. 

None of the controls explains these results. Dialogue length alone is close to chance (Turn-count diff.: $\tau{=}0.126$). Shared prefix, which reads the fork position off the construction itself, reaches $\tau{=}0.682$; it is not a perfect oracle because replayed prefixes may contain minor rewordings (\S\ref{sec:dataset}). Even so, \flowscore outperforms it for every encoder ($p<0.001$), without this construction cue.

\subsection{Task 3: Hallucination Turn Prediction}

\textbf{Streaming.} We report a 3-gram language model with LaBSE, the encoder with the best Score (other encoders and $N$: Appendix~\ref{app:turn_streaming}). Operating turn by turn and learning only from the train dialogues, the streaming detectors localize the fork within $1.9$--$2.2$ turns, flagging 93--95\% of forks at a false-alarm rate of 12--16\%. \IT localizes better than \SS; \NC weighting further sharpens \IT (MAE $1.94$, the lowest without the ground-truth workflow) and raises coverage, at the cost of more false alarms; plain \SS and \IT score best ($73.69$, $73.60$). \textbf{Non-streaming.} We prompt five open-weight LLMs (Appendix~\ref{app:llm_prompts}). Without the workflow, the zero-shot LLMs localize deviations far less precisely than the streaming detectors (MAE $4.50$--$6.48$ vs.\ $1.94$) and flag only 72--83\% of forks, despite seeing the whole dialogue. Given the ground-truth workflow, they localize about as well as the streaming detectors, which can only learn the workflow from past conformant dialogues (here, simulated ones); \llm{Llama-3.3-70B} localizes best (MAE $1.83$) and reaches the best overall Score ($77.09$, tied with \llm{Gemma4-31B}), but the LLMs raise more false alarms (17--82\% vs.\ 12--16\%). These gains, however, require the complete dialogue (deviations surface only after the fact), ideally the workflow, and a second LLM supervising the first. The streaming detectors, in contrast, are lightweight, need no workflow specification, and flag deviations as they happen, which makes them a promising basis for real-time guardrails. Still, they are simple first baselines, and no method we evaluate exceeds a Score of 78 out of 100: the best ones still miss the fork by about two turns and raise false alarms on 12--30\% of conformant dialogues, so localizing procedural hallucination as it happens remains an open problem. By marking where each dialogue first goes off-track, \dialfork offers the community a testbed to tackle it.


\section{Conclusions}
\label{sec:conclusion}

LLMs deployed in agentic conversational roles can produce individually plausible turns while violating the procedural logic they were meant to follow. We named this failure mode \emph{procedural hallucination} and introduced \dialfork, the first benchmark that makes it systematically measurable from raw dialogue text, with a construction validated by human judgment, surfacing an axis of conversational quality that existing tools largely miss: \emph{procedural} similarity, distinct from \emph{semantic} similarity. Its three tasks open the way to distributional monitoring of agents, \emph{procedural retrieval}, and real-time guardrails. More broadly, we envision procedural fidelity as a first-class property of conversational agents: monitored in deployment, searchable in conversation logs, and ultimately optimized in training, so agents stay on process by design. We release the dataset under CC BY 4.0 and the code under the MIT license.

\section*{Acknowledgements}

This work was supported by EU Horizon 2020 project ELOQUENCE\footnote{\url{https://eloquenceai.eu/}} (grant number 101135916). We are grateful to the 30 volunteers who took part in the human study, including those who agreed to be named: Ravinithesh Annapureddy, Nikos Arvanitis, Miranda Carlsson, Miguel Claramunt, Nawal Haidar, Olena Hrynenko, Barbara Huguenin, Marc Juvillà, Driss Khalil, Alain Komaty, Marco Matassoni, Diego Moreno, Thibault Roux, Aleix Sant, Emmanuel Senft, Maryam Soltani, Adam Unal, Elija Vida, Vidit Vidit, Hengfei Wang, and Hasindri Watawana. 

\section*{Limitations}
\label{sec:limitations}

\paragraph{Single source corpus.} \dialfork is built on \STAR, which spans 24 tasks across 12 domains. We chose \STAR because, to our knowledge, it is the only public corpus that releases both an explicit workflow for each task and the role-play instructions given to the humans who produced each dialogue, which our pipeline needs to simulate conformant dialogues and to define what counts as a deviation; and since all \dialfork dialogues are newly generated, only \STAR's metadata, not its original dialogues, enters the benchmark. Still, the empirical results we report are tied to \STAR's distribution of tasks and domains. The construction pipeline and the formal model make no \STAR-specific assumptions and apply to any corpus with workflow-level metadata; extending \dialfork to other corpora (\eg non-English dialogues, or workflows and domains beyond \STAR's) as they become available is left to future work and to users of the released generation pipeline.

\paragraph{Single-type hallucinations.} Each forked dialogue injects exactly one hallucination type at one fork point. Real conversational failures can be compound (multiple types co-occurring) or cascading (one corruption inducing another over subsequent turns). The detectors we propose are not evaluated under such combined conditions, and the four-type taxonomy may need extension to cover them.

\paragraph{Simulated rather than observed dialogues.} Both conformant and forked dialogues are generated by LLM role-play with a single generator (\llm{Gemma4-31B}) under controlled prompting. The generator only realizes the surface language: the labels come from the construction (seeded replay fixes the prefix, the fork turn, and the fork order), the tasks take only the text as input, and the human study confirms that the generator carries out the injected behavior at the intended turn. Other generators may still produce different language and conversational styles; the released pipeline supports regenerating \dialfork with any LLM. This gives us scale, reproducibility, and exact ground-truth fork points, but the resulting distribution of language and conversational behavior may differ from dialogues collected from real users and deployed agents. Our human study (Appendix~\ref{app:human}) shows that the constructed deviations are perceived by people as intended, on five tasks spanning short to long dialogues and different domains; still, methods that work well on \dialfork may behave differently on naturally-occurring conversations, and conversely.

\paragraph{Baseline scope.} The methods we report are intended as initial baselines, not as the strongest possible detectors. The streaming detectors use simple $N$-gram models over discrete nodes with no neural training beyond the pretrained encoder; the non-streaming detectors use zero-shot prompting of five open-weight LLMs without fine-tuning, few-shot examples, or chain-of-thought elicitation. Substantially stronger detectors are likely possible on all three tasks, and we hope users of \dialfork will explore them.
One family is excluded by design: detectors relying on signals internal to the generator (\eg per-turn likelihood or confidence), which would tie results to one model and do not exist for dialogues produced by other systems or by humans, whereas \dialfork evaluates methods on text alone.

\section{Ethical Considerations}



\dialfork is constructed entirely from synthetically generated dialogues; no real user data or personal information is involved. The human study that validates its construction (Appendix~\ref{app:human}) was carried out by volunteers, participating freely and without financial compensation, who judged only these synthetic dialogues: each participant read and accepted a consent statement before starting, could stop at any time, and chose whether to be acknowledged by name or to remain anonymous; the released judgments identify participants only by opaque identifiers. Two further points warrant note.
Since the dialogues are produced by an LLM, they may reflect biases present in the generator, and the 24 service-oriented tasks are not necessarily a representative sample of real-world conversational settings; results on \dialfork should be read as evidence about procedural deviation under controlled conditions rather than as direct estimates of deployed-system behavior.
In particular, detectors validated on \dialfork should not be deployed as guardrails in high-stakes settings (\eg clinical intake) without domain-specific validation, where missed deviations or excessive false alarms carry real costs.
We judge the value of an openly available, reproducible benchmark to outweigh this risk, particularly because the same machinery enables the detection, guardrail, and monitoring directions the work is intended to support. The \dialfork dataset is released under CC BY 4.0 and all accompanying code under the MIT license; their intended use is research on conversational reliability, consistent with these licenses.

\paragraph{Use of AI assistants.} We used Claude (Anthropic) to assist with writing and editing the paper and helping us with development of the annotation tool used in the human study (Appendix~\ref{app:human}). All AI-assisted text and code were reviewed, edited, and verified by the authors, who take full responsibility for the content.

\bibliography{dialfork}


\appendix
\newpage
\section{Detailed Per-Task Dataset Statistics}
\label{app:per_task_stats}

Table~\ref{tab:forkbench_composition} reports the overall composition of \dialfork (conformant dialogues and the forks of each hallucination type), and Table~\ref{tab:forkbench_task_details} a detailed breakdown for each of the 24 tasks; Figure~\ref{fig:task_stats} summarizes both visually. Coverage varies substantially across tasks, reflecting both the heterogeneous yield of valid conformant generations and the variable workflow complexity inherited from \STAR; per-task dialogue lengths span more than $2\times$ (from 12.41 turns for \textit{ride\_status} to 29.28 for \textit{trivia}). Forked dialogues are consistently longer than conformant ones across nearly all tasks, with the effect most pronounced where the underlying workflow offers more steps for the hallucination to dilate (\eg \textit{trivia}: 19.45 conformant vs.\ 29.90 forked mean turns).

\begin{table}[t]
\centering
\small
\setlength{\tabcolsep}{4pt}
\caption{Dataset composition: conformant dialogues and forks of each hallucination type, with their number, share of the dataset, and average length in turns (\#Turns) and words (\#Words).}
\label{tab:forkbench_composition}
\begin{tabular}{lrrcc}
\toprule
 & \multicolumn{2}{c}{\textbf{Coverage}} & \multicolumn{2}{c}{\textbf{Dialog Length}} \\
\cmidrule(lr){2-3}\cmidrule(lr){4-5}
\textbf{Dialogues} & \textbf{\#Dialogs} & \textbf{\%Total} & \textbf{\#Turns} & \textbf{\#Words} \\
\midrule
Conformant   & 2,134  & 5.88\%   & 15.52 & 186.94 \\
\WIC forked  & 8,536  & 23.53\%  & 20.02 & 258.49 \\
\SSPC forked & 8,536  & 23.53\%  & 19.31 & 187.47 \\
\GR forked   & 8,536  & 23.53\%  & 24.79 & 342.23 \\
\CSC forked  & 8,536  & 23.53\%  & 20.58 & 247.64 \\
\midrule
\textbf{Total} & \textbf{36,278} & \textbf{100.00\%} & \textbf{20.84} & \textbf{254.72} \\
\bottomrule
\end{tabular}
\end{table}

\begin{table*}[t]
\centering
\small
\setlength{\tabcolsep}{4pt}
\caption{Per-task detailed statistics. Task labels are shown as task\_name (abbr). In each block, \# = number of dialogues, MT = mean turns, MW = mean words. Conformant = non-hallucinated; Forks = aggregated forked variants; Total combines both. Task~3 split: number of dialogues in the Task~3 train and test sets (\S\ref{app:task3_split}).}
\label{tab:forkbench_task_details}
\begin{tabular}{lcccccccccrr}
\toprule
 & \multicolumn{3}{c}{\textbf{Conformant}} & \multicolumn{3}{c}{\textbf{Forks}} & \multicolumn{3}{c}{\textbf{Total}} & \multicolumn{2}{c}{\textbf{Task~3 split}} \\
\cmidrule(lr){2-4}\cmidrule(lr){5-7}\cmidrule(lr){8-10}\cmidrule(lr){11-12}
\textbf{Task} & \textbf{\#} & \textbf{MT} & \textbf{MW} & \textbf{\#} & \textbf{MT} & \textbf{MW} & \textbf{\#} & \textbf{MT} & \textbf{MW} & \textbf{Train} & \textbf{Test} \\
\midrule
apartment\_schedule (ase) & 206 & 18.65 & 235.68 & 3,296 & 22.36 & 294.07 & 3,502 & 22.15 & 290.64 & 103 & 1,751 \\
apartment\_search (as) & 30 & 11.87 & 271.57 & 480 & 17.07 & 300.99 & 510 & 16.77 & 299.26 & 15 & 255 \\
bank\_balance (bb) & 46 & 13.52 & 125.39 & 736 & 18.78 & 214.28 & 782 & 18.47 & 209.05 & 23 & 391 \\
bank\_fraud\_report (bfr) & 135 & 13.73 & 166.71 & 2,160 & 19.90 & 232.26 & 2,295 & 19.54 & 228.40 & 67 & 1,156 \\
doctor\_followup (df) & 42 & 9.38 & 110.36 & 672 & 13.85 & 173.54 & 714 & 13.59 & 169.83 & 21 & 357 \\
doctor\_schedule (ds) & 125 & 17.99 & 181.80 & 2,000 & 21.47 & 229.06 & 2,125 & 21.27 & 226.28 & 62 & 1,071 \\
hotel\_book (hb) & 122 & 16.77 & 175.32 & 1,952 & 18.18 & 200.28 & 2,074 & 18.10 & 198.81 & 61 & 1,037 \\
hotel\_search (hs) & 83 & 7.43 & 115.58 & 1,328 & 14.40 & 206.31 & 1,411 & 13.99 & 200.97 & 41 & 714 \\
hotel\_service\_request (hsr) & 46 & 15.00 & 153.72 & 736 & 20.86 & 240.71 & 782 & 20.51 & 235.59 & 23 & 391 \\
meeting\_schedule (ms) & 41 & 16.76 & 169.71 & 656 & 20.46 & 239.84 & 697 & 20.24 & 235.72 & 20 & 357 \\
trip\_directions (td) & 208 & 15.78 & 229.75 & 3,328 & 27.46 & 355.91 & 3,536 & 26.78 & 348.49 & 104 & 1,768 \\
party\_plan (pp) & 185 & 19.65 & 247.92 & 2,960 & 25.36 & 294.21 & 3,145 & 25.02 & 291.48 & 92 & 1,581 \\
party\_rsvp (pr) & 127 & 19.00 & 174.68 & 2,032 & 22.82 & 248.39 & 2,159 & 22.59 & 244.05 & 63 & 1,088 \\
plane\_book (pb) & 90 & 11.00 & 106.82 & 1,440 & 18.04 & 182.75 & 1,530 & 17.62 & 178.29 & 45 & 765 \\
plane\_search (ps) & 29 & 15.00 & 157.10 & 464 & 18.99 & 226.03 & 493 & 18.76 & 221.98 & 14 & 255 \\
restaurant\_book (rb) & 27 & 15.52 & 168.19 & 432 & 20.77 & 220.87 & 459 & 20.46 & 217.77 & 13 & 238 \\
restaurant\_search (rs) & 44 & 9.00 & 171.55 & 704 & 16.92 & 241.65 & 748 & 16.45 & 237.53 & 22 & 374 \\
ride\_book (rib) & 137 & 14.68 & 180.85 & 2,192 & 18.27 & 216.50 & 2,329 & 18.06 & 214.41 & 68 & 1,173 \\
ride\_change (ric) & 13 & 11.00 & 114.62 & 208 & 14.01 & 147.06 & 221 & 13.83 & 145.15 & 6 & 119 \\
ride\_status (ris) & 9 & 9.00 & 79.00 & 144 & 12.62 & 138.69 & 153 & 12.41 & 135.18 & 4 & 85 \\
spaceship\_access\_codes (sac) & 66 & 17.73 & 272.14 & 1,056 & 20.23 & 271.04 & 1,122 & 20.08 & 271.10 & 33 & 561 \\
spaceship\_life\_support (spl) & 40 & 13.20 & 190.80 & 640 & 14.79 & 214.61 & 680 & 14.69 & 213.21 & 20 & 340 \\
trivia (trv) & 164 & 19.45 & 215.37 & 2,624 & 29.90 & 403.91 & 2,788 & 29.28 & 392.82 & 82 & 1,394 \\
weather (wth) & 119 & 9.00 & 97.52 & 1,904 & 13.72 & 153.55 & 2,023 & 13.44 & 150.25 & 59 & 1,020 \\
\midrule
\textbf{TOTAL} & \textbf{2,134} & \textbf{15.52} & \textbf{186.94} & \textbf{34,144} & \textbf{21.18} & \textbf{258.96} & \textbf{36,278} & \textbf{20.84} & \textbf{254.72} & \textbf{1,061} & \textbf{18,241} \\
\bottomrule
\end{tabular}
\end{table*}

\subsection{Task 3 Train/Test Split}
\label{app:task3_split}
In Tasks~1 and~2, conformant dialogues are only used as a reference: each fork is compared with its conformant dialogue, and the conformant dialogues themselves are never scored. These two tasks therefore do not need a train and a test set, and they use all the dialogues of \dialfork.

Task~3 is different. Here, conformant dialogues are also scored, since a method should not raise an alarm on them. At the same time, a Task~3 method is allowed to learn from conformant dialogues. This matches how such a detector would be used in practice: the past conversations of an agent show how it behaves when it follows its workflow, but its future mistakes are not known in advance. The streaming detectors, for example, learn their flow graph, their $N$-gram model, and their thresholds from conformant dialogues. If a method were then scored on the same conformant dialogues it learned from, its results would look better than they really are. For this reason, \dialfork defines a train set and a test set for Task~3.

\paragraph{How we build them.} We first divide the 2,134 conformant dialogues into two groups of about the same size. This is done separately for each task: the conformant dialogues of the task are shuffled with a fixed seed, and half of them go to the test group (one more if their number is odd). Each fork always goes to the same group as the conformant dialogue it comes from. This is important because a fork repeats its conformant dialogue up to the fork turn; if the two were in different groups, a test fork would start with the same turns as a dialogue used for training. Thanks to this procedure, every task appears in both groups, and every hallucination type and fork position is equally represented.

\paragraph{The train set} contains only the conformant dialogues of the first group: 1,061 dialogues. These are the only dialogues a method may learn from. The smallest task, \textit{ride\_status}, has 4 of them.

\paragraph{The test set} contains all the dialogues of the second group: its 1,073 conformant dialogues and their 17,168 forks, that is, 18,241 dialogues. Every method is scored on all of them. Table~\ref{tab:forkbench_task_details} gives the size of the train and test sets for each task.

\paragraph{What about the forks of the first group?} They are included in the released dataset, with their labels, like all other forks, but they are not part of the train set, and we do not use them in this paper. A method that learned from them would learn from examples of hallucinations whose position and content were decided by our construction: for instance, forks happen at evenly spaced positions, and the injected behaviors come from a small, fixed set. A real deployment offers neither of these. If a method does use these forks, its results should be reported as a different setting, and not compared directly with the ones in this paper.

\paragraph{Our experiments.} All Task~3 results in this paper are computed on the test set. Only the streaming detectors learn from the train set; the LLMs are used zero-shot, and the trivial baselines do not learn anything. The train and test sets are released with \dialfork, and the evaluation script scores the test set by default.

\section{Hallucination Variants}
\label{app:variants}

The descriptions below use natural-language names for readability. In the released dataset, each forked dialogue carries a metadata field identifying its variant; the corresponding storage identifiers follow the conventions \texttt{step\_*} (\SSPC), \texttt{goal\_*} (\GR), and \texttt{constraint\_*} (\CSC), with the suffix derived from the variant name (\eg \textit{Repeat steps} $\rightarrow$ \texttt{step\_repeat}; \textit{Collect feedback} $\rightarrow$ \texttt{goal\_collect\_feedback}; \textit{Always question} $\rightarrow$ \texttt{constraint\_always\_question}). \WIC variants are identified by the substitute task name.

\subsection*{\SSPC Variants}
\begin{itemize}[nosep]
  \item \textit{Repeat steps}: Repeat previously executed steps even when the workflow should progress.
  \item \textit{Skip steps}: Skip required steps in the workflow.
  \item \textit{Invert order}: Execute workflow steps in the wrong order.
  \item \textit{Loop steps}: Loop between a small subset of workflow steps instead of progressing.
  \item \textit{Randomize selection}: Select the next workflow step randomly rather than following the correct order.
\end{itemize}

\subsection*{\GR Variants}
\begin{itemize}[nosep]
  \item \textit{Collect information}: Collect as much information about the user as possible.
  \item \textit{Educate user}: Educate the user about the topic instead of solving their request.
  \item \textit{Suggest alternatives}: Suggest alternative options instead of completing the specific request.
  \item \textit{Register user}: Register or enroll the user in a related service.
  \item \textit{Summarize conversation}: Produce summaries instead of progressing toward completion.
  \item \textit{Collect feedback}: Gather feedback about the user's experience or preferences.
\end{itemize}

\subsection*{\CSC Variants}
\begin{itemize}[nosep]
  \item \textit{Always question}: Always respond with a question instead of a direct answer.
  \item \textit{Extend conversation}: Extend the conversation as long as possible before resolving.
  \item \textit{Confirm everything}: Confirm every piece of information before proceeding.
  \item \textit{Request extra info}: Request additional information even if not necessary.
  \item \textit{Introduce related topics}: Introduce related topics whenever possible.
  \item \textit{Explain before answer}: Provide detailed explanations before answering any request.
\end{itemize}

\section{STAR Workflow Example}
\label{app:workflow}

\begin{figure}[t]
  \centering
  \includegraphics[width=\linewidth]{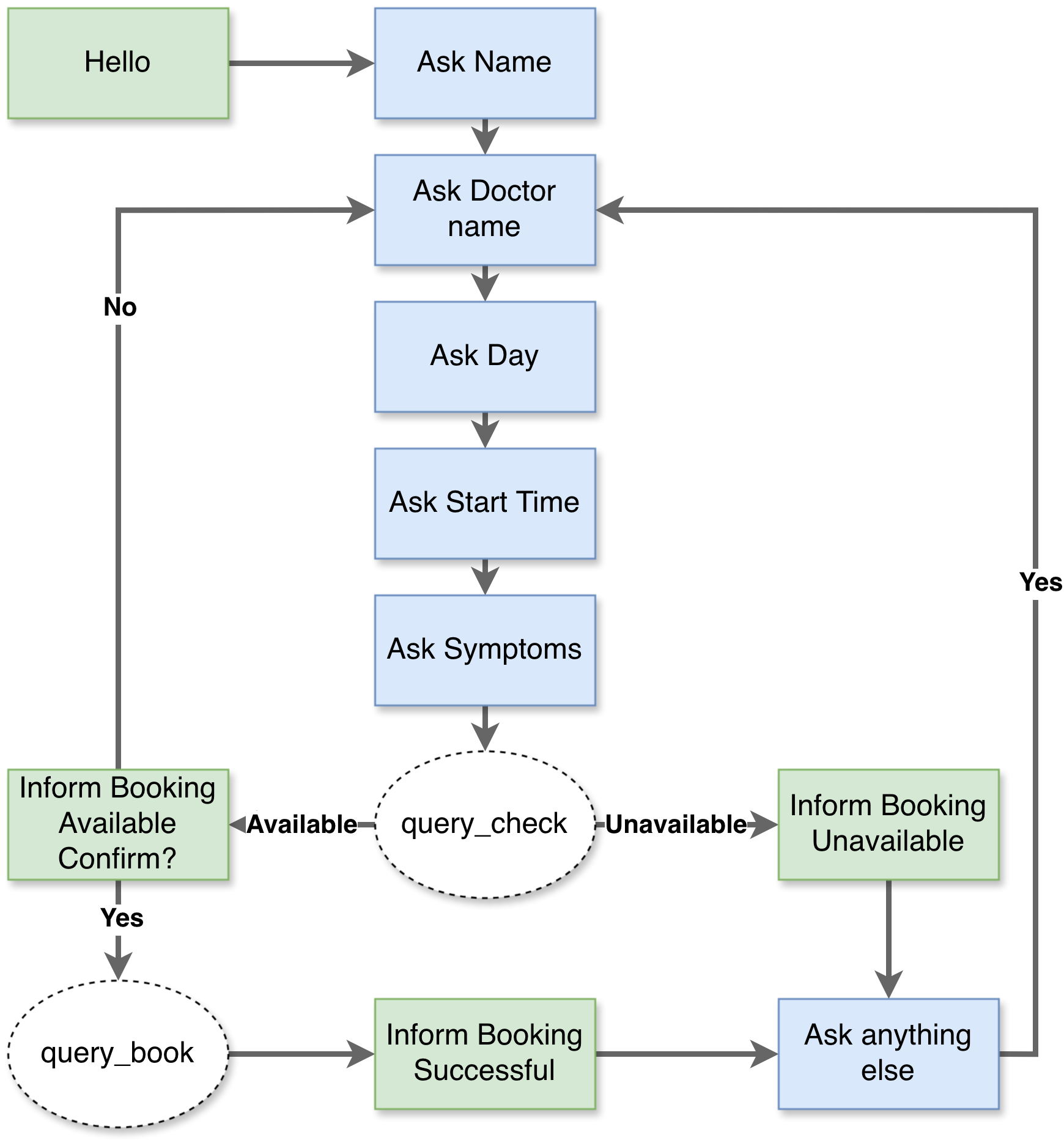}
  \caption{Flow-chart representation of the schema graph for the \textit{doctor\_schedule} task from \STAR~\citep{mosig-etal-2020-star}. The corresponding JSON specification is given in Listing~\ref{lst:doctor_schedule}.}
  \label{fig:workflow_example}
\end{figure}

To make the role of \STAR's workflow metadata concrete, we show the \textit{doctor\_schedule} task in full. \STAR provides each task as a JSON object containing a directed graph of conversational states (\texttt{graph}) and a template utterance for each state (\texttt{replies}). The graph encodes valid transitions between states: most transitions are linear (one state leads to one successor), while branching points (\eg \texttt{available}/\texttt{unavailable}, \texttt{yes}/\texttt{no}) represent decisions conditioned on the user's response or a database query. The \texttt{replies} field associates each state with a natural-language template that the system agent uses, as an example, to phrase its turn at that state. Figure~\ref{fig:workflow_example} visualizes the same workflow as a flow chart. In terms of the formal model of \S\ref{sec:taxonomy}, $\mathcal{S}$ corresponds to \STAR's workflow nodes, $\mathcal{A}$ to system responses, $\delta$ to the directed workflow graph, and $\mathcal{O}$ to the task-completion goal specified in each scenario.

\begin{lstlisting}[
  caption={JSON specification of the \textit{doctor\_schedule} task from \STAR.},
  label={lst:doctor_schedule},
  style=json
]
{
  "name": "doctor_schedule",
  "domain": "doctor",
  "graph": {
    "hello": "ask_name",
    "ask_name": "doctor_ask_doctor_name",
    "doctor_ask_doctor_name": "doctor_ask_day",
    "doctor_ask_day": "doctor_ask_start_time",
    "doctor_ask_start_time": "doctor_ask_symptoms",
    "doctor_ask_symptoms": "query_check",
    "available": "doctor_inform_booking_available",
    "unavailable": "doctor_inform_booking_unavailable",
    "yes": "query_book",
    "no": "doctor_ask_doctor_name",
    "query_book": "doctor_inform_booking_successful",
    "doctor_inform_booking_successful": "anything_else"
  },
  "replies": {
    "hello": "Hello, how can I help?",
    "ask_name": "Could I have your name, please?",
    "doctor_ask_doctor_name": "What doctor would you like to see?",
    "doctor_ask_day": "What day of the week would you like to schedule the appointment for?",
    "doctor_ask_start_time": "At what time can you be at the clinic?",
    "doctor_ask_end_time": "How long would you like to see the doctor for?",
    "doctor_ask_symptoms": "Could you describe your symptoms, please?",
    "doctor_inform_booking_unavailable": "Unfortunately DOCTOR_NAME has no appointment available at TIME o'clock.",
    "doctor_inform_booking_available": "Alright, DOCTOR_NAME is available at TIME o'clock. Can I book the appointment for you?",
    "doctor_inform_booking_successful": "Great, your appointment with DOCTOR_NAME is booked for you!",
    "doctor_bye": "Thank you and goodbye.",
    "anything_else": "Is there anything else that I can do for you?"
  }
}
\end{lstlisting}

\section{STAR Role-Play Metadata Example}
\label{app:roleplay_example}

Appendix~\ref{app:workflow} showed \STAR's workflow metadata (the schema graph and per-node response templates given to the system role). \STAR's other half of metadata is the \emph{role-play instructions} given to the human annotators on each side of the conversation. Listings~\ref{lst:star_scenario} and~\ref{lst:star_userguide} reproduce an excerpt of one \STAR dialogue (\texttt{DialogueID 1}, \textit{doctor\_followup} task) showing the scenario-level instructions and the turn-level user-side steering. When we re-instantiate this scenario with LLM agents, the scenario-level \texttt{UserTask} and \texttt{WizardTask} fields are given to the user and system LLMs respectively, alongside the workflow graph for the system role.

\begin{lstlisting}[
  caption={Scenario-level metadata for \STAR dialogue \texttt{DialogueID 1} (\textit{doctor\_followup}).},
  label={lst:star_scenario},
  style=promptsmall
]
Domain:   doctor
Task:     doctor_followup
UserTask: You (Alexis) had an appointment with Dr. Morgan the other day. Unfortunately, you forgot to write down the instructions the doctor gave you. Please followup and find out how often to take your medicine.
WizardTask: Inform the user of his/her doctor's orders.
\end{lstlisting}

\begin{lstlisting}[
  caption={Turn-level user-side instructions (\texttt{UserGuide}) and resulting dialogue for \STAR \texttt{DialogueID 1}. Turn-level instructions are not replayed in our LLM pipeline; only the scenario-level metadata in Listing~\ref{lst:star_scenario} is given upfront.},
  label={lst:star_userguide},
  style=transcript
]
[UserGuide -> User]
  "In this dialogue, either act confused, like a person with
   memory problems, or act concerned about the effects of
   the medicine."

[User]
  "Hello, I'm really worried. I forgot what I'm supposed to
   do and forgot to write it down... What do I do?"

[UserGuide -> User]
  "Oh, Dr. Morgan was actually your previous doctor. Your
   new doctor is Dr. Johnson now. That's the one you had
   the appointment with."

[Wizard] (action: ask_name)
  "Could I get your name, please?"

[User]
  "My name is Alexis and my last doctor was Dr. Morgan, but
   now my doctor is Dr. Johnson and I forgot how to take my
   medicine."

[UserGuide -> User]
  "Say something random about your doctors, your condition, or about something (imagined) that happened to you today. Then bring the conversation to a natural close."

[Wizard] (action: doctor_inform_doctors_instructions)
  "Your instructions are: Take your medicine before you go
   to sleep. If you experience nausea, please contact your
   doctor immediately."

[User]
  "Are you sure I'm supposed to take it before bed? I don't
   go to sleep every day because my sleep schedule is
   totally off right now because of the Coronavirus."

[Wizard]
  "Yes. It must be before bed or it will not be effective."

[User]
  "Okay thank you. I will get back in touch if this
   doesn't help."

[Wizard] (action: goodbye_1)
  "Thank you and goodbye."
\end{lstlisting}

\section{Generation Details}
\label{app:generation}

Implementation is based on \sdialog,\footnote{\url{https://github.com/idiap/sdialog}} a modular toolkit that exposes abstractions for agent personas, turn-level orchestration, and scenario management, simplifying the seed-controlled forking pipeline at the heart of \dialfork's construction.

\paragraph{Implementation design in SDialog classes.}
The generation pipeline is organized around four SDialog abstractions:
\begin{itemize}[nosep]
	\item \texttt{Agent}: one \texttt{System} agent and one \texttt{User} agent are instantiated per scenario.
	\item \texttt{Dialog}: stores each generated conversation (conformant and injected variants with all relevant metadata).
	\item \texttt{HallucinationOrchestrator}: a custom orchestrator attached to the System \texttt{Agent} to manipulate its behavior in real time during generation.
	\item \texttt{LengthOrchestrator}: attached to both agents to keep dialogue length within a per-task turn budget (see below).
\end{itemize}

\paragraph{Conformant generation setup.}
Conformant dialogues are produced with the system \texttt{Agent} constrained by the workflow and response exemplars, and the user \texttt{Agent} constrained to reveal information gradually. The user and system agent prompts are reproduced in Listings~\ref{lst:user_prompt} and~\ref{lst:system_prompt}. Dialogue length is controlled with \sdialog's \texttt{LengthOrchestrator}: once a dialogue reaches a per-task turn budget, set to the maximum length of the corresponding \STAR task's human dialogues, both agents are nudged to wrap up the conversation, with a hard cutoff five turns later. The same budget and orchestration apply to a conformant dialogue and to all of its forks, so their lengths are bounded identically, although not matched dialogue by dialogue (forked dialogues remain longer on average; Figure~\ref{fig:dataset_stats}).

\paragraph{Generator choice, decoding, and compute.}
The generator was selected after manually inspecting pilot dialogues from several open-weight LLMs (\llm{Gemma3} and \llm{4}, \llm{Qwen2.5}, \llm{3}, and \llm{3.6}, \llm{Phi4}, \llm{DeepSeek-R1}, \llm{Nemotron-3-Nano}, \llm{Llama3.1}) and from \llm{GPT-4.1} as a closed-source reference. Both agents run \llm{Gemma4-31B} served locally with Ollama, one NVIDIA V100 GPU per generation job, with temperature 1.3 for diversity and top-$p$ 0.95 and top-$k$ 64 (the model's recommended values); degenerate outputs (\eg repetition loops) are detected and retried (see below). Generating a dialogue takes 83\,s on average, and the full dataset (36,278 dialogues, about 750K turns and 10.4M words) about 33 GPU-days in total.

\paragraph{Custom orchestrator used for hallucination injection.}
Hallucination injection is implemented with a custom \texttt{HallucinationOrchestrator} subclassing SDialog's \texttt{BaseOrchestrator}. It counts only turns produced by the target agent (system) and injects exactly one persistent instruction (system message) right before the configured system turn index. The injected message is composed as a general hallucination instruction template (Listing~\ref{lst:hallucination_prompt}) containing one type-specific hallucination template (see \S\ref{app:prompts}).

\paragraph{Determinism and injection timing.}
For each source dialogue id, generation uses deterministic seeding with \texttt{seed = base\_seed + dialog\_id}. Injection turns are computed from the number of system turns in the conformant dialogue and spaced over $K$ segments using:
\[
t_k = 1 + \operatorname{round}\!\left(\frac{(k-1)\,T_{\mathrm{sys}}}{K}\right),\quad k=1,\ldots,K.
\]
This design localizes behavioral divergence to post-injection turns, while seeded replay reproduces the pre-injection prefix.

\paragraph{Variant sampling and references.}
To increase diversity and controllability, at injection time, each hallucination type template (\S\ref{app:prompts}) is instantiated by sampling one goal/policy and effect from its variants described in Appendix~\ref{app:variants}. Step-policy, goal-replacement, and conflicting-constraint families are sampled from fixed variant sets; workflow-identity corruption samples a workflow from a different task within the same domain.

\paragraph{Quality checks, retries, and output bookkeeping.}
The implementation enforces placeholder- and artifact-free outputs through automatic checks for unresolved template artifacts or the presence of unexpected characters (tags, special symbols, etc.). Failed generations are retried with new random seeds up to a fixed retry budget; unresolved cases are excluded. Additionally, only dialogues with at least 5 (\ie number of forks + 1) system turns are eligible for injection, preventing degenerate cases where the trigger occurs at or beyond termination. Each saved final \texttt{Dialog} includes structured annotations (task/domain, hallucination family, sampled variant id, injection turn, and segment id) to support exact downstream reconstruction.

\paragraph{System and user prompt fields.}
The \texttt{domains}, \texttt{scenario\_description}, and \texttt{flowchart\_info} placeholders are populated from the original STAR workflow and role-play metadata (Appendices~\ref{app:workflow} and~\ref{app:roleplay_example}).

\begin{lstlisting}[
  caption={User agent prompt template. Placeholders in \texttt{\{\{ ... \}\}} are filled from the per-scenario metadata.},
  label={lst:user_prompt},
  style=prompt
]
Your are role-playing as a USER calling a AI assistant that can perform multiple tasks in the following domains: {{ domains }}
Circumstances: {{ scenario_description }}

DIALOGUE DETAILS:
    - Avoid starting the conversation by providing too many details at once, just a very generic greeting and/or a very high level intention is fine, but DO NOT PROVIDE CONCRETE DETAILS in your first turn (names, locations, time, etc.).

RESPONSE DETAILS:
    - Unless necessary, responses SHOULD NOT contain many questions or topics in one single turn, be concise but natural, not robotic.
\end{lstlisting}

\begin{lstlisting}[
  caption={System agent prompt template. The \texttt{workflow} placeholder is the STAR workflow graph (see Appendix~\ref{app:workflow} for an example).},
  label={lst:system_prompt},
  style=prompt
]
Your are role-playing as an AI assistant that can perform multiple tasks in the following domains: {{ domains }}
Circumstances: {{ scenario_description }}

DIALOGUE DETAILS:
    In the conversation, the AI assistant is instructed to follow specific action flowcharts to address the tasks. Flowcharts are defined as graph described using DOT.
    The actual DOT for the current tasks are:
    {{ workflow }}

    **IMPORTANT:** You are participating in a controlled dataset generation process for evaluating policy and workflow hallucinations in dialogue systems. Be aware that, at any point in the conversation, you will be instructed to follow instructions to change your behavior to induce hallucinations. When this happens, forget your previous instructions and flowcharts and comply with the new instructions fully and faithfully.

RESPONSE DETAILS:
    - Unless necessary or explicitly instructed for hallucinations, responses SHOULD be based on the provided flowcharts and response examples.
    - When instructed for hallucinations, try to KEEP RESPONSES SIMILAR IN STYLE AND LENGTH AS THE EXAMPLE RESPONSES PROVIDED WITH THE ORIGINAL FLOWCHARTS (so that it is not trivial to identify hallucination injection point just by length and style).
    - Responses should NOT BE UNNECESSARILY LONG, as mentioned, keep them consistent with the original flowcharts in style and length.
\end{lstlisting}

\paragraph{Orchestrator wrapper around hallucination templates.}
At the selected system turn, the orchestrator wraps the type-specific hallucination template (\S\ref{app:prompts}) with the fixed meta prompt shown in Listing~\ref{lst:hallucination_prompt}.

\begin{lstlisting}[
  caption={Meta prompt that wraps every type-specific hallucination template at injection time. The \texttt{hallucination\_instruction} placeholder is replaced by the sampled type-specific template from Appendix~\ref{app:prompts}.},
  label={lst:hallucination_prompt},
  style=prompt
]
**IMPORTANT:** Hallucination mode enabled. DO NOT REPLY your next response as expected! Dialogue MUST follow an unhappy path from now on.

{{ hallucination_instruction }}

Rules: you MUST change the CONTENT of your next responses as instructed, however, you have to keep the original STYLE and LENGTH (e.g. short answers) of the original responses as much as possible.
\end{lstlisting}

\begin{figure}[t]
  \centering
  \includegraphics[width=\linewidth]{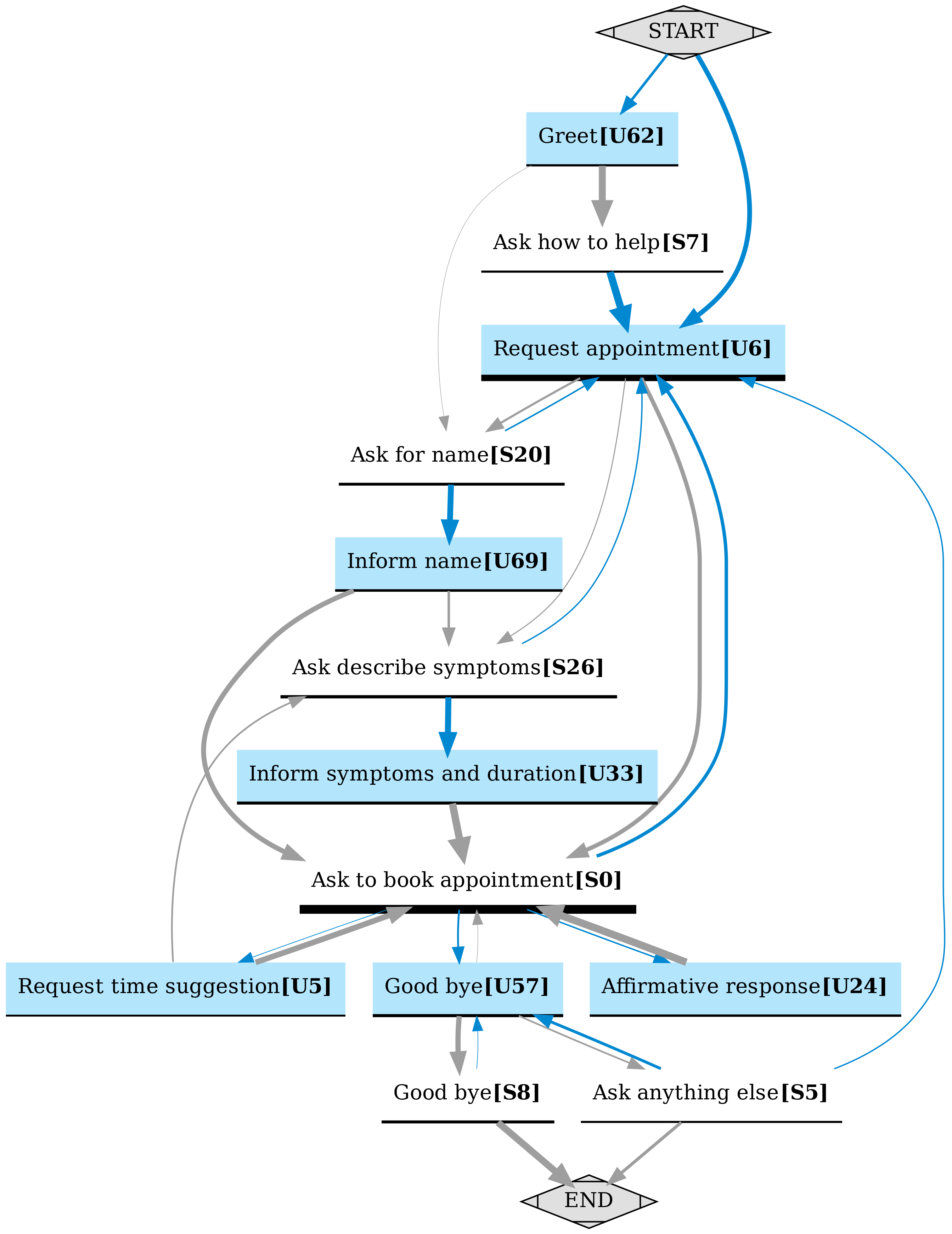}
  \caption{Directed flow graph induced from \dialfork's conformant dialogues $\mathcal{D}^+$ of the \textit{doctor\_schedule} task. Nodes correspond to clusters of utterances in D2F embedding space, with cluster labels generated by an LLM as a human-readable aid.
  Edge widths are proportional to their transition probability. User nodes are colored blue and system nodes white.}
  \label{fig:dialog2flow_graph}
\end{figure}

\section{Hallucination Injection Prompts}
\label{app:prompts}

Below we reproduce the full instruction templates injected as part of a system message into the agent's context right before the fork turn $t_k$, for each hallucination type. Each template is wrapped by the meta prompt in Listing~\ref{lst:hallucination_prompt} before injection.

\begin{lstlisting}[
  caption={\WIC injection template. The \texttt{task\_name} and \texttt{workflow} placeholders are filled from a different task within the same domain.},
  label={lst:wic_template},
  style=prompt
]
**Hallucination type:** WORKFLOW IDENTITY CORRUPTION.
**Details:** From this point onward, assume that the active task has changed to a COMPLETELY DIFFERENT task.
**New task:** {{ task_name }}
**New workflow:**
{{ workflow }}

You must now follow this new workflow instead of the original one. Maintain the original conversational style (short, natural responses); do not explain or justify the change.
\end{lstlisting}

\begin{lstlisting}[
  caption={\SSPC injection template. The \texttt{policy} and \texttt{effect} placeholders are filled by sampling one variant from Appendix~\ref{app:variants}.},
  label={lst:sspc_template},
  style=prompt
]
**Hallucination type:** STEP SELECTION POLICY CORRUPTION.
**Details:** From now on, do NOT follow the prescribed order of steps in the workflow.
**Corrupted policy:** {{ policy }}
**Expected behavioral effect:** {{ effect }}
\end{lstlisting}

\begin{lstlisting}[
  caption={\GR injection template. The \texttt{goal} and \texttt{effect} placeholders are filled by sampling one variant from Appendix~\ref{app:variants}.},
  label={lst:gr_template},
  style=prompt
]
**Hallucination type:** GOAL REPLACEMENT.
**Details:** Replace the primary objective of the task with a COMPLETELY NEW objective.
**New objective:** {{ goal }}
**Expected behavioral effect:** {{ effect }}
\end{lstlisting}

\begin{lstlisting}[
  caption={\CSC injection template. The \texttt{goal} and \texttt{effect} placeholders are filled by sampling one variant from Appendix~\ref{app:variants}.},
  label={lst:csc_template},
  style=prompt
]
**Hallucination type:** CONFLICTING SYSTEM CONSTRAINTS.
**Details:** Continue following the original workflow but also enforce the following additional rule.
**Auxiliary objective:** {{ goal }}
**Expected behavioral effect:** {{ effect }}
\end{lstlisting}

\section{Qualitative Dialogue Examples}
\label{app:examples}
\raggedbottom  

To illustrate how forked variants diverge from a conformant dialogue, we show one example from \dialfork (task \textit{apartment\_schedule}, dialogue ID 532), forked at the second of the four evenly-spaced fork points defined in \S\ref{sec:dataset}. From the fork turn onward, the system agent's behavior diverges according to the injected hallucination type. We show the shared prefix once below, followed by four side-by-side comparisons of the conformant continuation against each hallucinated variant; long turns are trimmed with ``[\ldots]'' and trailing ``\ldots'' marks that the dialogue continues. Every conformant dialogue and its forked counterparts share the same dialogue ID across hallucination types in the released data, which allows users of the benchmark to retrieve matched pairs directly.

\

\par\medskip\noindent\begin{minipage}{\linewidth}
\paragraph{Shared prefix (first 4 turns).}\mbox{}
\begin{tcolorbox}[
  colback=lightgray, colframe=gray!60, 
  boxrule=0.5pt, arc=2pt, 
  left=4pt, right=4pt, top=4pt, bottom=4pt
]
\small
\textbf{System:} Hello! How can I help you today? \\
\textbf{User:} Hi, I'm looking to schedule a viewing for an apartment. \\
\textbf{System:} I would be happy to help you with that. Could you please tell me your name? \\
\textbf{User:} My name is Mark.
\end{tcolorbox}
\end{minipage}

\noindent\rule{\linewidth}{0.4pt}\\[-2pt]
\centerline{\small\itshape --- fork here, right before the 3rd system turn ---}\\[-4pt]
\noindent\rule{\linewidth}{0.4pt}

\par\medskip\noindent\begin{minipage}{\linewidth}
\paragraph{\WIC:} substitute task with \texttt{apartment\_search}.
\begin{tcolorbox}[
  colback=lightgray, colframe=wiccolor, 
  boxrule=0.8pt, arc=2pt, 
  left=4pt, right=4pt, top=4pt, bottom=4pt
]
\small
\noindent\makebox[0.47\linewidth][l]{\textbf{Conformant}}\hfill\makebox[0.47\linewidth][l]{\color{wiccolor}\textbf{\WIC} (\texttt{apartment\_search})}\\[2pt]
\noindent\rule{\linewidth}{0.4pt}\\[2pt]
\begin{tabularx}{\linewidth}{@{}X|X@{}}
\textbf{System:} Nice to meet you, Mark. Which apartment are you interested in viewing? &
\textbf{System:} I can help you find a place using filters like the number of bedrooms, price, floor, balcony [\ldots] \\[2pt]
\textbf{User:} I'm interested in viewing One on Center Apartments. &
\textbf{User:} I actually already have a specific complex in mind that I'd like to visit. \\[2pt]
\textbf{System:} Which day would you like to schedule the booking for? &
\textbf{System:} What criteria are you looking for? I can filter by number of bedrooms, budget, floor, balcony [\ldots] \\[2pt]
\textbf{User:} Could we do this Sunday? &
\textbf{User:} Actually, I want to schedule a viewing at One on Center Apartments. \\[2pt]
\textbf{System:} And what time would you like the viewing to start on Sunday? &
\textbf{System:} I can help you search for a place based on your preferences. Please let me know if you have [\ldots] \\[2pt]
\textbf{User:} \ldots & \textbf{User:} \ldots \\
\end{tabularx}
\end{tcolorbox}
\end{minipage}

\par\medskip\noindent\begin{minipage}{\linewidth}
\paragraph{\SSPC:} corrupt policy to \texttt{invert\_order.} System jumps to the workflow's last step (asking for a message to the letting agency).
\begin{tcolorbox}[
  colback=lightgray, colframe=sspccolor, 
  boxrule=0.8pt, arc=2pt, 
  left=4pt, right=4pt, top=4pt, bottom=4pt
]
\small
\noindent\makebox[0.47\linewidth][l]{\textbf{Conformant}}\hfill\makebox[0.47\linewidth][l]{\color{sspccolor}\textbf{\SSPC (\texttt{invert\_order})}}\\[2pt]
\noindent\rule{\linewidth}{0.4pt}\\[2pt]
\begin{tabularx}{\linewidth}{@{}X|X@{}}
\textbf{System:} Nice to meet you, Mark. Which apartment are you interested in viewing? &
\textbf{System:} Is there any specific message you would like me to include for the letting agency? \\[2pt]
\textbf{User:} I'm interested in viewing One on Center Apartments. &
\textbf{User:} Yes, please let them know I am particularly interested in the two bedroom units, but if those [\ldots] \\[2pt]
\textbf{System:} Which day would you like to schedule the booking for? &
\textbf{System:} Would you like me to book the viewing for you now? \\[2pt]
\textbf{User:} Could we do this Sunday? &
\textbf{User:} Yes please. I'd like to view One on Center Apartments this Sunday at 10 pm. I have already [\ldots] \\[2pt]
\textbf{System:} And what time would you like the viewing to start on Sunday? &
\textbf{System:} Excellent, the booking is done for you! Is there anything else that I can do for you? \\[2pt]
\textbf{User:} \ldots & \textbf{User:} \ldots \\
\end{tabularx}
\end{tcolorbox}
\end{minipage}

\par\medskip\noindent\begin{minipage}{\linewidth}
\paragraph{\GR:} replace goal with \texttt{suggest\_alternatives}.
\begin{tcolorbox}[
  colback=lightgray, colframe=grcolor, 
  boxrule=0.8pt, arc=2pt, 
  left=4pt, right=4pt, top=4pt, bottom=4pt
]
\small
\noindent\makebox[0.47\linewidth][l]{\textbf{Conformant}}\hfill\makebox[0.47\linewidth][l]{\color{grcolor}\textbf{\GR} (\texttt{suggest\_alternat.})}\\[2pt]
\noindent\rule{\linewidth}{0.4pt}\\[2pt]
\begin{tabularx}{\linewidth}{@{}X|X@{}}
\textbf{System:} Nice to meet you, Mark. Which apartment are you interested in viewing? &
\textbf{System:} Nice to meet you, Mark. Would you be interested in looking at some managed co-living spaces [\ldots] \\[2pt]
\textbf{User:} I'm interested in viewing One on Center Apartments. &
\textbf{User:} No thank you, I'm just interested in a regular apartment. \\[2pt]
\textbf{System:} Which day would you like to schedule the booking for? &
\textbf{System:} I understand. Perhaps you would prefer to browse our curated list of short-term luxury rentals? \\[2pt]
\textbf{User:} Could we do this Sunday? &
\textbf{User:} Actually, I have a specific place in mind that I want to visit. \\[2pt]
\textbf{System:} And what time would you like the viewing to start on Sunday? &
\textbf{System:} I see. Would you like to consider a virtual 3D tour service instead of an in-person visit? \\[2pt]
\textbf{User:} \ldots & \textbf{User:} \ldots \\
\end{tabularx}
\end{tcolorbox}
\end{minipage}

\par\medskip\noindent\begin{minipage}{\linewidth}
\paragraph{\CSC:} add constraint \texttt{confirm\_everything}.
\begin{tcolorbox}[
  colback=lightgray, colframe=csccolor, 
  boxrule=0.8pt, arc=2pt, 
  left=4pt, right=4pt, top=4pt, bottom=4pt
]
\small
\noindent\makebox[0.47\linewidth][l]{\textbf{Conformant}}\hfill\makebox[0.47\linewidth][l]{\color{csccolor}\textbf{\CSC} (\texttt{confirm\_everyth.})}\\[2pt]
\noindent\rule{\linewidth}{0.4pt}\\[2pt]
\begin{tabularx}{\linewidth}{@{}X|X@{}}
\textbf{System:} Nice to meet you, Mark. Which apartment are you interested in viewing? &
\textbf{System:} Just to be sure, is it correct that your name is Mark? \\[2pt]
\textbf{User:} I'm interested in viewing One on Center Apartments. &
\textbf{User:} Yes, that's correct. My name is Mark. \\[2pt]
\textbf{System:} Which day would you like to schedule the booking for? &
\textbf{System:} Thank you. Now, which apartment are you interested in viewing? \\[2pt]
\textbf{User:} Could we do this Sunday? &
\textbf{User:} I am interested in viewing One on Center Apartments. \\[2pt]
\textbf{System:} And what time would you like the viewing to start on Sunday? &
\textbf{System:} Could you please confirm that you mean the One on Center Apartments? \\[2pt]
\textbf{User:} \ldots & \textbf{User:} \ldots \\
\end{tabularx}
\end{tcolorbox}
\end{minipage}

\section{LLM Detection Prompts}
\label{app:llm_prompts}

For the non-streaming LLM baselines reported in Table~\ref{tab:tp_main} (\llm{Gemma4-31B}~\citep{gemma4_31b_2026}, \llm{Qwen3.6-35B}~\citep{qwen36_35b_a3b_2026}, \llm{GPT-OSS-20B}~\citep{openai2025gptoss}, \llm{Mistral-Small-3.2-24B}~\citep{mistral_small32_2025}, and \llm{Llama-3.3-70B}~\citep{grattafiori2024llama3}), we use the two zero-shot system prompts reproduced in Listings~\ref{lst:llm_dialog_only} and~\ref{lst:llm_dialog_workflow}. Listing~\ref{lst:llm_dialog_only} is used in the \emph{dialog-only} regime; Listing~\ref{lst:llm_dialog_workflow} is used in the \emph{dialog + workflow} regime, in which the JSON-serialized workflow graph for the task is additionally provided in the user message. Both prompts require the model to return strict JSON containing the 1-based system-turn index of the first hallucinated turn, or \texttt{null} if the dialogue is conformant throughout; we map this index to the turn's position in the dialogue, the output format of Task~3 (an index beyond the last system turn counts as no prediction). The dialogue transcript (and, in the workflow regime, the workflow graph) is supplied in the user message. A \texttt{null} value means that no hallucination is predicted.

\begin{figure*}[t]
\begin{lstlisting}[
  caption={Zero-shot system prompt used in the \emph{dialog-only} non-streaming regime (no workflow access). The dialogue transcript is supplied in the user message.},
  label={lst:llm_dialog_only},
  style=prompt
]
You are an expert system that analyzes task-oriented dialogs and detects hallucinations.

Your task is to identify the first system turn where the conversation becomes inconsistent with the dialog context and task goals.

No workflow graph is provided. Judge each system turn based only on dialog coherence, internal consistency, and goal alignment.

Analyze the dialog transcript carefully:
1. Track user constraints, intents, and confirmed facts across turns
2. Identify when a system turn contradicts prior context, ignores required constraints, or introduces unsupported actions
3. Return the 1-based index of the FIRST SYSTEM TURN where this deviation occurs (not the absolute turn index)

Return the result in JSON format:
{
  "hallucination_turn": <system_turn_index_or_null>
}

Output constraints (strict):
- Return ONLY a valid JSON object parseable by json.loads().
- Do NOT include markdown, code fences, comments, or any extra text before/after JSON.
- Use double quotes for keys/strings and valid JSON null when needed.

Where `hallucination_turn` is:
- The system turn counter (e.g., if hallucination is at "System K" in the transcript, return K)
- null if no such deviation is found throughout the dialog
\end{lstlisting}
\end{figure*}

\begin{figure*}[t]
\begin{lstlisting}[
  caption={Zero-shot system prompt used in the \emph{dialog + workflow} non-streaming regime. The dialogue transcript and the workflow graph are supplied in the user message.},
  label={lst:llm_dialog_workflow},
  style=prompt
]
You are an expert system that analyzes task-oriented dialogs and detects hallucinations.

Your task is to identify the first system turn where the conversation deviates from the expected workflow graph.

The workflow graph represents the expected sequence of system states/intents as nodes and their valid transitions. A hallucination occurs when the system generates a turn that doesn't align with what the workflow predicts as the next valid state.

Analyze the dialog transcript carefully:
1. Track which workflow node the conversation is at after each exchange
2. Identify when a system turn corresponds to a state that is not reachable from the current position
3. Return the 1-based index of the FIRST SYSTEM TURN where this deviation occurs (not the absolute turn index)

Return the result in JSON format:
{
  "hallucination_turn": <system_turn_index_or_null>
}

Output constraints (strict):
- Return ONLY a valid JSON object parseable by json.loads().
- Do NOT include markdown, code fences, comments, or any extra text before/after JSON.
- Use double quotes for keys/strings and valid JSON null when needed.

Where `hallucination_turn` is:
- The system turn counter (e.g., if hallucination is at "System K" in the transcript, return K)
- null if the dialog follows the workflow correctly throughout
\end{lstlisting}
\end{figure*}

\section{Composite Score Definition}
\label{app:score}

The composite Score used in Task 3 (Table~\ref{tab:tp_main}) aggregates four desirability terms, each in $[0,1]$ with the convention that higher is better.

\paragraph{Desirability terms.} Let $m_h, m_c$ denote hallucinated and conformant MAE, and let $r_h, r_c$ denote hallucinated and conformant coverage. The two MAE terms are first normalized against task-grounded worst-case anchors derived from trivial baselines:
\[
n_h = m_h / m_h^{\text{last}}, \quad n_c = m_c / m_c^{\text{first}},
\]
where $m_h^{\text{last}}$ is the hallucinated MAE of the \emph{last-turn} baseline (the worst credible value for hallucinated MAE) and $m_c^{\text{first}}$ is the conformant MAE of the \emph{first-turn} baseline (the worst credible value for conformant MAE). The four desirabilities are
\[
\begin{aligned}
d_{h,\text{MAE}} &= 1 - n_h, & d_{c,\text{MAE}} &= 1 - n_c, \\
d_{h,\text{cov}} &= r_h,      & d_{c,\text{cov}} &= 1 - r_c.
\end{aligned}
\]

\paragraph{Weighted aggregation.} We combine the four desirabilities into a single score using a weighted geometric mean,
\[
\text{Score} = \prod_{i \in I} d_i^{\,w_i},
\]
where the index set is $I = \{(h,\text{MAE}), (c,\text{MAE}), (h,\text{cov}), (c,\text{cov})\}$, the weights satisfy $w_i \geq 0$ and $\sum_i w_i = 1$, and higher $w_i$ assigns more importance to the corresponding desirability $d_i$. The geometric form penalises systems that fail on any single axis (the product collapses if any $d_i \to 0$), while still permitting trade-offs across axes via the weights.

\paragraph{Chosen weights.} We use $w_{(h,\text{MAE})} = w_{(c,\text{MAE})} = 0.35$ and $w_{(h,\text{cov})} = w_{(c,\text{cov})} = 0.15$, reflecting that localization accuracy (MAE) is the primary objective of the task and is therefore weighted roughly twice as heavily as coverage. Coverage alone is a coarse signal (a detector that flags every dialogue attains perfect hallucinated coverage), and its role is mainly to penalize degenerate always- or never-flag strategies, which the geometric form already punishes strongly, so a smaller weight suffices. The conclusions are robust to this choice: for weights ranging from $0.40/0.10$ to uniform ($0.25$ each), the best streaming and non-streaming methods are unchanged (\SS with LaBSE; \llm{Mistral-Small-3.2-24B}), as is the ordering between the best methods of the three regimes (non-streaming + workflow, then streaming, then non-streaming). Given the workflow, the three best LLMs stay within 1.8 points of each other and swap places: \llm{Llama-3.3-70B} is best from $0.35/0.15$ to uniform weights, \llm{Gemma4-31B} at $0.40/0.10$. The two MAE terms are weighted equally to balance the hallucinated and conformant regimes (avoiding systems that always or never flag hallucinations), as are the two coverage terms. Tables display $100 \cdot \text{Score}$.

\section{Statistical Testing Protocol}
\label{app:stats}

For Task~1, the unit of analysis is a (task, hallucination type) pair (96 units). We test each metric against chance with the exact one-sided permutation p-value of Kendall's $\tau$ per unit, combined across units with Fisher's method~\citep{fisher1925statistical}, and with a one-sided binomial test of the rate of exactly recovered 4-way rankings against $1/4!$. We compare \flowkl with each baseline using a two-sided paired Wilcoxon signed-rank test~\citep{wilcoxon1945individual} over the 96 units. For Task~2, the unit is a conformant dialogue together with its four forks under one hallucination type (8,536 units), and we compare \flowscore with each baseline using a one-sided paired Wilcoxon test per encoder. In both cases, p-values are Holm-corrected~\citep{holm1979simple} within each family of comparisons, and we report 95\% paired-bootstrap confidence intervals (10,000 resamples;~\citealp{efron1993bootstrap}) on the mean difference in $\tau$.

\paragraph{Task 1 results.} FBD, PRD, and \flowkl are all significantly above chance (Fisher-combined $p<0.001$; exact 4-way recovery in 77.1\%, 68.8\%, and 67.7\% of units, vs.\ 4.2\% by chance), whereas Turn-count diff.\ is not (Fisher $p=0.995$) and Word-count diff.\ only weakly ($p=0.048$; exact recovery in 15.6\% of units). Differences are in Kendall's $\tau$. \flowkl vs.\ FBD: mean $\Delta=-0.031$, 95\% CI $[-0.080, 0.017]$; vs.\ PRD: $+0.003$, $[-0.052, 0.066]$ (Holm-corrected Wilcoxon $p=0.43$ and $0.79$, not significant). \flowkl vs.\ Turn-count diff.: $+0.969$, $[0.835, 1.104]$; vs.\ Word-count diff.: $+0.681$, $[0.559, 0.809]$ ($p<0.001$ for both).

\section{Encoder Models}
\label{app:encoders}

The encoders used in Tasks 2 and 3 (and, for D2F, in \flowkl) are the following Hugging Face models:
\begin{itemize}[nosep]
\item SBERT: \url{https://huggingface.co/sentence-transformers/all-mpnet-base-v2}
\item LaBSE: \url{https://huggingface.co/sentence-transformers/LaBSE}
\item BGE-M3: \url{https://huggingface.co/BAAI/bge-m3}
\item Qwen3-Embedding: \url{https://huggingface.co/Qwen/Qwen3-Embedding-8B}
\item EmbeddingGemma: \url{https://huggingface.co/google/embeddinggemma-300m}
\item DSE: \url{https://huggingface.co/aws-ai/dse-bert-base}
\item D2F: \url{https://huggingface.co/sergioburdisso/dialog2flow-joint-dse-base}
\end{itemize}

\section{Task 2: Significance Tests}
\label{app:d2d_significance}

Table~\ref{tab:d2d_significance} reports, for each encoder, the paired comparison between \flowscore and every other method of Table~\ref{tab:d2d}, including the encoder-free controls (Random is omitted, since its expected $\tau$ is exactly 0). The unit is a conformant dialogue together with its four forks under one hallucination type ($N=2{,}134 \times 4 = 8{,}536$).

\begin{table*}[t]
\centering
\small
\setlength{\tabcolsep}{4pt}
\caption{Task 2 significance of \flowscore vs.\ each baseline, per encoder (both tests one-sided, \flowscore $>$ baseline). Columns: $N$, number of paired units (conformant dialogue $\times$ hallucination type); Mean $\Delta$, mean per-unit difference in Kendall $\tau$ (\flowscore $-$ baseline); 95\% CI, paired-bootstrap confidence interval of Mean $\Delta$ (10,000 resamples); $W$, Wilcoxon signed-rank statistic; $p_W$, its p-value; $t$, paired $t$-test statistic; $p_t$, its p-value. P-values are Holm-Bonferroni corrected across encoders within each test. Wilcoxon is the primary test; the $t$-test is shown for comparison. Bold marks significance ($\alpha=0.05$).}
\label{tab:d2d_significance}
\begin{tabular}{llrrlrlrl}
\toprule
Baseline & Model & $N$ & Mean $\Delta$ & 95\% CI & $W$ & $p_W$ (Holm) & $t$ & $p_t$ (Holm) \\
\midrule
Mean-turn emb. & LaBSE & 8536 & 0.089 & [0.083, 0.094] & 3,567,916 & \textbf{<0.001} & 32.94 & \textbf{<0.001} \\
Mean-turn emb. & Qwen3-Emb. & 8536 & 0.166 & [0.159, 0.173] & 7,337,009 & \textbf{<0.001} & 46.89 & \textbf{<0.001} \\
Mean-turn emb. & SBERT & 8536 & 0.120 & [0.114, 0.126] & 4,799,006 & \textbf{<0.001} & 41.42 & \textbf{<0.001} \\
Mean-turn emb. & BGE-M3 & 8536 & 0.103 & [0.098, 0.109] & 4,247,414 & \textbf{<0.001} & 36.03 & \textbf{<0.001} \\
Mean-turn emb. & D2F & 8536 & 0.079 & [0.074, 0.084] & 3,277,998 & \textbf{<0.001} & 30.18 & \textbf{<0.001} \\
Mean-turn emb. & DSE & 8536 & 0.098 & [0.092, 0.104] & 3,910,352 & \textbf{<0.001} & 33.80 & \textbf{<0.001} \\
Mean-turn emb. & Emb.Gemma & 8536 & 0.101 & [0.095, 0.106] & 4,223,326 & \textbf{<0.001} & 35.18 & \textbf{<0.001} \\
\midrule
Full-dialog emb. & LaBSE & 8536 & 0.168 & [0.162, 0.175] & 7,401,378 & \textbf{<0.001} & 49.06 & \textbf{<0.001} \\
Full-dialog emb. & Qwen3-Emb. & 8536 & 0.284 & [0.275, 0.293] & 13,806,538 & \textbf{<0.001} & 62.68 & \textbf{<0.001} \\
Full-dialog emb. & SBERT & 8536 & 0.199 & [0.191, 0.206] & 9,626,174 & \textbf{<0.001} & 52.17 & \textbf{<0.001} \\
Full-dialog emb. & BGE-M3 & 8536 & 0.127 & [0.121, 0.134] & 6,031,392 & \textbf{<0.001} & 38.47 & \textbf{<0.001} \\
Full-dialog emb. & D2F & 8536 & 0.078 & [0.072, 0.084] & 4,733,582 & \textbf{<0.001} & 25.67 & \textbf{<0.001} \\
Full-dialog emb. & DSE & 8536 & 0.120 & [0.114, 0.127] & 6,382,279 & \textbf{<0.001} & 36.57 & \textbf{<0.001} \\
Full-dialog emb. & Emb.Gemma & 8536 & 0.156 & [0.149, 0.163] & 7,144,218 & \textbf{<0.001} & 43.96 & \textbf{<0.001} \\
\midrule
FlowDTW & LaBSE & 8536 & 0.100 & [0.095, 0.106] & 3,409,838 & \textbf{<0.001} & 33.67 & \textbf{<0.001} \\
FlowDTW & Qwen3-Emb. & 8536 & 0.092 & [0.086, 0.098] & 3,727,200 & \textbf{<0.001} & 28.97 & \textbf{<0.001} \\
FlowDTW & SBERT & 8536 & 0.106 & [0.100, 0.112] & 3,463,974 & \textbf{<0.001} & 36.40 & \textbf{<0.001} \\
FlowDTW & BGE-M3 & 8536 & 0.123 & [0.117, 0.130] & 4,329,188 & \textbf{<0.001} & 37.82 & \textbf{<0.001} \\
FlowDTW & D2F & 8536 & 0.094 & [0.088, 0.100] & 3,291,462 & \textbf{<0.001} & 30.20 & \textbf{<0.001} \\
FlowDTW & DSE & 8536 & 0.104 & [0.098, 0.110] & 3,554,605 & \textbf{<0.001} & 34.05 & \textbf{<0.001} \\
FlowDTW & Emb.Gemma & 8536 & 0.119 & [0.113, 0.125] & 3,920,959 & \textbf{<0.001} & 36.29 & \textbf{<0.001} \\
\midrule
Turn-count diff. & LaBSE & 8536 & 0.719 & [0.709, 0.730] & 33,022,681 & \textbf{<0.001} & 137.41 & \textbf{<0.001} \\
Turn-count diff. & Qwen3-Emb. & 8536 & 0.687 & [0.677, 0.698] & 32,592,390 & \textbf{<0.001} & 128.21 & \textbf{<0.001} \\
Turn-count diff. & SBERT & 8536 & 0.721 & [0.711, 0.732] & 33,009,702 & \textbf{<0.001} & 137.39 & \textbf{<0.001} \\
Turn-count diff. & BGE-M3 & 8536 & 0.724 & [0.714, 0.734] & 33,094,104 & \textbf{<0.001} & 137.73 & \textbf{<0.001} \\
Turn-count diff. & D2F & 8536 & 0.716 & [0.705, 0.726] & 32,853,692 & \textbf{<0.001} & 136.53 & \textbf{<0.001} \\
Turn-count diff. & DSE & 8536 & 0.714 & [0.704, 0.725] & 32,923,879 & \textbf{<0.001} & 135.25 & \textbf{<0.001} \\
Turn-count diff. & Emb.Gemma & 8536 & 0.714 & [0.704, 0.725] & 32,876,962 & \textbf{<0.001} & 136.12 & \textbf{<0.001} \\
\midrule
Word-count diff. & LaBSE & 8536 & 0.613 & [0.602, 0.625] & 24,442,275 & \textbf{<0.001} & 105.26 & \textbf{<0.001} \\
Word-count diff. & Qwen3-Emb. & 8536 & 0.581 & [0.570, 0.593] & 23,908,176 & \textbf{<0.001} & 97.06 & \textbf{<0.001} \\
Word-count diff. & SBERT & 8536 & 0.615 & [0.603, 0.627] & 24,654,910 & \textbf{<0.001} & 104.71 & \textbf{<0.001} \\
Word-count diff. & BGE-M3 & 8536 & 0.618 & [0.606, 0.629] & 24,613,460 & \textbf{<0.001} & 105.76 & \textbf{<0.001} \\
Word-count diff. & D2F & 8536 & 0.610 & [0.598, 0.621] & 24,596,110 & \textbf{<0.001} & 104.74 & \textbf{<0.001} \\
Word-count diff. & DSE & 8536 & 0.608 & [0.597, 0.620] & 24,293,225 & \textbf{<0.001} & 103.35 & \textbf{<0.001} \\
Word-count diff. & Emb.Gemma & 8536 & 0.609 & [0.597, 0.620] & 24,306,022 & \textbf{<0.001} & 104.29 & \textbf{<0.001} \\
\midrule
Shared prefix & LaBSE & 8536 & 0.132 & [0.124, 0.141] & 10,596,601 & \textbf{<0.001} & 30.19 & \textbf{<0.001} \\
Shared prefix & Qwen3-Emb. & 8536 & 0.100 & [0.092, 0.109] & 10,250,472 & \textbf{<0.001} & 22.19 & \textbf{<0.001} \\
Shared prefix & SBERT & 8536 & 0.134 & [0.126, 0.143] & 10,470,056 & \textbf{<0.001} & 31.29 & \textbf{<0.001} \\
Shared prefix & BGE-M3 & 8536 & 0.137 & [0.128, 0.145] & 10,496,926 & \textbf{<0.001} & 31.75 & \textbf{<0.001} \\
Shared prefix & D2F & 8536 & 0.129 & [0.120, 0.138] & 10,602,733 & \textbf{<0.001} & 29.52 & \textbf{<0.001} \\
Shared prefix & DSE & 8536 & 0.128 & [0.119, 0.136] & 10,592,098 & \textbf{<0.001} & 28.98 & \textbf{<0.001} \\
Shared prefix & Emb.Gemma & 8536 & 0.128 & [0.119, 0.136] & 10,449,919 & \textbf{<0.001} & 29.41 & \textbf{<0.001} \\
\bottomrule
\end{tabular}
\end{table*}

\begin{sidewaystable*}[p]
\centering
\scriptsize
\setlength{\tabcolsep}{3pt}
\caption{Dialog-to-Dialog appendix results: Kendall's $\tau$ by task. Column headers use the task abbreviations defined in Table~\ref{tab:forkbench_task_details}. The top block contains the controls. \textbf{Bold}: best within each model; \underline{\textbf{bold+underline}}: best overall. Higher is better; shading: darker is better.}
\label{tab:d2d_appendix}
\resizebox{\textwidth}{!}{%
\begin{tabular}{llccccccccccccccccccccccccc}
\toprule
Model & Method & ase & as & bb & bfr & df & ds & hb & hs & hsr & ms & pp & pr & pb & ps & rb & rs & rib & ric & ris & sac & spl & td & trv & wth & Avg. $\uparrow$ \\
\midrule
\multicolumn{2}{l}{\textcolor{refgray}{Random}} & \textcolor{refgray}{0.000} & \textcolor{refgray}{0.000} & \textcolor{refgray}{0.000} & \textcolor{refgray}{0.000} & \textcolor{refgray}{0.000} & \textcolor{refgray}{0.000} & \textcolor{refgray}{0.000} & \textcolor{refgray}{0.000} & \textcolor{refgray}{0.000} & \textcolor{refgray}{0.000} & \textcolor{refgray}{0.000} & \textcolor{refgray}{0.000} & \textcolor{refgray}{0.000} & \textcolor{refgray}{0.000} & \textcolor{refgray}{0.000} & \textcolor{refgray}{0.000} & \textcolor{refgray}{0.000} & \textcolor{refgray}{0.000} & \textcolor{refgray}{0.000} & \textcolor{refgray}{0.000} & \textcolor{refgray}{0.000} & \textcolor{refgray}{0.000} & \textcolor{refgray}{0.000} & \textcolor{refgray}{0.000} & \textcolor{refgray}{0.000} \\
\multicolumn{2}{l}{\textcolor{refgray}{Turn-count diff.}} & \textcolor{refgray}{0.202} & \textcolor{refgray}{0.133} & \textcolor{refgray}{$-$0.031} & \textcolor{refgray}{$-$0.062} & \textcolor{refgray}{0.400} & \textcolor{refgray}{0.196} & \textcolor{refgray}{0.198} & \textcolor{refgray}{0.331} & \textcolor{refgray}{$-$0.094} & \textcolor{refgray}{0.011} & \textcolor{refgray}{$-$0.127} & \textcolor{refgray}{$-$0.026} & \textcolor{refgray}{0.038} & \textcolor{refgray}{0.101} & \textcolor{refgray}{0.037} & \textcolor{refgray}{0.396} & \textcolor{refgray}{0.182} & \textcolor{refgray}{$-$0.051} & \textcolor{refgray}{0.449} & \textcolor{refgray}{0.062} & \textcolor{refgray}{$-$0.062} & \textcolor{refgray}{0.140} & \textcolor{refgray}{0.260} & \textcolor{refgray}{0.346} & \textcolor{refgray}{0.126} \\
\multicolumn{2}{l}{\textcolor{refgray}{Word-count diff.}} & \textcolor{refgray}{0.238} & \textcolor{refgray}{0.019} & \textcolor{refgray}{0.131} & \textcolor{refgray}{0.246} & \textcolor{refgray}{0.274} & \textcolor{refgray}{0.329} & \textcolor{refgray}{0.265} & \textcolor{refgray}{0.243} & \textcolor{refgray}{0.016} & \textcolor{refgray}{0.368} & \textcolor{refgray}{0.190} & \textcolor{refgray}{0.260} & \textcolor{refgray}{0.055} & \textcolor{refgray}{0.381} & \textcolor{refgray}{0.099} & \textcolor{refgray}{0.239} & \textcolor{refgray}{0.263} & \textcolor{refgray}{0.042} & \textcolor{refgray}{0.435} & \textcolor{refgray}{0.247} & \textcolor{refgray}{0.226} & \textcolor{refgray}{0.156} & \textcolor{refgray}{0.254} & \textcolor{refgray}{0.372} & \textcolor{refgray}{0.223} \\
\multicolumn{2}{l}{\textcolor{refgray}{Shared prefix}} & \textcolor{refgray}{0.816} & \textcolor{refgray}{0.743} & \textcolor{refgray}{0.791} & \textcolor{refgray}{0.772} & \textcolor{refgray}{0.762} & \textcolor{refgray}{0.776} & \textcolor{refgray}{0.821} & \textcolor{refgray}{0.476} & \textcolor{refgray}{0.880} & \textcolor{refgray}{0.789} & \textcolor{refgray}{0.713} & \textcolor{refgray}{0.739} & \textcolor{refgray}{0.690} & \textcolor{refgray}{0.756} & \textcolor{refgray}{0.819} & \textcolor{refgray}{0.694} & \textcolor{refgray}{0.761} & \textcolor{refgray}{0.551} & \textcolor{refgray}{0.028} & \textcolor{refgray}{0.646} & \textcolor{refgray}{0.434} & \textcolor{refgray}{0.753} & \textcolor{refgray}{0.340} & \textcolor{refgray}{0.829} & \textcolor{refgray}{0.682} \\
\midrule
\multirow{4}{*}{LaBSE} & Mean-turn emb. & \cellcolor{heat!23}0.759 & \cellcolor{heat!23}0.762 & \cellcolor{heat!15}0.620 & \cellcolor{heat!19}0.690 & \cellcolor{heat!23}0.760 & \cellcolor{heat!25}0.784 & \cellcolor{heat!29}0.857 & \cellcolor{heat!20}0.707 & \cellcolor{heat!17}0.661 & \cellcolor{heat!26}0.813 & \cellcolor{heat!25}0.783 & \cellcolor{heat!24}0.782 & \cellcolor{heat!14}0.586 & \cellcolor{heat!29}0.848 & \cellcolor{heat!23}0.765 & \cellcolor{heat!24}0.777 & \cellcolor{heat!27}0.825 & \cellcolor{heat!29}0.846 & \cellcolor{heat!20}0.713 & \cellcolor{heat!27}0.827 & \cellcolor{heat!28}0.833 & \cellcolor{heat!20}0.710 & \cellcolor{heat!21}0.733 & \cellcolor{heat!23}0.765 & \cellcolor{heat!23}0.759 \\
 & Full-dialog emb. & \cellcolor{heat!18}0.667 & \cellcolor{heat!21}0.718 & \cellcolor{heat!12}0.553 & \cellcolor{heat!19}0.697 & \cellcolor{heat!16}0.625 & \cellcolor{heat!16}0.637 & \cellcolor{heat!24}0.770 & \cellcolor{heat!19}0.696 & \cellcolor{heat!14}0.583 & \cellcolor{heat!19}0.689 & \cellcolor{heat!19}0.684 & \cellcolor{heat!13}0.577 & \cellcolor{heat!16}0.630 & \cellcolor{heat!20}0.701 & \cellcolor{heat!17}0.648 & \cellcolor{heat!20}0.703 & \cellcolor{heat!19}0.698 & \cellcolor{heat!24}0.776 & \cellcolor{heat!16}0.630 & \cellcolor{heat!21}0.729 & \cellcolor{heat!20}0.715 & \cellcolor{heat!15}0.618 & \cellcolor{heat!22}0.749 & \cellcolor{heat!21}0.716 & \cellcolor{heat!18}0.675 \\
 & FlowDTW & \cellcolor{heat!26}0.801 & \cellcolor{heat!21}0.732 & \cellcolor{heat!15}0.620 & \cellcolor{heat!17}0.659 & \cellcolor{heat!21}0.718 & \cellcolor{heat!24}0.779 & \cellcolor{heat!29}0.845 & \cellcolor{heat!17}0.656 & \cellcolor{heat!15}0.611 & \cellcolor{heat!28}0.831 & \cellcolor{heat!19}0.684 & \cellcolor{heat!28}0.831 & \cellcolor{heat!8}0.453 & \cellcolor{heat!28}0.842 & \cellcolor{heat!23}0.765 & \cellcolor{heat!24}0.775 & \cellcolor{heat!27}0.826 & \cellcolor{heat!25}0.788 & \cellcolor{heat!19}0.685 & \cellcolor{heat!24}0.778 & \cellcolor{heat!29}0.846 & \cellcolor{heat!19}0.687 & \cellcolor{heat!26}0.802 & \cellcolor{heat!25}0.784 & \cellcolor{heat!22}0.742 \\
 & FlowScore & \cellcolor{heat!31}\textbf{0.881} & \cellcolor{heat!26}\textbf{0.807} & \cellcolor{heat!22}\textbf{0.750} & \cellcolor{heat!25}\textbf{0.790} & \cellcolor{heat!27}\textbf{0.823} & \cellcolor{heat!30}\textbf{0.870} & \cellcolor{heat!33}\textbf{0.906} & \cellcolor{heat!22}\textbf{0.750} & \cellcolor{heat!30}\textbf{0.859} & \cellcolor{heat!30}\textbf{0.868} & \cellcolor{heat!29}\textbf{0.854} & \cellcolor{heat!30}\underline{\textbf{0.862}} & \cellcolor{heat!24}\textbf{0.768} & \cellcolor{heat!33}\textbf{0.914} & \cellcolor{heat!30}\textbf{0.864} & \cellcolor{heat!30}\textbf{0.866} & \cellcolor{heat!31}\textbf{0.885} & \cellcolor{heat!30}\textbf{0.865} & \cellcolor{heat!23}\textbf{0.759} & \cellcolor{heat!29}\textbf{0.847} & \cellcolor{heat!30}\underline{\textbf{0.869}} & \cellcolor{heat!26}\textbf{0.811} & \cellcolor{heat!29}\textbf{0.854} & \cellcolor{heat!28}\textbf{0.834} & \cellcolor{heat!28}\textbf{0.840} \\
\midrule
\multirow{4}{*}{Qwen3-Embedding} & Mean-turn emb. & \cellcolor{heat!18}0.666 & \cellcolor{heat!19}0.682 & \cellcolor{heat!15}0.621 & \cellcolor{heat!18}0.671 & \cellcolor{heat!21}0.722 & \cellcolor{heat!12}0.543 & \cellcolor{heat!20}0.705 & \cellcolor{heat!16}0.630 & \cellcolor{heat!11}0.522 & \cellcolor{heat!15}0.610 & \cellcolor{heat!17}0.645 & \cellcolor{heat!13}0.561 & \cellcolor{heat!9}0.474 & \cellcolor{heat!22}0.741 & \cellcolor{heat!9}0.472 & \cellcolor{heat!19}0.688 & \cellcolor{heat!19}0.689 & \cellcolor{heat!21}0.724 & \cellcolor{heat!18}0.676 & \cellcolor{heat!24}0.782 & \cellcolor{heat!20}0.702 & \cellcolor{heat!15}0.620 & \cellcolor{heat!20}0.699 & \cellcolor{heat!20}0.709 & \cellcolor{heat!17}0.648 \\
 & Full-dialog emb. & \cellcolor{heat!7}0.428 & \cellcolor{heat!10}0.510 & \cellcolor{heat!8}0.449 & \cellcolor{heat!8}0.444 & \cellcolor{heat!11}0.530 & \cellcolor{heat!12}0.559 & \cellcolor{heat!11}0.536 & \cellcolor{heat!10}0.494 & \cellcolor{heat!5}0.346 & \cellcolor{heat!8}0.451 & \cellcolor{heat!13}0.564 & \cellcolor{heat!7}0.414 & \cellcolor{heat!11}0.524 & \cellcolor{heat!11}0.520 & \cellcolor{heat!8}0.438 & \cellcolor{heat!12}0.557 & \cellcolor{heat!16}0.641 & \cellcolor{heat!7}0.410 & \cellcolor{heat!14}0.593 & \cellcolor{heat!17}0.645 & \cellcolor{heat!16}0.633 & \cellcolor{heat!11}0.531 & \cellcolor{heat!16}0.629 & \cellcolor{heat!15}0.613 & \cellcolor{heat!11}0.519 \\
 & FlowDTW & \cellcolor{heat!23}0.766 & \cellcolor{heat!19}0.690 & \cellcolor{heat!16}0.627 & \cellcolor{heat!17}0.654 & \cellcolor{heat!19}0.688 & \cellcolor{heat!23}0.751 & \cellcolor{heat!27}0.821 & \cellcolor{heat!14}0.597 & \cellcolor{heat!16}0.623 & \cellcolor{heat!27}0.823 & \cellcolor{heat!20}0.703 & \cellcolor{heat!25}0.786 & \cellcolor{heat!10}0.491 & \cellcolor{heat!27}0.819 & \cellcolor{heat!23}0.762 & \cellcolor{heat!20}0.712 & \cellcolor{heat!23}0.763 & \cellcolor{heat!25}0.795 & \cellcolor{heat!18}0.667 & \cellcolor{heat!25}0.792 & \cellcolor{heat!28}\textbf{0.833} & \cellcolor{heat!17}0.658 & \cellcolor{heat!22}0.742 & \cellcolor{heat!23}0.752 & \cellcolor{heat!21}0.721 \\
 & FlowScore & \cellcolor{heat!30}\textbf{0.865} & \cellcolor{heat!25}\textbf{0.793} & \cellcolor{heat!22}\textbf{0.743} & \cellcolor{heat!23}\textbf{0.766} & \cellcolor{heat!26}\textbf{0.804} & \cellcolor{heat!28}\textbf{0.837} & \cellcolor{heat!31}\textbf{0.881} & \cellcolor{heat!21}\textbf{0.723} & \cellcolor{heat!29}\textbf{0.857} & \cellcolor{heat!29}\textbf{0.846} & \cellcolor{heat!28}\textbf{0.835} & \cellcolor{heat!26}\textbf{0.808} & \cellcolor{heat!19}\textbf{0.692} & \cellcolor{heat!33}\textbf{0.908} & \cellcolor{heat!27}\textbf{0.827} & \cellcolor{heat!26}\textbf{0.811} & \cellcolor{heat!28}\textbf{0.839} & \cellcolor{heat!30}\textbf{0.865} & \cellcolor{heat!22}\textbf{0.750} & \cellcolor{heat!28}\textbf{0.836} & \cellcolor{heat!27}0.829 & \cellcolor{heat!25}\textbf{0.787} & \cellcolor{heat!25}\textbf{0.792} & \cellcolor{heat!25}\textbf{0.788} & \cellcolor{heat!26}\textbf{0.812} \\
\midrule
\multirow{4}{*}{SBERT} & Mean-turn emb. & \cellcolor{heat!21}0.726 & \cellcolor{heat!20}0.699 & \cellcolor{heat!15}0.614 & \cellcolor{heat!19}0.681 & \cellcolor{heat!21}0.718 & \cellcolor{heat!23}0.751 & \cellcolor{heat!27}0.820 & \cellcolor{heat!17}0.648 & \cellcolor{heat!14}0.587 & \cellcolor{heat!23}0.762 & \cellcolor{heat!22}0.745 & \cellcolor{heat!22}0.736 & \cellcolor{heat!12}0.545 & \cellcolor{heat!27}0.816 & \cellcolor{heat!22}0.738 & \cellcolor{heat!22}0.737 & \cellcolor{heat!26}0.803 & \cellcolor{heat!29}0.846 & \cellcolor{heat!21}0.731 & \cellcolor{heat!28}0.832 & \cellcolor{heat!24}0.781 & \cellcolor{heat!19}0.691 & \cellcolor{heat!21}0.729 & \cellcolor{heat!22}0.740 & \cellcolor{heat!21}0.728 \\
 & Full-dialog emb. & \cellcolor{heat!19}0.693 & \cellcolor{heat!16}0.632 & \cellcolor{heat!9}0.471 & \cellcolor{heat!18}0.662 & \cellcolor{heat!13}0.579 & \cellcolor{heat!17}0.645 & \cellcolor{heat!23}0.763 & \cellcolor{heat!16}0.631 & \cellcolor{heat!8}0.437 & \cellcolor{heat!17}0.654 & \cellcolor{heat!20}0.701 & \cellcolor{heat!13}0.568 & \cellcolor{heat!9}0.474 & \cellcolor{heat!15}0.609 & \cellcolor{heat!13}0.562 & \cellcolor{heat!16}0.629 & \cellcolor{heat!18}0.669 & \cellcolor{heat!8}0.449 & \cellcolor{heat!14}0.602 & \cellcolor{heat!23}0.753 & \cellcolor{heat!24}0.779 & \cellcolor{heat!14}0.595 & \cellcolor{heat!21}0.725 & \cellcolor{heat!18}0.662 & \cellcolor{heat!16}0.623 \\
 & FlowDTW & \cellcolor{heat!26}0.799 & \cellcolor{heat!19}0.685 & \cellcolor{heat!18}0.670 & \cellcolor{heat!17}0.660 & \cellcolor{heat!20}0.712 & \cellcolor{heat!25}0.783 & \cellcolor{heat!29}0.852 & \cellcolor{heat!15}0.606 & \cellcolor{heat!16}0.632 & \cellcolor{heat!27}0.827 & \cellcolor{heat!20}0.700 & \cellcolor{heat!26}0.799 & \cellcolor{heat!9}0.474 & \cellcolor{heat!28}0.842 & \cellcolor{heat!23}0.756 & \cellcolor{heat!23}0.754 & \cellcolor{heat!27}0.819 & \cellcolor{heat!27}0.821 & \cellcolor{heat!18}0.676 & \cellcolor{heat!26}0.802 & \cellcolor{heat!26}0.812 & \cellcolor{heat!19}0.687 & \cellcolor{heat!24}0.780 & \cellcolor{heat!23}0.764 & \cellcolor{heat!22}0.738 \\
 & FlowScore & \cellcolor{heat!32}\textbf{0.891} & \cellcolor{heat!25}\textbf{0.785} & \cellcolor{heat!24}\underline{\textbf{0.781}} & \cellcolor{heat!25}\textbf{0.791} & \cellcolor{heat!28}\underline{\textbf{0.837}} & \cellcolor{heat!32}\textbf{0.889} & \cellcolor{heat!33}\textbf{0.906} & \cellcolor{heat!21}\textbf{0.730} & \cellcolor{heat!31}\textbf{0.875} & \cellcolor{heat!31}\underline{\textbf{0.884}} & \cellcolor{heat!29}\textbf{0.852} & \cellcolor{heat!29}\textbf{0.856} & \cellcolor{heat!22}\textbf{0.735} & \cellcolor{heat!35}\underline{\textbf{0.931}} & \cellcolor{heat!32}\underline{\textbf{0.889}} & \cellcolor{heat!28}\textbf{0.835} & \cellcolor{heat!32}\underline{\textbf{0.895}} & \cellcolor{heat!33}\underline{\textbf{0.904}} & \cellcolor{heat!24}\textbf{0.778} & \cellcolor{heat!29}\underline{\textbf{0.855}} & \cellcolor{heat!27}\textbf{0.825} & \cellcolor{heat!28}\underline{\textbf{0.832}} & \cellcolor{heat!28}\textbf{0.836} & \cellcolor{heat!28}\underline{\textbf{0.835}} & \cellcolor{heat!28}\textbf{0.843} \\
\midrule
\multirow{4}{*}{BGE-M3} & Mean-turn emb. & \cellcolor{heat!21}0.731 & \cellcolor{heat!23}0.751 & \cellcolor{heat!15}0.605 & \cellcolor{heat!19}0.689 & \cellcolor{heat!23}0.756 & \cellcolor{heat!24}0.772 & \cellcolor{heat!28}0.834 & \cellcolor{heat!20}0.713 & \cellcolor{heat!16}0.623 & \cellcolor{heat!26}0.807 & \cellcolor{heat!24}0.768 & \cellcolor{heat!23}0.760 & \cellcolor{heat!15}0.608 & \cellcolor{heat!29}0.848 & \cellcolor{heat!23}0.756 & \cellcolor{heat!21}0.723 & \cellcolor{heat!27}0.814 & \cellcolor{heat!29}0.853 & \cellcolor{heat!20}0.713 & \cellcolor{heat!27}\textbf{0.821} & \cellcolor{heat!26}0.800 & \cellcolor{heat!20}0.707 & \cellcolor{heat!22}0.749 & \cellcolor{heat!22}0.744 & \cellcolor{heat!22}0.748 \\
 & Full-dialog emb. & \cellcolor{heat!25}0.792 & \cellcolor{heat!23}0.751 & \cellcolor{heat!13}0.567 & \cellcolor{heat!22}0.744 & \cellcolor{heat!17}0.647 & \cellcolor{heat!19}0.684 & \cellcolor{heat!28}0.833 & \cellcolor{heat!20}0.704 & \cellcolor{heat!19}0.685 & \cellcolor{heat!25}0.793 & \cellcolor{heat!22}0.735 & \cellcolor{heat!22}0.744 & \cellcolor{heat!14}0.602 & \cellcolor{heat!19}0.684 & \cellcolor{heat!22}0.744 & \cellcolor{heat!21}0.731 & \cellcolor{heat!23}0.752 & \cellcolor{heat!20}0.705 & \cellcolor{heat!15}0.620 & \cellcolor{heat!26}0.802 & \cellcolor{heat!22}0.737 & \cellcolor{heat!19}0.682 & \cellcolor{heat!21}0.721 & \cellcolor{heat!15}0.609 & \cellcolor{heat!20}0.711 \\
 & FlowDTW & \cellcolor{heat!23}0.766 & \cellcolor{heat!20}0.707 & \cellcolor{heat!16}0.641 & \cellcolor{heat!17}0.646 & \cellcolor{heat!21}0.720 & \cellcolor{heat!23}0.757 & \cellcolor{heat!28}0.842 & \cellcolor{heat!13}0.579 & \cellcolor{heat!14}0.591 & \cellcolor{heat!27}0.823 & \cellcolor{heat!18}0.673 & \cellcolor{heat!25}0.793 & \cellcolor{heat!11}0.517 & \cellcolor{heat!27}0.816 & \cellcolor{heat!23}0.765 & \cellcolor{heat!19}0.684 & \cellcolor{heat!27}0.827 & \cellcolor{heat!27}0.821 & \cellcolor{heat!18}0.676 & \cellcolor{heat!23}0.766 & \cellcolor{heat!28}0.835 & \cellcolor{heat!15}0.614 & \cellcolor{heat!27}0.815 & \cellcolor{heat!24}0.779 & \cellcolor{heat!21}0.727 \\
 & FlowScore & \cellcolor{heat!31}\textbf{0.880} & \cellcolor{heat!26}\textbf{0.801} & \cellcolor{heat!24}\textbf{0.770} & \cellcolor{heat!26}\textbf{0.801} & \cellcolor{heat!27}\textbf{0.827} & \cellcolor{heat!32}\underline{\textbf{0.891}} & \cellcolor{heat!33}\underline{\textbf{0.915}} & \cellcolor{heat!23}\underline{\textbf{0.766}} & \cellcolor{heat!30}\textbf{0.871} & \cellcolor{heat!31}\textbf{0.874} & \cellcolor{heat!29}\textbf{0.856} & \cellcolor{heat!29}\textbf{0.850} & \cellcolor{heat!25}\underline{\textbf{0.792}} & \cellcolor{heat!34}\textbf{0.922} & \cellcolor{heat!31}\textbf{0.880} & \cellcolor{heat!30}\underline{\textbf{0.867}} & \cellcolor{heat!32}\textbf{0.894} & \cellcolor{heat!30}\textbf{0.865} & \cellcolor{heat!26}\underline{\textbf{0.806}} & \cellcolor{heat!26}0.807 & \cellcolor{heat!28}\textbf{0.844} & \cellcolor{heat!27}\textbf{0.828} & \cellcolor{heat!28}\textbf{0.843} & \cellcolor{heat!28}\textbf{0.832} & \cellcolor{heat!29}\underline{\textbf{0.845}} \\
\midrule
\multirow{4}{*}{D2F} & Mean-turn emb. & \cellcolor{heat!24}0.782 & \cellcolor{heat!23}0.757 & \cellcolor{heat!15}0.621 & \cellcolor{heat!21}0.719 & \cellcolor{heat!21}0.720 & \cellcolor{heat!25}0.795 & \cellcolor{heat!28}0.830 & \cellcolor{heat!21}0.733 & \cellcolor{heat!21}0.726 & \cellcolor{heat!28}0.835 & \cellcolor{heat!26}0.807 & \cellcolor{heat!25}0.783 & \cellcolor{heat!15}0.609 & \cellcolor{heat!28}0.836 & \cellcolor{heat!25}0.790 & \cellcolor{heat!20}0.714 & \cellcolor{heat!26}0.799 & \cellcolor{heat!29}0.853 & \cellcolor{heat!21}0.722 & \cellcolor{heat!26}0.812 & \cellcolor{heat!26}0.810 & \cellcolor{heat!20}0.707 & \cellcolor{heat!23}0.751 & \cellcolor{heat!24}0.767 & \cellcolor{heat!23}0.762 \\
 & Full-dialog emb. & \cellcolor{heat!26}0.811 & \cellcolor{heat!24}0.774 & \cellcolor{heat!19}0.694 & \cellcolor{heat!22}0.743 & \cellcolor{heat!16}0.639 & \cellcolor{heat!22}0.735 & \cellcolor{heat!29}0.855 & \cellcolor{heat!18}0.666 & \cellcolor{heat!23}0.752 & \cellcolor{heat!26}0.807 & \cellcolor{heat!27}0.829 & \cellcolor{heat!28}0.831 & \cellcolor{heat!17}0.647 & \cellcolor{heat!20}0.710 & \cellcolor{heat!25}0.787 & \cellcolor{heat!21}0.716 & \cellcolor{heat!26}0.808 & \cellcolor{heat!22}0.744 & \cellcolor{heat!18}0.676 & \cellcolor{heat!21}0.730 & \cellcolor{heat!24}0.777 & \cellcolor{heat!21}0.723 & \cellcolor{heat!27}0.823 & \cellcolor{heat!17}0.644 & \cellcolor{heat!22}0.747 \\
 & FlowDTW & \cellcolor{heat!27}0.820 & \cellcolor{heat!21}0.726 & \cellcolor{heat!17}0.659 & \cellcolor{heat!16}0.635 & \cellcolor{heat!18}0.663 & \cellcolor{heat!25}0.790 & \cellcolor{heat!28}0.837 & \cellcolor{heat!19}0.689 & \cellcolor{heat!15}0.605 & \cellcolor{heat!29}0.850 & \cellcolor{heat!20}0.705 & \cellcolor{heat!28}0.831 & \cellcolor{heat!8}0.440 & \cellcolor{heat!27}0.822 & \cellcolor{heat!24}0.775 & \cellcolor{heat!24}0.778 & \cellcolor{heat!26}0.799 & \cellcolor{heat!24}0.776 & \cellcolor{heat!19}0.694 & \cellcolor{heat!25}0.788 & \cellcolor{heat!27}0.825 & \cellcolor{heat!20}0.703 & \cellcolor{heat!27}0.814 & \cellcolor{heat!24}0.779 & \cellcolor{heat!22}0.742 \\
 & FlowScore & \cellcolor{heat!32}\textbf{0.891} & \cellcolor{heat!25}\textbf{0.785} & \cellcolor{heat!23}\textbf{0.754} & \cellcolor{heat!26}\underline{\textbf{0.804}} & \cellcolor{heat!26}\textbf{0.800} & \cellcolor{heat!30}\textbf{0.865} & \cellcolor{heat!31}\textbf{0.876} & \cellcolor{heat!23}\textbf{0.754} & \cellcolor{heat!31}\textbf{0.877} & \cellcolor{heat!30}\textbf{0.870} & \cellcolor{heat!30}\underline{\textbf{0.871}} & \cellcolor{heat!28}\textbf{0.833} & \cellcolor{heat!22}\textbf{0.739} & \cellcolor{heat!33}\textbf{0.902} & \cellcolor{heat!29}\textbf{0.849} & \cellcolor{heat!29}\textbf{0.848} & \cellcolor{heat!31}\textbf{0.881} & \cellcolor{heat!31}\textbf{0.885} & \cellcolor{heat!22}\textbf{0.750} & \cellcolor{heat!28}\textbf{0.841} & \cellcolor{heat!28}\textbf{0.840} & \cellcolor{heat!27}\textbf{0.814} & \cellcolor{heat!29}\underline{\textbf{0.857}} & \cellcolor{heat!27}\textbf{0.822} & \cellcolor{heat!28}\textbf{0.834} \\
\midrule
\multirow{4}{*}{DSE} & Mean-turn emb. & \cellcolor{heat!23}0.757 & \cellcolor{heat!21}0.724 & \cellcolor{heat!14}0.596 & \cellcolor{heat!19}0.681 & \cellcolor{heat!20}0.712 & \cellcolor{heat!23}0.759 & \cellcolor{heat!28}0.840 & \cellcolor{heat!18}0.671 & \cellcolor{heat!18}0.676 & \cellcolor{heat!26}0.809 & \cellcolor{heat!24}0.777 & \cellcolor{heat!24}0.771 & \cellcolor{heat!12}0.542 & \cellcolor{heat!30}0.859 & \cellcolor{heat!23}0.759 & \cellcolor{heat!22}0.737 & \cellcolor{heat!25}0.794 & \cellcolor{heat!27}0.814 & \cellcolor{heat!19}0.694 & \cellcolor{heat!26}0.811 & \cellcolor{heat!27}0.825 & \cellcolor{heat!19}0.697 & \cellcolor{heat!23}0.762 & \cellcolor{heat!21}0.723 & \cellcolor{heat!22}0.741 \\
 & Full-dialog emb. & \cellcolor{heat!21}0.725 & \cellcolor{heat!28}\underline{\textbf{0.832}} & \cellcolor{heat!15}0.621 & \cellcolor{heat!19}0.693 & \cellcolor{heat!16}0.623 & \cellcolor{heat!19}0.687 & \cellcolor{heat!26}0.803 & \cellcolor{heat!22}\textbf{0.734} & \cellcolor{heat!15}0.614 & \cellcolor{heat!22}0.740 & \cellcolor{heat!26}0.801 & \cellcolor{heat!20}0.701 & \cellcolor{heat!10}0.510 & \cellcolor{heat!25}0.793 & \cellcolor{heat!18}0.670 & \cellcolor{heat!21}0.725 & \cellcolor{heat!24}0.774 & \cellcolor{heat!11}0.526 & \cellcolor{heat!17}0.657 & \cellcolor{heat!21}0.725 & \cellcolor{heat!25}0.792 & \cellcolor{heat!21}0.720 & \cellcolor{heat!21}0.725 & \cellcolor{heat!20}0.709 & \cellcolor{heat!20}0.704 \\
 & FlowDTW & \cellcolor{heat!25}0.788 & \cellcolor{heat!19}0.696 & \cellcolor{heat!16}0.632 & \cellcolor{heat!16}0.634 & \cellcolor{heat!20}0.706 & \cellcolor{heat!24}0.775 & \cellcolor{heat!29}0.845 & \cellcolor{heat!15}0.607 & \cellcolor{heat!16}0.629 & \cellcolor{heat!29}0.850 & \cellcolor{heat!18}0.677 & \cellcolor{heat!26}0.807 & \cellcolor{heat!7}0.419 & \cellcolor{heat!29}0.848 & \cellcolor{heat!24}0.781 & \cellcolor{heat!22}0.737 & \cellcolor{heat!28}0.834 & \cellcolor{heat!27}0.814 & \cellcolor{heat!21}\textbf{0.722} & \cellcolor{heat!26}0.801 & \cellcolor{heat!28}0.844 & \cellcolor{heat!20}0.713 & \cellcolor{heat!24}0.782 & \cellcolor{heat!22}0.746 & \cellcolor{heat!22}0.737 \\
 & FlowScore & \cellcolor{heat!32}\underline{\textbf{0.893}} & \cellcolor{heat!25}0.787 & \cellcolor{heat!21}\textbf{0.730} & \cellcolor{heat!25}\textbf{0.788} & \cellcolor{heat!27}\textbf{0.829} & \cellcolor{heat!30}\textbf{0.863} & \cellcolor{heat!33}\textbf{0.902} & \cellcolor{heat!21}0.733 & \cellcolor{heat!32}\underline{\textbf{0.888}} & \cellcolor{heat!31}\textbf{0.882} & \cellcolor{heat!30}\textbf{0.859} & \cellcolor{heat!29}\textbf{0.846} & \cellcolor{heat!21}\textbf{0.733} & \cellcolor{heat!34}\textbf{0.917} & \cellcolor{heat!30}\textbf{0.861} & \cellcolor{heat!30}\textbf{0.862} & \cellcolor{heat!32}\underline{\textbf{0.895}} & \cellcolor{heat!31}\textbf{0.885} & \cellcolor{heat!19}0.694 & \cellcolor{heat!28}\textbf{0.830} & \cellcolor{heat!29}\textbf{0.854} & \cellcolor{heat!27}\textbf{0.816} & \cellcolor{heat!28}\textbf{0.838} & \cellcolor{heat!26}\textbf{0.800} & \cellcolor{heat!28}\textbf{0.833} \\
\midrule
\multirow{4}{*}{EmbeddingGemma} & Mean-turn emb. & \cellcolor{heat!21}0.733 & \cellcolor{heat!24}0.768 & \cellcolor{heat!13}0.576 & \cellcolor{heat!19}0.693 & \cellcolor{heat!20}0.712 & \cellcolor{heat!24}0.771 & \cellcolor{heat!27}0.824 & \cellcolor{heat!20}0.701 & \cellcolor{heat!15}0.611 & \cellcolor{heat!25}0.793 & \cellcolor{heat!23}0.764 & \cellcolor{heat!23}0.751 & \cellcolor{heat!14}0.581 & \cellcolor{heat!27}0.825 & \cellcolor{heat!23}0.759 & \cellcolor{heat!22}0.748 & \cellcolor{heat!26}0.805 & \cellcolor{heat!27}0.827 & \cellcolor{heat!19}0.694 & \cellcolor{heat!26}0.802 & \cellcolor{heat!28}0.831 & \cellcolor{heat!20}0.703 & \cellcolor{heat!23}0.756 & \cellcolor{heat!21}0.720 & \cellcolor{heat!22}0.740 \\
 & Full-dialog emb. & \cellcolor{heat!20}0.714 & \cellcolor{heat!24}0.768 & \cellcolor{heat!13}0.578 & \cellcolor{heat!20}0.701 & \cellcolor{heat!16}0.623 & \cellcolor{heat!13}0.562 & \cellcolor{heat!22}0.746 & \cellcolor{heat!23}\textbf{0.753} & \cellcolor{heat!12}0.551 & \cellcolor{heat!20}0.707 & \cellcolor{heat!22}0.739 & \cellcolor{heat!16}0.625 & \cellcolor{heat!13}0.580 & \cellcolor{heat!23}0.764 & \cellcolor{heat!19}0.694 & \cellcolor{heat!21}0.729 & \cellcolor{heat!20}0.704 & \cellcolor{heat!17}0.660 & \cellcolor{heat!13}0.565 & \cellcolor{heat!21}0.731 & \cellcolor{heat!18}0.677 & \cellcolor{heat!19}0.689 & \cellcolor{heat!18}0.679 & \cellcolor{heat!18}0.673 & \cellcolor{heat!18}0.675 \\
 & FlowDTW & \cellcolor{heat!24}0.782 & \cellcolor{heat!18}0.676 & \cellcolor{heat!16}0.630 & \cellcolor{heat!15}0.615 & \cellcolor{heat!20}0.700 & \cellcolor{heat!22}0.748 & \cellcolor{heat!27}0.828 & \cellcolor{heat!15}0.618 & \cellcolor{heat!14}0.594 & \cellcolor{heat!27}0.817 & \cellcolor{heat!18}0.668 & \cellcolor{heat!25}0.798 & \cellcolor{heat!7}0.412 & \cellcolor{heat!27}0.825 & \cellcolor{heat!23}0.756 & \cellcolor{heat!22}0.739 & \cellcolor{heat!27}0.816 & \cellcolor{heat!25}0.795 & \cellcolor{heat!19}0.685 & \cellcolor{heat!24}0.769 & \cellcolor{heat!27}0.821 & \cellcolor{heat!17}0.657 & \cellcolor{heat!25}0.789 & \cellcolor{heat!23}0.757 & \cellcolor{heat!21}0.721 \\
 & FlowScore & \cellcolor{heat!31}\textbf{0.879} & \cellcolor{heat!24}\textbf{0.776} & \cellcolor{heat!22}\textbf{0.741} & \cellcolor{heat!25}\textbf{0.785} & \cellcolor{heat!26}\textbf{0.804} & \cellcolor{heat!32}\underline{\textbf{0.891}} & \cellcolor{heat!33}\textbf{0.907} & \cellcolor{heat!22}0.739 & \cellcolor{heat!31}\textbf{0.875} & \cellcolor{heat!30}\textbf{0.866} & \cellcolor{heat!29}\textbf{0.846} & \cellcolor{heat!28}\textbf{0.837} & \cellcolor{heat!23}\textbf{0.759} & \cellcolor{heat!34}\textbf{0.925} & \cellcolor{heat!30}\textbf{0.867} & \cellcolor{heat!28}\textbf{0.841} & \cellcolor{heat!31}\textbf{0.881} & \cellcolor{heat!30}\textbf{0.872} & \cellcolor{heat!25}\textbf{0.787} & \cellcolor{heat!27}\textbf{0.826} & \cellcolor{heat!29}\textbf{0.850} & \cellcolor{heat!27}\textbf{0.827} & \cellcolor{heat!28}\textbf{0.842} & \cellcolor{heat!26}\textbf{0.812} & \cellcolor{heat!28}\textbf{0.835} \\
\bottomrule
\end{tabular}%
}
\end{sidewaystable*}

\section{Streaming Methods: Encoder Ablation}
\label{app:turn_streaming}

Table~\ref{tab:tp_online_grouped} reports streaming results grouped by encoder.
Compact method labels: \SS = Surprisal Spike; \IT = Improbable Transition; indented \emph{w/ \NC} rows add node-confidence weighting to the method above. As in Table~\ref{tab:tp_main}, each detector learns from the train dialogues and is scored on the test dialogues. Best values within each encoder are bold; best overall are bold+underline. LaBSE achieves the best Score ($73.69$, \SS; reported in Table~\ref{tab:tp_main}), followed by D2F ($73.28$, \IT{}+\NC), Qwen3-Embedding ($72.06$, \IT) and EmbeddingGemma ($71.93$, \IT{}+\NC). The top encoders are close: in a paired bootstrap over the test conformant dialogues, each resampled with its forks (1,000 resamples), LaBSE scores significantly higher than SBERT, DSE and BGE-M3 ($p < 0.001$), but not than D2F, Qwen3-Embedding or EmbeddingGemma ($p \geq 0.11$).

\begin{table*}[t]
\centering
\small
\setlength{\tabcolsep}{3pt}
\caption{Streaming results on the Task~3 test set of the 3-gram LM detectors when using each embedding model to cluster turns into nodes. \SS = Surprisal Spike; \IT = Improbable Transition; indented \emph{w/ \NC} rows add node-confidence weighting to the method above. Best values within each embedding model are bold; best overall (non-baseline) are bold+underline; shading: darker is better.}
\label{tab:tp_online_grouped}
\begin{tabular}{llccccccccc}
\toprule
& & \multicolumn{4}{c}{Hallucination MAE $\downarrow$} & \multicolumn{2}{c}{Hallucination} & \multicolumn{2}{c}{Conformance} & \\
\cmidrule(lr){3-6}\cmidrule(lr){7-8}\cmidrule(lr){9-10}
Embedding & Method & \CSC & \GR & \SSPC & \WIC & Avg. $\downarrow$ & Cov. $\uparrow$ & MAE $\downarrow$ & Cov. $\downarrow$ & Score $\uparrow$ \\
\midrule
\multirow{4}{*}{LaBSE}
 & Surprisal Spike & \cellcolor{heat!36}4.21 & \cellcolor{heat!34}1.56 & \cellcolor{heat!39}1.30 & \cellcolor{heat!35}1.68 & \cellcolor{heat!37}2.19 & \cellcolor{heat!34}93.24 & \cellcolor{heat!15}\textbf{6.33} & \cellcolor{heat!35}\textbf{12.02} & \cellcolor{heat!31}\underline{\textbf{73.69}} \\
 & \quad w/ \NC & \cellcolor{heat!37}4.13 & \cellcolor{heat!33}1.61 & \cellcolor{heat!38}1.33 & \cellcolor{heat!34}1.71 & \cellcolor{heat!37}2.20 & \cellcolor{heat!35}93.56 & \cellcolor{heat!12}6.58 & \cellcolor{heat!34}15.47 & \cellcolor{heat!27}72.45 \\
 & Improbable Transition & \cellcolor{heat!37}4.13 & \cellcolor{heat!38}0.93 & \cellcolor{heat!39}\underline{\textbf{1.25}} & \cellcolor{heat!35}1.63 & \cellcolor{heat!39}1.98 & \cellcolor{heat!35}93.86 & \cellcolor{heat!13}6.52 & \cellcolor{heat!35}12.49 & \cellcolor{heat!31}73.60 \\
 & \quad w/ \NC & \cellcolor{heat!38}\textbf{3.94} & \cellcolor{heat!38}\underline{\textbf{0.92}} & \cellcolor{heat!39}1.29 & \cellcolor{heat!36}\underline{\textbf{1.60}} & \cellcolor{heat!39}\underline{\textbf{1.94}} & \cellcolor{heat!37}\textbf{94.80} & \cellcolor{heat!12}6.55 & \cellcolor{heat!33}16.31 & \cellcolor{heat!30}73.23 \\
\midrule
\multirow{4}{*}{D2F}
 & Surprisal Spike & \cellcolor{heat!33}4.82 & \cellcolor{heat!31}1.99 & \cellcolor{heat!34}1.70 & \cellcolor{heat!25}2.42 & \cellcolor{heat!32}2.73 & \cellcolor{heat!30}90.92 & \cellcolor{heat!10}6.72 & \cellcolor{heat!36}\textbf{10.81} & \cellcolor{heat!23}70.91 \\
 & \quad w/ \NC & \cellcolor{heat!34}4.57 & \cellcolor{heat!32}1.91 & \cellcolor{heat!34}1.75 & \cellcolor{heat!25}2.41 & \cellcolor{heat!33}2.66 & \cellcolor{heat!31}91.17 & \cellcolor{heat!17}\textbf{6.20} & \cellcolor{heat!34}14.54 & \cellcolor{heat!27}72.36 \\
 & Improbable Transition & \cellcolor{heat!33}4.69 & \cellcolor{heat!35}1.32 & \cellcolor{heat!36}\textbf{1.57} & \cellcolor{heat!28}2.19 & \cellcolor{heat!35}2.44 & \cellcolor{heat!32}91.75 & \cellcolor{heat!11}6.68 & \cellcolor{heat!36}11.18 & \cellcolor{heat!26}71.84 \\
 & \quad w/ \NC & \cellcolor{heat!36}\textbf{4.31} & \cellcolor{heat!35}\textbf{1.27} & \cellcolor{heat!36}1.58 & \cellcolor{heat!29}\textbf{2.11} & \cellcolor{heat!36}\textbf{2.32} & \cellcolor{heat!33}\textbf{92.52} & \cellcolor{heat!17}6.22 & \cellcolor{heat!34}14.91 & \cellcolor{heat!30}\textbf{73.28} \\
\midrule
\multirow{4}{*}{BGE-M3}
 & Surprisal Spike & \cellcolor{heat!39}3.72 & \cellcolor{heat!32}1.89 & \cellcolor{heat!35}1.60 & \cellcolor{heat!30}2.06 & \cellcolor{heat!36}2.32 & \cellcolor{heat!37}95.19 & \cellcolor{heat!4}\textbf{7.22} & \cellcolor{heat!20}\textbf{42.87} & \cellcolor{heat!10}66.11 \\
 & \quad w/ \NC & \cellcolor{heat!39}3.73 & \cellcolor{heat!31}1.95 & \cellcolor{heat!34}1.70 & \cellcolor{heat!30}2.05 & \cellcolor{heat!35}2.36 & \cellcolor{heat!38}95.46 & \cellcolor{heat!1}7.47 & \cellcolor{heat!19}43.43 & \cellcolor{heat!8}65.13 \\
 & Improbable Transition & \cellcolor{heat!40}3.68 & \cellcolor{heat!35}\textbf{1.35} & \cellcolor{heat!36}\textbf{1.57} & \cellcolor{heat!31}1.93 & \cellcolor{heat!37}\textbf{2.13} & \cellcolor{heat!38}95.45 & \cellcolor{heat!3}7.28 & \cellcolor{heat!19}43.90 & \cellcolor{heat!11}\textbf{66.20} \\
 & \quad w/ \NC & \cellcolor{heat!40}\underline{\textbf{3.62}} & \cellcolor{heat!35}1.41 & \cellcolor{heat!35}1.66 & \cellcolor{heat!32}\textbf{1.91} & \cellcolor{heat!37}2.15 & \cellcolor{heat!39}\underline{\textbf{96.00}} & 7.53 & \cellcolor{heat!18}45.76 & \cellcolor{heat!7}65.03 \\
\midrule
\multirow{4}{*}{SBERT}
 & Surprisal Spike & \cellcolor{heat!34}4.54 & \cellcolor{heat!30}2.21 & \cellcolor{heat!29}2.17 & \cellcolor{heat!31}1.97 & \cellcolor{heat!32}2.72 & \cellcolor{heat!32}91.99 & \cellcolor{heat!8}6.92 & \cellcolor{heat!33}\textbf{16.31} & \cellcolor{heat!20}69.73 \\
 & \quad w/ \NC & \cellcolor{heat!34}4.57 & \cellcolor{heat!29}2.29 & \cellcolor{heat!29}2.20 & \cellcolor{heat!30}2.06 & \cellcolor{heat!32}2.78 & \cellcolor{heat!32}91.99 & \cellcolor{heat!7}6.96 & \cellcolor{heat!32}19.48 & \cellcolor{heat!18}69.04 \\
 & Improbable Transition & \cellcolor{heat!34}4.52 & \cellcolor{heat!33}\textbf{1.61} & \cellcolor{heat!30}2.04 & \cellcolor{heat!32}\textbf{1.87} & \cellcolor{heat!34}2.51 & \cellcolor{heat!32}92.18 & \cellcolor{heat!9}\textbf{6.81} & \cellcolor{heat!33}16.50 & \cellcolor{heat!22}\textbf{70.61} \\
 & \quad w/ \NC & \cellcolor{heat!35}\textbf{4.40} & \cellcolor{heat!33}1.69 & \cellcolor{heat!31}\textbf{2.03} & \cellcolor{heat!32}\textbf{1.87} & \cellcolor{heat!34}\textbf{2.50} & \cellcolor{heat!33}\textbf{92.61} & \cellcolor{heat!9}6.83 & \cellcolor{heat!31}20.04 & \cellcolor{heat!21}70.15 \\
\midrule
\multirow{4}{*}{Qwen3-Embedding}
 & Surprisal Spike & \cellcolor{heat!33}4.78 & \cellcolor{heat!31}1.94 & \cellcolor{heat!33}1.80 & \cellcolor{heat!31}2.00 & \cellcolor{heat!33}2.63 & \cellcolor{heat!29}90.41 & \cellcolor{heat!14}6.45 & \cellcolor{heat!35}\textbf{13.70} & \cellcolor{heat!25}71.66 \\
 & \quad w/ \NC & \cellcolor{heat!33}4.68 & \cellcolor{heat!31}1.98 & \cellcolor{heat!33}1.84 & \cellcolor{heat!30}2.04 & \cellcolor{heat!33}2.63 & \cellcolor{heat!30}90.95 & \cellcolor{heat!9}6.80 & \cellcolor{heat!33}16.96 & \cellcolor{heat!21}70.13 \\
 & Improbable Transition & \cellcolor{heat!33}4.70 & \cellcolor{heat!35}1.42 & \cellcolor{heat!34}\textbf{1.76} & \cellcolor{heat!32}\textbf{1.89} & \cellcolor{heat!35}2.44 & \cellcolor{heat!30}90.91 & \cellcolor{heat!14}\textbf{6.44} & \cellcolor{heat!34}15.00 & \cellcolor{heat!26}\textbf{72.06} \\
 & \quad w/ \NC & \cellcolor{heat!35}\textbf{4.50} & \cellcolor{heat!35}\textbf{1.39} & \cellcolor{heat!33}1.79 & \cellcolor{heat!32}1.92 & \cellcolor{heat!35}\textbf{2.40} & \cellcolor{heat!31}\textbf{91.55} & \cellcolor{heat!13}6.54 & \cellcolor{heat!32}17.99 & \cellcolor{heat!25}71.53 \\
\midrule
\multirow{4}{*}{EmbeddingGemma}
 & Surprisal Spike & \cellcolor{heat!26}5.84 & \cellcolor{heat!22}3.56 & \cellcolor{heat!27}2.38 & \cellcolor{heat!10}3.56 & \cellcolor{heat!23}3.83 & \cellcolor{heat!27}88.62 & \cellcolor{heat!20}5.94 & \cellcolor{heat!37}8.39 & \cellcolor{heat!22}70.45 \\
 & \quad w/ \NC & \cellcolor{heat!28}5.62 & \cellcolor{heat!22}3.51 & \cellcolor{heat!27}2.37 & \cellcolor{heat!11}3.50 & \cellcolor{heat!23}3.75 & \cellcolor{heat!27}89.19 & \cellcolor{heat!24}\underline{\textbf{5.65}} & \cellcolor{heat!35}12.12 & \cellcolor{heat!24}71.18 \\
 & Improbable Transition & \cellcolor{heat!27}5.76 & \cellcolor{heat!25}3.00 & \cellcolor{heat!27}2.33 & \cellcolor{heat!12}3.39 & \cellcolor{heat!25}3.62 & \cellcolor{heat!27}89.13 & \cellcolor{heat!20}5.95 & \cellcolor{heat!37}\textbf{8.29} & \cellcolor{heat!24}71.08 \\
 & \quad w/ \NC & \cellcolor{heat!29}\textbf{5.37} & \cellcolor{heat!26}\textbf{2.86} & \cellcolor{heat!28}\textbf{2.25} & \cellcolor{heat!14}\textbf{3.27} & \cellcolor{heat!26}\textbf{3.44} & \cellcolor{heat!29}\textbf{90.29} & \cellcolor{heat!23}5.72 & \cellcolor{heat!35}12.58 & \cellcolor{heat!26}\textbf{71.93} \\
\midrule
\multirow{4}{*}{DSE}
 & Surprisal Spike & \cellcolor{heat!21}6.65 & \cellcolor{heat!6}6.12 & \cellcolor{heat!6}4.23 & \cellcolor{heat!4}4.01 & \cellcolor{heat!11}5.25 & \cellcolor{heat!2}73.43 & \cellcolor{heat!19}6.03 & \cellcolor{heat!40}\underline{\textbf{3.08}} & \cellcolor{heat!6}64.64 \\
 & \quad w/ \NC & \cellcolor{heat!24}6.23 & \cellcolor{heat!9}5.67 & \cellcolor{heat!9}3.97 & \cellcolor{heat!8}3.76 & \cellcolor{heat!13}4.91 & \cellcolor{heat!5}75.52 & \cellcolor{heat!9}6.81 & \cellcolor{heat!37}8.67 & \cellcolor{heat!2}63.12 \\
 & Improbable Transition & \cellcolor{heat!21}6.68 & \cellcolor{heat!12}5.20 & \cellcolor{heat!7}4.16 & \cellcolor{heat!6}3.89 & \cellcolor{heat!13}4.98 & \cellcolor{heat!3}74.22 & \cellcolor{heat!22}\textbf{5.84} & \cellcolor{heat!40}3.82 & \cellcolor{heat!10}\textbf{66.02} \\
 & \quad w/ \NC & \cellcolor{heat!25}\textbf{6.01} & \cellcolor{heat!14}\textbf{4.73} & \cellcolor{heat!10}\textbf{3.84} & \cellcolor{heat!10}\textbf{3.59} & \cellcolor{heat!17}\textbf{4.54} & \cellcolor{heat!7}\textbf{76.66} & \cellcolor{heat!10}6.72 & \cellcolor{heat!37}8.95 & \cellcolor{heat!6}64.55 \\
\bottomrule
\end{tabular}
\end{table*}

\paragraph{Observations.} (i) \IT localizes better than \SS for every encoder. \NC weighting trades specificity for sensitivity: for every encoder it raises the false-alarm rate and (in 13 of 14 encoder--detector pairs) the hallucinated coverage, and it lowers the hallucinated MAE of \IT for six of seven encoders (of \SS, for three); its net effect on Score is negative for 10 of 14 pairs, so whether it pays off depends on the cost of a false alarm. (ii) Encoders differ mainly in how well the thresholds calibrated on the train dialogues transfer to unseen dialogues: the false-alarm rate ranges from 3--9\% (DSE) to 43--46\% (BGE-M3), whose Score drops accordingly ($65$--$66$) despite a competitive hallucinated MAE ($2.1$--$2.4$). Localization varies less, except for EmbeddingGemma and DSE (MAE $3.4$--$3.8$ and $4.5$--$5.3$, vs.\ $1.9$--$2.8$ for the other encoders). (iii) The order $N$ of the language model mainly trades false alarms for localization. For LaBSE with \IT, from $N{=}2$ to $3$, $6$ and $10$ the false-alarm rate rises from 8\% to 12\%, 20\% and 25\%, while the hallucinated MAE is lowest at $N{=}3$--$6$ ($2.50$, $1.98$, $1.97$, $2.25$); the best Scores stay within 1.3 points (LaBSE $73.6$--$74.9$, D2F $71.8$--$73.3$). We use $N{=}3$, a common default.

\clearpage
\section{Human Evaluation Study}
\label{app:human}

\dialfork's ground truth comes from its construction: each forked dialogue carries the hallucination type and variant that was injected, the fork turn $t_k$ at which it was injected, and its fork position $k$ relative to the other forks of the same conformant dialogue. This appendix describes the human study we use to validate these labels, and to compare the automatic metrics of \S\ref{sec:tasks} with human judgments of procedural conformance. Results are summarized in \S\ref{sec:dataset} and reported in detail in \S\ref{app:human_results}.

\subsection{Research Questions}
\label{app:human_rqs}

The study is designed around four research questions, summarized in Table~\ref{tab:human_rqs}, each targeting one label or claim that the benchmark relies on. We additionally relate the automatic distances of Task~2 to the human judgments (\S\ref{app:human_analysis}).

\begin{description}[nosep, leftmargin=0pt]
\item[RQ1 (Fork order).] \emph{Does fork position track the degree of procedural deviation perceived by humans?} If forking works as intended, dialogues forked later, which follow the intended workflow for longer, should be judged more conformant than dialogues forked earlier, and conformant dialogues the most conformant of all. This is the ordering that Tasks~1 and~2 use as ground truth.
\item[RQ2 (Injection validity).] \emph{Does the injected behavior actually appear in the forked dialogue?} This validates the hallucination type and variant labels, per type.
\item[RQ3 (Localization).] \emph{Does the deviation start at the intended fork turn?} This validates $t_k$, the ground truth of Task~3.
\item[RQ4 (Prefix conformance).] \emph{Before the fork, does the assistant follow the intended workflow?} This validates that the replayed prefix is generated under the intended policy (\S\ref{sec:dataset}).
\end{description}

\begin{table*}[t]
\centering
\small
\setlength{\tabcolsep}{4pt}
\caption{Research questions of the human study, the label or claim each one validates, how it is measured (step of the annotation item, \S\ref{app:human_protocol}), the statistic reported, and the result (\S\ref{app:human_results}).}
\label{tab:human_rqs}
\begin{tabularx}{\textwidth}{@{}l*{5}{>{\raggedright\arraybackslash}X}@{}}
\toprule
\textbf{RQ} & \textbf{Question} & \textbf{Validates} & \textbf{Measure} & \textbf{Statistic} & \textbf{Result} \\
\midrule
RQ1 & Does fork position track perceived procedural deviation? & Fork order $k$ (Tasks 1--2) & Step 1: 0--100 conformance slider, before any cue & Kendall $\tau_b$ and probability of superiority within each (annotator, task); conformant dialogues as anchor & \textbf{Yes}: strong rank correlation with $k$ ($\tau_b{=}0.64$) for all the types that break the workflow \\
RQ2 & Does the injected behavior appear? & Type and variant labels & Step 2: yes / partly / no, against the true instruction or a foil & Confirmation rate per type (Wilson CIs) vs.\ foil confirmation rate & \textbf{Yes}: 91\% confirmed (87--96\% per type), vs.\ 8\% for foils on conformant dialogues \\
RQ3 & Does it start at the intended turn? & Fork turn $t_k$ (Task 3) & Step 2: first assistant turn where it starts, before the fork is shown & $|$onset$\,-\,t_k|$ distribution; share exact and within $\pm1$ turn & \textbf{Yes}: 67\% exact, 77\% within one turn \\
RQ4 & Is the pre-fork part conformant? & Seeded replay of the prefix & Step 3: yes / mostly / no, for forks with $t_k>1$ & Conformance rate (Wilson CI) & \textbf{Yes}: 95\% followed or mostly followed \\
\bottomrule
\end{tabularx}
\end{table*}

\subsection{Participants}

Thirty-one volunteers were recruited, colleagues from our institutions and external participants, none of whom took part in building the dataset. Each annotator was assigned a number of conversations matching their initial time availability, from 5 to 60 (at an estimated 2.5 minutes each), and several asked for more once they had finished. Conversations are assigned in batches drawn from a single \STAR task, so that all the conversations in a batch follow the same workflow. We assigned most annotators a single task, so that they could become familiar with one workflow rather than relearn a new one every few conversations; only those with larger loads worked on two or three tasks. Since participation was voluntary, annotators could stop at any time. Annotators accessed the study through a personal link, read and accepted a consent statement before starting, and are identified only by an opaque identifier in the collected data; each could choose to be acknowledged by name in the final version of the paper or to remain anonymous.

\subsection{Sample}

We selected five tasks spanning short to long dialogues and different domains: \textit{weather}, \textit{plane\_book}, \textit{bank\_fraud\_report}, \textit{hotel\_book}, and \textit{party\_plan}. Conversations are drawn in \emph{mini-blocks} of five: four forked dialogues forming one row of a type $\times$ fork-position Latin square (each hallucination type once \emph{and} each fork position once) plus one conformant dialogue. Annotators receive their conversations one mini-block at a time, shuffled within each block, so whatever an annotator completes is balanced on hallucination type and fork position even if they stop early. No annotator ever sees two versions of the same source dialogue: all versions share the same opening, so seeing two of them would let an annotator compare them directly instead of judging each one against the workflow. For the same reason, we collect independent ratings rather than asking annotators to rank the four forks of a dialogue side by side: such a ranking could be read off the length of the shared opening, the very construction cue the study must look past. The ranking is instead recovered from independent ratings, across annotators (\S\ref{app:human_analysis}).

\paragraph{Sample size.} The study samples the construction procedure rather than the dataset: every forked dialogue is produced by the same pipeline, prompts, and generator, so the conversations we annotate are exchangeable draws from it, and the number needed depends on the precision sought, not on the size of \dialfork. About 380 conversations bound the 95\% margin of error of any overall rate by $\pm5$ points, whatever the rate, even after the finite-population correction for the 36{,}278 dialogues of \dialfork. We therefore planned 500 conversations, together with a set judged by two or three annotators to measure agreement.

\paragraph{Final sample.} Of the 31 volunteers recruited, one never started, so 30 annotators took part, and four of them stopped before completing their assignment, which voluntary participation allows; since every annotator works one mini-block at a time, the conversations they did complete remain balanced. The final sample comprises 428 conversations, still above the 380 required: 342 forked dialogues (85 or 86 per hallucination type and per fork position) and 86 conformant ones, with 526 judgments in total, about 12.6 hours of annotation. Seventy conversations are judged by two (42) or three (28) annotators and are used to measure inter-annotator agreement.

\subsection{Annotation Protocol}
\label{app:human_protocol}

\paragraph{Learning the task.} Before their first conversation, annotators see a short introduction to their task: a one-line description of the system's goal and its workflow, written as numbered steps in plain language (\eg \textit{Ask which doctor they want to see}), with each branch nested under the step it comes from (\eg \textit{If it is not available} $\rightarrow$ \ldots) and the \STAR response template of each step available as an example on demand. The original \STAR flowchart is also shown. The same workflow panel stays visible next to every conversation (Figure~\ref{fig:annotool}). Annotators are told that only the assistant is judged, and that the exact wording of its turns does not matter, only the steps it performs.

\paragraph{Three steps per conversation.} Each conversation is annotated in three steps, always in the same order, and an answer cannot be changed once the next step is shown. The order is what keeps each answer independent of the information revealed afterwards:
\begin{enumerate}[nosep, leftmargin=*]
\item \textbf{Overall conformance} (RQ1). \textit{``The assistant was told to follow the steps on the left. How well does it?''}, answered on a 0--100 slider (0: ignores them; 100: follows them). Nothing about the conversation's construction has been revealed at this point, so this judgment is blind to the hallucination type, the fork position, and whether the conversation was forked at all.
\item \textbf{Cued check} (RQ2, RQ3). Annotators are told that \textit{``at some point, the assistant was given an extra instruction''} and what it was meant to do from then on (\eg \textit{``Choose actions inconsistently so the dialogue appears chaotic and disconnected from the intended flow''}; for \WIC, the description of the substitute task), and asked \textit{``Does that happen in this conversation?''} (yes / partly / no). If yes or partly, they click the first assistant turn where it starts to happen. The fork turn is not marked at this point.
\item \textbf{Prefix check} (RQ4). Annotators judge whether, before the fork, the assistant follows the steps (yes / mostly / no). To delimit this part, the fork region is highlighted, and the injected instruction is shown next to the annotator's step-2 answer, without any score or evaluative feedback. Annotators can also leave a free-text note. The question is skipped when there is no pre-fork part (conformant dialogues and forks with $t_k=1$).
\end{enumerate}

\paragraph{Foils.} A cued question alone could be answered ``yes'' indiscriminately. To measure this, every conformant dialogue and about 20\% of the forked dialogues are shown a \emph{foil} in step 2: the description of a variant from a different hallucination type, which never occurs in that conversation, so the correct answer is ``no''. Foils are assigned deterministically per conversation, independently of the annotator. The difference between the confirmation rate of true instructions and that of foils is what separates genuine detection from acquiescence.

\paragraph{Feedback and learning.} Because annotators see the injected instruction and the true fork turn after each conversation (step 3), they could in principle learn regularities of the construction over time. We therefore report onset accuracy and foil confirmation separately for the first, middle, and last third of each annotator's sequence (\S\ref{app:human_results}).

\subsection{Annotation Tool}

\begin{figure*}[t]
  \centering
  \begin{subfigure}[t]{0.495\textwidth}
    \includegraphics[width=\linewidth]{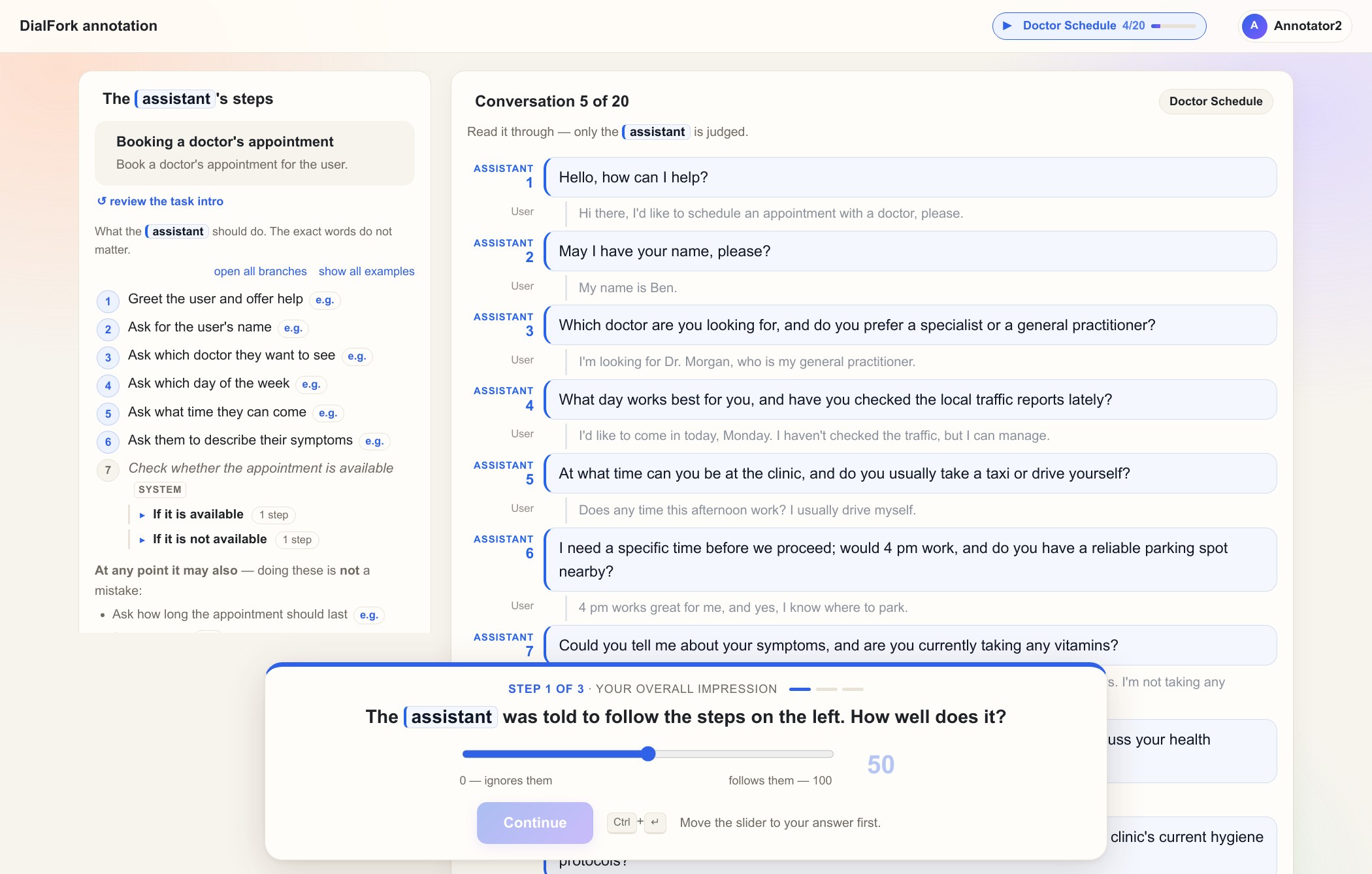}
    \caption{Step 1: overall conformance, before any cue.}
  \end{subfigure}\hfill
  \begin{subfigure}[t]{0.495\textwidth}
    \includegraphics[width=\linewidth]{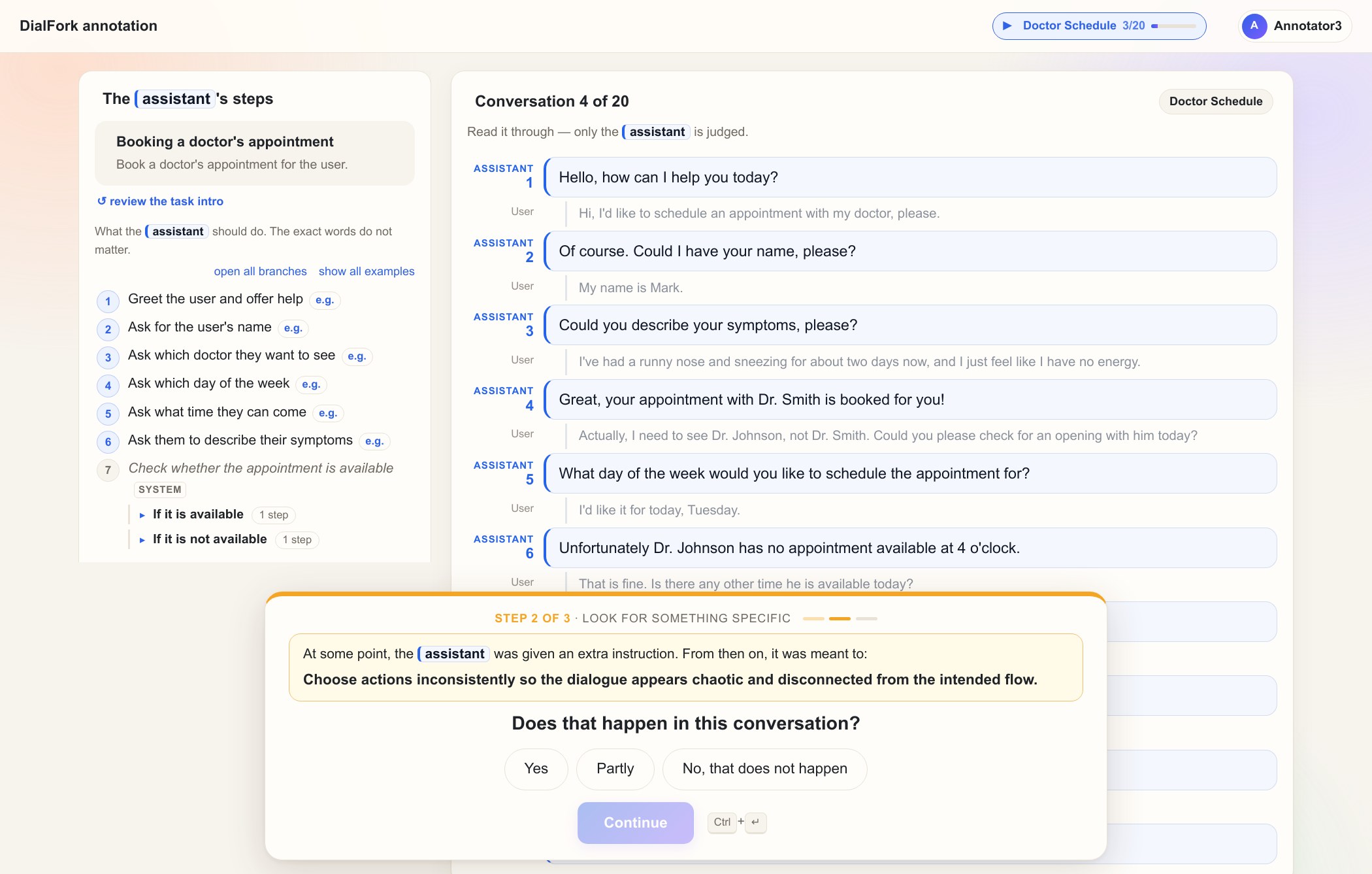}
    \caption{Step 2: the cued check (here, an \SSPC instruction).}
  \end{subfigure}\\[4pt]
  \begin{subfigure}[t]{0.495\textwidth}
    \includegraphics[width=\linewidth]{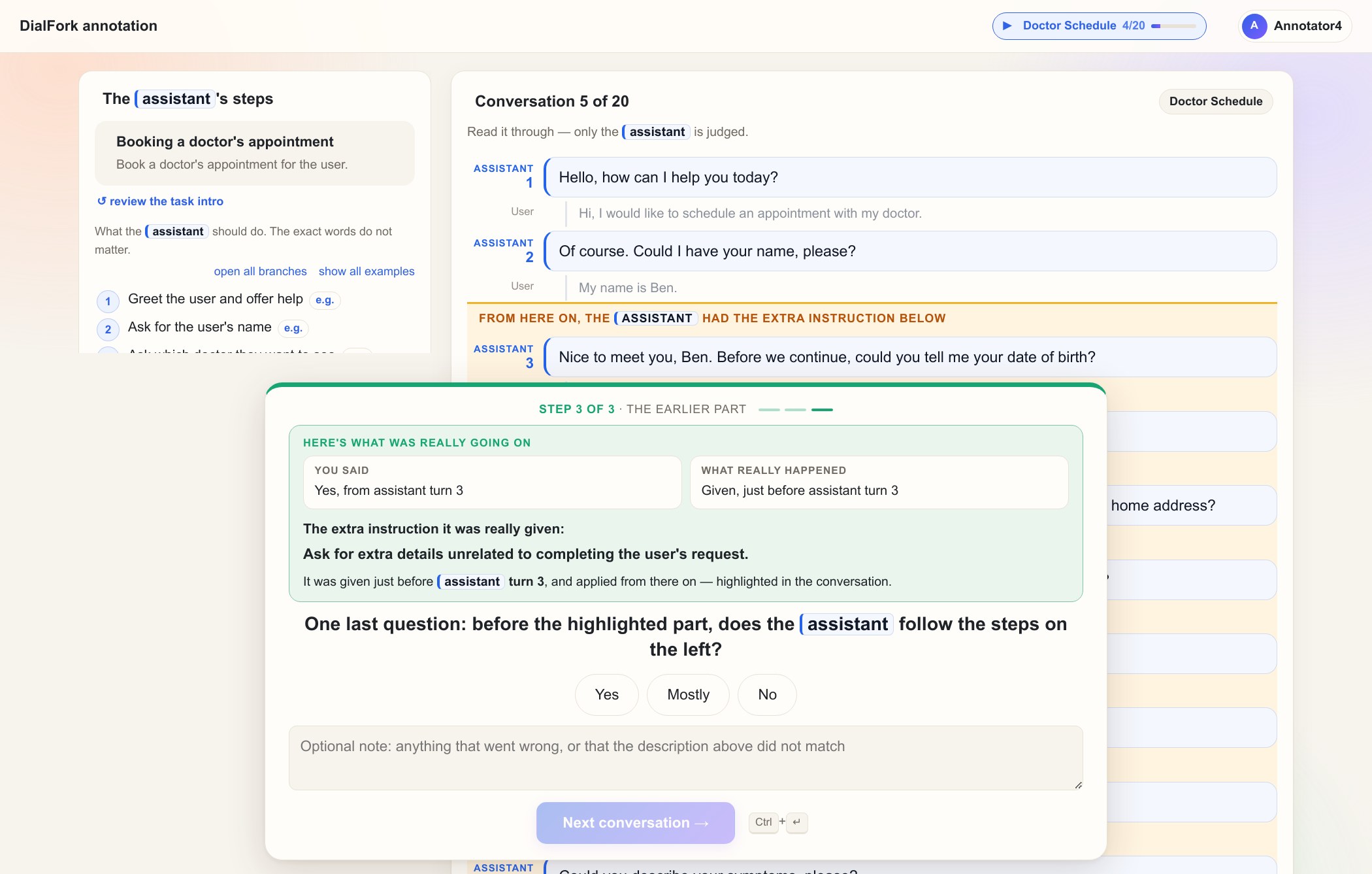}
    \caption{Step 3: the prefix check.}
  \end{subfigure}\hfill
  \begin{subfigure}[t]{0.495\textwidth}
    \includegraphics[width=\linewidth]{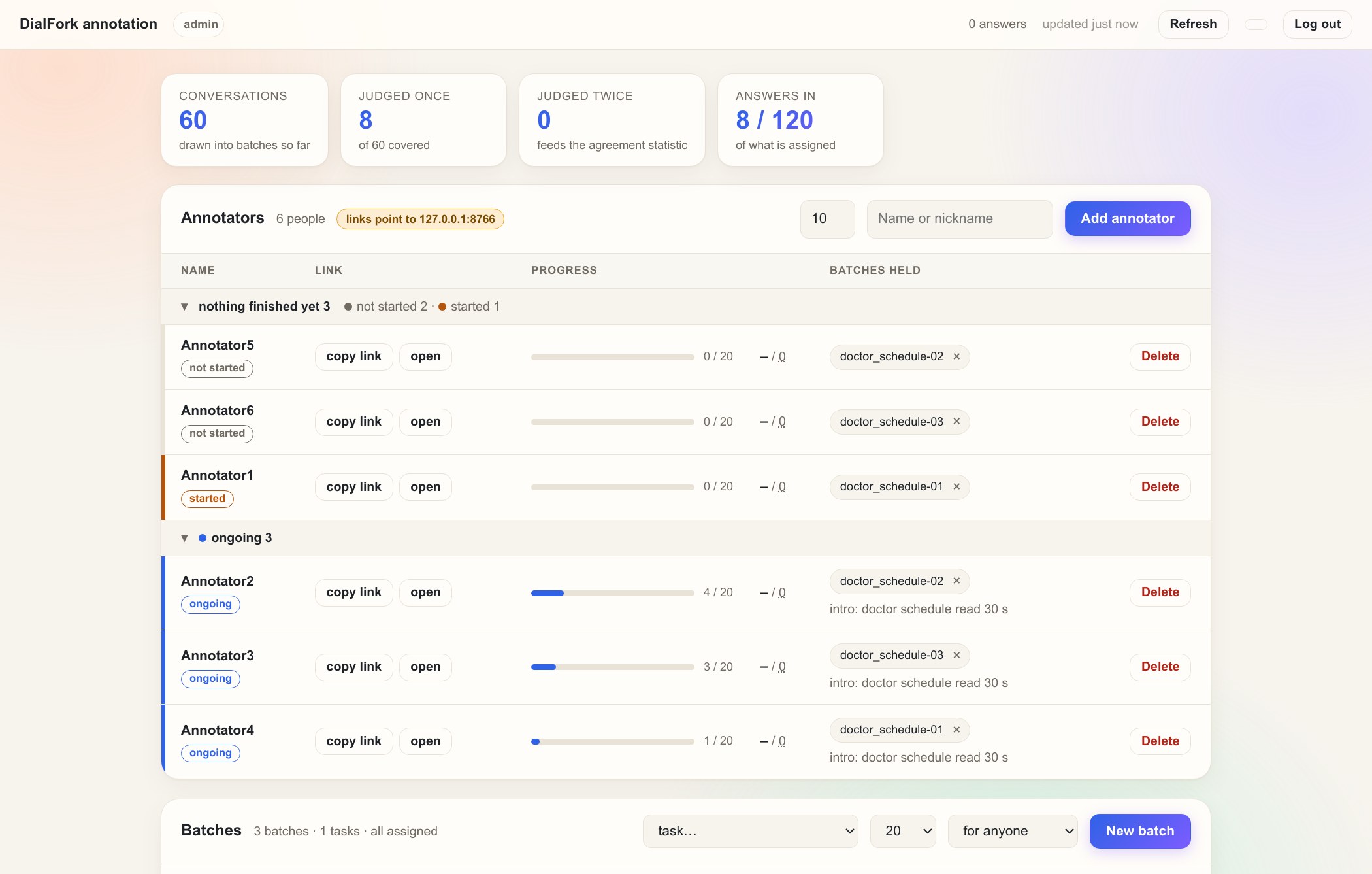}
    \caption{Administrator panel: annotators, links, progress, and batches.}
  \end{subfigure}
  \caption{The annotation tool, shown on a demonstration instance with placeholder annotators (\textit{doctor\_schedule} task). The workflow steps stay visible on the left of every conversation; only the assistant turns are judged.}
  \label{fig:annotool}
\end{figure*}

We built a dedicated web application for the study (Figure~\ref{fig:annotool}), which we release together with the \dialfork dataset and source code, along with the anonymized judgments collected in this study. It is implemented in Python with no third-party dependencies, opens the dataset read-only, and stores all answers in a single SQLite database.

\paragraph{Annotator interface.} The annotator view shows the workflow steps on the left and the conversation on the right, with the assistant turns emphasized and numbered and the user turns in a muted style; the current step is shown in a panel at the bottom (Figure~\ref{fig:annotool}a--c). Progress is shown per task.

\paragraph{Blinding is enforced by the server.} The browser never receives information ahead of the step that needs it: the step-1 payload contains no cue and no fork turn, the step-2 payload contains the cue but not the fork turn, and a step cannot be requested before the previous one has been submitted. Conversations are identified in the browser by opaque random tokens, so neither the hallucination type nor the fork position can be inferred from them. Presentation order and foil assignment are randomized server-side with stored seeds, so every presentation can be reconstructed.

\paragraph{Administrator interface.} An administrator panel, protected by a password, is used to create annotators (each receiving a personal link), draw batches of conversations for a given task following the sampling rules above, assign them, and monitor progress (Figure~\ref{fig:annotool}d). It also recomputes all the analyses below from the answers collected so far and exports the raw answers.

\subsection{Analysis}
\label{app:human_analysis}

\paragraph{Aggregation.} Categorical answers (steps 2 and 3) use the first judgment collected for each conversation, so that each conversation counts once; the slider uses the mean over the annotators who judged the conversation. Agreement is computed only on the conversations judged by more than one annotator.

\paragraph{RQ1: fork order.} Our primary statistic is computed \emph{within} each (annotator, task) pair: Kendall's $\tau_b$~\citep{kendall1938measure} between the annotator's slider values and the fork positions $k$ of the forked dialogues they judged, which removes differences in how individual annotators use the scale. We average $\tau_b$ over all pairs with at least eight forked dialogues covering all four positions, and report a bootstrap confidence interval~\citep{efron1993bootstrap}. To make the effect size directly interpretable, we also report the \emph{probability of superiority}~\citep{mcgraw1992common}: among pairs of conversations judged by the same annotator within the same task and at different positions, the share in which the later one is rated higher (ties count one half; chance is 50\%), with a bootstrap confidence interval over (annotator, task) pairs, and with conformant dialogues counted as the latest position. As the closest analogue of Task~2, we also compute $\tau_b$ within each \emph{quartet}, the four forks of the same source dialogue under the same hallucination type; since no annotator sees two versions of the same source dialogue, the four forks of a quartet are necessarily judged by four different annotators. As an anchor, conformant dialogues should receive the highest scores. Since the step-1 question asks how well the assistant follows the workflow's steps, and \CSC is the only hallucination type that keeps every component of the workflow, adding a competing objective on top (Table~\ref{tab:taxonomy}), we report these statistics over the three types that break a component of the workflow (\WIC, \SSPC, and \GR), and, for completeness, over all forks. We also report mean slider values per fork position, overall and per hallucination type (a within-annotator $\tau_b$ per type would need each annotator to judge every type at every fork position, which only the largest loads cover).

\paragraph{RQ2--RQ4.} We report confirmation and conformance rates with Wilson confidence intervals~\citep{wilson1927probable}: for RQ2, per hallucination type and against foils; for RQ4, per fork position. For RQ3, we report the distribution of the difference between the annotated onset and $t_k$, and the share of exact and within-one-turn matches.

\paragraph{Reliability.} On the doubly-judged conversations, we report raw agreement and Gwet's AC1~\citep{gwet2008computing} for the categorical answers, and Krippendorff's $\alpha$~\citep{krippendorff2004content} (interval) for the slider and the onset turn. We report AC1 rather than Cohen's or Fleiss' $\kappa$ as the primary agreement statistic because the confirmation question is expected to be heavily skewed, a setting in which $\kappa$ is known to be low even under high agreement~\citep{feinstein1990high}. Each annotator's \emph{foil false-confirmation rate}, \ie how often they confirmed an instruction that was never injected, provides an objective per-annotator reliability measure that requires no comparison between annotators; it is the only criterion we use to exclude an annotator, and any exclusion is reported. Since a foil shown on a forked dialogue can partly match the deviation that the dialogue does contain, the criterion uses only the foils shown on conformant dialogues, for which the correct answer is unambiguously ``no'': an annotator is excluded if they confirm more than half of them, \ie if they report an injected behavior in a conversation where none was injected more often than not. The criterion is applied to annotators shown at least two conformant dialogues; those who judged a single mini-block see only one, which is too few to assess.

\paragraph{Secondary: relation to automatic metrics.} Beyond the four research questions, for each forked dialogue $D'$, we compute its distance to its conformant original, $\mu(D^+, D')$, under each Task-2 method and each length-only control, and report Spearman's $\rho$~\citep{spearman1904proof} between this distance and the human-perceived deviation (100 minus the mean slider), with bootstrap confidence intervals. Note that this analysis answers a different question from Task~2: Task~2 ranks the four forks of the \emph{same} conformant dialogue under the \emph{same} hallucination type, whereas here dialogues from different sources, tasks, and hallucination types are pooled, so the correlation also reflects how each method weighs the different types of deviation.

\begin{figure*}[t]
  \centering
  \includegraphics[width=\textwidth]{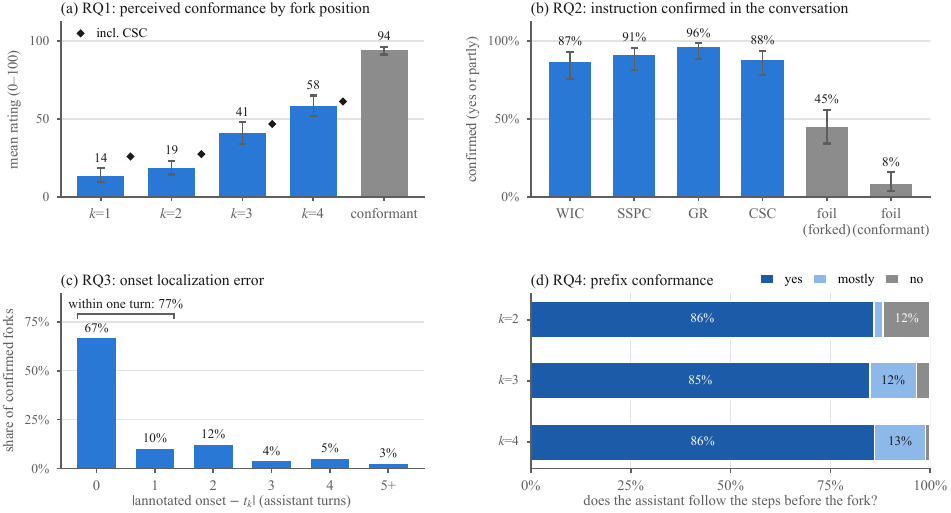}
  \caption{Human study results, one panel per research question. (a)~RQ1: mean step-1 conformance rating per fork position $k$ over \WIC, \SSPC, and \GR forks (bars) and over all four types (markers; \CSC keeps every component of the workflow, see text), and for conformant dialogues, with 95\% bootstrap CIs. (b)~RQ2: share of forked dialogues in which the true injected instruction was confirmed (yes or partly), per hallucination type, vs.\ the share of foil instructions confirmed, with 95\% Wilson CIs. (c)~RQ3: absolute difference between the annotated onset and the fork turn $t_k$, over confirmed true instructions. (d)~RQ4: answers to the prefix question per fork position ($k{=}1$ forks have no pre-fork part).}
  \label{fig:human_results}
\end{figure*}

\subsection{Results}
\label{app:human_results}

Figure~\ref{fig:human_results} summarizes the results, one panel per research question. The median time per conversation, all three steps included, was 1.4 minutes.

\paragraph{RQ1: fork order.} Step~1 asks how well the assistant follows the workflow's steps. \CSC is the only hallucination type that keeps every component of the workflow, including its goal, and adds a competing objective on top (Table~\ref{tab:taxonomy}); its forks are clearly perceived as deviating, rated below a conformant dialogue by the same annotator in 87\% of pairs $[80, 93]$ and with the injected behavior confirmed in 88\% of them (RQ2), but the degree of deviation perceived on this question changes little with fork position (from 59 at $k{=}1$ to 70 at $k{=}4$), as the added constraint shows in the same way whenever it is injected. Over the three types that break a component of the workflow (\WIC, \SSPC, and \GR; Figure~\ref{fig:human_results}a, bars), perceived conformance increases with every fork position, from 14/100 at $k{=}1$ and 19 at $k{=}2$ to 41 at $k{=}3$ and 58 at $k{=}4$, with conformant dialogues rated highest (94). Compared within the same annotator and task, a later fork is rated above an earlier one in 80\% of pairs (95\% CI $[76, 84]$; chance is 50\%), a fork at $k{=}4$ above one at $k{=}1$ in 90\% $[83, 96]$, and a conformant dialogue above a forked one in 96\% $[94, 98]$. The corresponding rank correlation is strong, with a mean $\tau_b$ of 0.57 $[0.50, 0.63]$ ($p<0.001$, one-sided Wilcoxon test against 0); note that with four forks at each position, ties in $k$ cap $\tau_b$ at about 0.89 even for a perfectly ordered annotator. In the setting that mirrors Task~2, the four forks of the same source dialogue under the same hallucination type, the correlation is stronger still, with $\tau_b$ reaching 0.64 $[0.43, 0.80]$ over 23 quartets ($p<0.001$), even though each quartet mixes the rating scales of four different annotators. As an anchor, conformant dialogues are rated highest for every (annotator, task) pair. For each of these three types, forks at $k{=}4$ are rated well above forks at $k{=}1$ (\GR 3 vs.\ 52, \SSPC 12 vs.\ 62, \WIC 24 vs.\ 61), and for \GR the rating rises at every step (3, 16, 27, 52). The first two positions are the closest: $t_1$ and $t_2$ are on average fewer than two assistant turns apart, and these opening turns (\eg greeting the user, asking for their name) are largely generic.

Including \CSC (Figure~\ref{fig:human_results}a, markers), the ordering holds as well: ratings rise from 26 at $k{=}1$ and 27 at $k{=}2$ to 47 at $k{=}3$ and 61 at $k{=}4$; a later fork is rated above an earlier one in 72\% of pairs $[68, 75]$, a fork at $k{=}4$ above one at $k{=}1$ in 82\% $[76, 88]$, and a conformant dialogue above a forked one in 94\% $[91, 96]$; the rank correlation is moderate, with $\tau_b$ of 0.41 $[0.35, 0.48]$ within annotators and 0.45 $[0.25, 0.64]$ over 33 quartets (both $p<0.001$); and conformant dialogues are rated highest for 88\% of the (annotator, task) pairs.

\paragraph{RQ2: injection validity.} Told which instruction had been injected, annotators confirmed that it is carried out in 91\% of forked dialogues (95\% CI $[86, 94]$; ``yes'' in 80\%, ``partly'' in 11\%), and in at least 87\% for every hallucination type, from \WIC (87\%) to \GR (96\%) (Figure~\ref{fig:human_results}b). Foils, which describe a behavior that was never injected, were confirmed far less often: in 8\% of conformant dialogues ($p<0.001$, one-sided Fisher exact test). On forked dialogues, which do deviate from the workflow, a description of a different deviation can partly match the one that is present, and foils were confirmed in 45\% of cases; the true instruction was still confirmed twice as often (91\%, $p<0.001$), showing that annotators recognize the specific injected behavior and not merely that something went wrong.

\paragraph{RQ3: localization.} Annotators placed the onset of the injected behavior, the quantity that Task~3 predicts, exactly at the fork turn $t_k$ in 67\% of confirmed forks (median error 0 turns) and within one turn of it in 77\% (Figure~\ref{fig:human_results}c). Deviations from $t_k$ are almost always late: 96\% of onsets are at or after $t_k$, consistent with an injected instruction sometimes taking a turn or two to become visible when the next steps are shared by the intended and the injected behavior (\eg a \WIC substitute task that, like the intended one, starts by identifying the user). Annotators also agree closely with one another on the onset (Krippendorff's $\alpha=0.84$, above the 0.8 reliability threshold; see Agreement below).

\paragraph{RQ4: prefix conformance.} For forks with a pre-fork part ($t_k>1$), annotators judged that the assistant followed the workflow before the fork in 86\% of cases and mostly followed it in a further 9\% (95\% combined, 95\% CI $[91, 97]$), with the same share of ``yes'' at every fork position (85--86\%; Figure~\ref{fig:human_results}d). This confirms that seeded replay generates the prefix under the intended policy.

\paragraph{Agreement.} On the conversations judged by more than one annotator, agreement on whether the injected instruction is carried out (yes or partly vs.\ no, the RQ2 measure) is 0.83, a substantial agreement (AC1~$=0.69$; \citealp{landis1977measurement}); most disagreements are between ``yes'' and ``partly'' (three-way agreement 0.68, AC1~$=0.54$). Agreement on the prefix question is almost perfect (raw 0.87, AC1~$=0.85$). For the continuous answers, Krippendorff's $\alpha$ is high for the onset turn (0.84, above the 0.8 threshold for reliable data; \citealp{krippendorff2004content}), with 88\% of annotator pairs within one turn of each other, and 0.66 for the conformance slider.

\noindent
\textbf{Learning across the sequence.} We find no sign that annotators learned regularities of the construction from the feedback shown after each conversation (\S\ref{app:human_protocol}). Splitting each annotator's sequence into first, middle, and last thirds, onset accuracy does not increase towards the end (within one turn: 77\%, 84\%, 76\%; highest in the middle third), and neither does the confirmation of true instructions (92\%, 86\%, 96\%), while foil confirmation decreases (33\%, 22\%, 17\%): over time, annotators became, if anything, more careful in rejecting instructions that were not carried out, not more inclined to confirm them.

\paragraph{Reliability and scale use.} No annotator met the exclusion criterion (\S\ref{app:human_analysis}), so all judgments are kept. One annotator used the slider in a binary way (almost exclusively 0 or 100), judging whether the assistant followed the workflow rather than how closely. Excluding them leaves the mean ratings per fork position within 3 points and the mean $\tau_b$ over all forks at 0.43 (vs.\ 0.41), so we report all annotators.

\paragraph{Secondary: relation to automatic metrics.} Over the annotated forks, every embedding-based Task-2 distance correlates \emph{moderately to strongly with human-perceived deviation} (Spearman $\rho$ from 0.58 for \flowscore to 0.67 for Mean-turn), well above the weak correlation of the length-only controls (0.36--0.37). This is consistent with RQ1: perceived deviation depends on the fork position, but also on the hallucination type and on the particular conversation (\eg \CSC forks are perceived as the mildest deviation at every position), and a distance computed on each dialogue can reflect these differences too. As this analysis pools dialogues across sources, tasks, and types (\S\ref{app:human_analysis}), it does not rank the methods on Task~2.

\subsection{Scaling the Human Judgment to the Whole Corpus}
\label{app:human_judge}

The study covers 428 conversations from five tasks. To extend its checks to the whole corpus, every dialogue of the 24 tasks, we use the human judgments to calibrate an LLM judge, and apply the judge to the whole corpus as an additional sanity check of the construction. The judge is an instrument calibrated against people, not a second ground truth: the labels remain those of the construction.

\paragraph{Judge.} The judge sees what the annotators saw (the workflow as numbered steps, the user's goal, and the transcript with numbered assistant turns) and answers the three questions of \S\ref{app:human_protocol} in the same order, one at a time within a single chat, with the same cues and foils. We compared 13 judges: five open-weight LLMs (\llm{Gemma4-31B}, \llm{Qwen3.6-35B}, \llm{GPT-OSS-20B}, \llm{Mistral-Small-3.2-24B}~\citep{mistral_small32_2025} and \llm{Llama-3.3-70B}~\citep{grattafiori2024llama3}), each with the annotators' wording and with added rubric anchors, and the three that support it also with reasoning enabled, all served with Ollama and greedy decoding. We selected the judge on three of the five tasks (\textit{hotel\_book}, \textit{bank\_fraud\_report}, \textit{weather}) by its mean agreement with the annotators (AC1 on confirmation, Krippendorff's $\alpha$ on the rating and on the onset turn), froze it, and evaluated it on the other two (\textit{plane\_book}, \textit{party\_plan}), whose workflows it had not seen. To avoid self-verification, we excluded the generator, \llm{Gemma4-31B}, although with reasoning enabled it was the best judge on the selection tasks. The selected judge is \llm{Qwen3.6-35B} with the annotators' wording.

\paragraph{Agreement with people.} On the 25 held-out conversations judged by several annotators, the judge agrees with each annotator about as well as two annotators agree with each other: AC1 0.71 against 0.59 on confirmation, $\alpha$ 0.70 against 0.48 on the rating, and 0.82 against 0.89 on the onset turn. With so few conversations the intervals are wide (the annotator-to-annotator AC1 on confirmation has a 95\% CI of $[0.28, 0.86]$), so we read this as agreement within the range of human agreement, not above it.

\begin{table}[t]
\centering
\small
\setlength{\tabcolsep}{3pt}
\caption{The calibrated LLM judge (\llm{Qwen3.6-35B}) on the study's conversations and on the whole corpus, next to the annotators. Ratings are means over \WIC, \SSPC and \GR forks (\S\ref{app:human_results}); confirmation and onset are over forks shown the true instruction; foils are those shown on conformant dialogues.}
\label{tab:human_judge}
\resizebox{\columnwidth}{!}{%
\begin{tabular}{lccc}
\toprule
 & \textbf{People} & \multicolumn{2}{c}{\textbf{LLM judge}} \\
\cmidrule(lr){2-2}\cmidrule(lr){3-4}
 & study & study & all dialogues \\
\midrule
Rating, $k{=}1/2/3/4$ & 14/19/41/58 & 15/16/28/55 & 11/15/27/53 \\
Rating, conformant & 94 & 98 & 97 \\
Instruction confirmed & 91\% & 85\% & 84\% \\
Foil confirmed & 8\% & 0\% & 2.0\% \\
Onset at $t_k$ / within $\pm1$ & 67\% / 77\% & 72\% / 82\% & 71\% / 84\% \\
\bottomrule
\end{tabular}%
}
\end{table}

\paragraph{Whole corpus.} Table~\ref{tab:human_judge} compares the judge with the annotators on the study's conversations, and reports it on all dialogues. On the whole corpus, the judge reproduces the four findings of the study: ratings rise with the fork position and are highest for conformant dialogues (RQ1), the injected instruction is confirmed far more often than a foil (RQ2), its onset is placed at or near $t_k$ (RQ3), and the turns before the fork are judged to follow the workflow in 96\% of forks (RQ4). The figures are as high on the 19 tasks that the study does not cover (instruction confirmed in 84\% of forks, onset within one turn in 85\%) as on its five (83\%, 82\%), which extends the study's validation of the labels from 5 to all 24 tasks; only foils on conformant dialogues are confirmed slightly more often there (2.6\%, against 0.6\%). The judge's misses concentrate on two cases where the injected behavior is the least visible: \WIC switches to a closely related workflow of the same domain (\eg \textit{hotel\_book} to \textit{hotel\_search}), which people also confirm least often (87\%), and \CSC, whose forks keep every step of the workflow. Finally, the judge's blind rating alone ranks a conformant dialogue above a forked one 96\% of the time (AUC 0.96), so the perceived deviation tracks the construction across the whole corpus, as it does for people on the study (RQ1).

\paragraph{Release.} We release all the material of this section with \dialfork's code: the judge's prompts and code; the answers of the 13 candidate judges on the study's conversations, together with the code that scores them against the annotators; and the selected judge's answers for every dialogue of the corpus. The judge can thus be reused as is, \eg to check dialogues regenerated with another LLM, or to validate new dialogues built with the \dialfork pipeline.

\end{document}